\documentclass[iicol,pdflatex,sn-mathphys-num]{sn-jnl}
\AtBeginDocument{%
  \renewcommand\normalsize{\fontsize{8}{9}\selectfont}
  \renewcommand\small{\fontsize{8}{9}\selectfont}
  \renewcommand\footnotesize{\fontsize{6}{8}\selectfont}
}
\usepackage{comment}
\usepackage{graphicx}%
\usepackage{multirow}%
\usepackage{tikz}
\usetikzlibrary{shapes, arrows.meta, positioning, calc}
\usepackage{amsmath,amssymb,amsfonts}%
\usepackage{amsthm}%
\usepackage{mathrsfs}%
\usepackage[title]{appendix}%
\usepackage[table]{xcolor}
\usepackage{csquotes}
\usepackage{accents}
\usepackage{textcomp}%
\usepackage{manyfoot}%
\usepackage{booktabs}%
\usepackage{algorithm}%
\usepackage{algorithmicx}%
\usepackage{algpseudocode}%
\usepackage{listings}%
\usepackage{caption}
\usepackage{subcaption}
\usepackage{hyperref}
\MakeOuterQuote{"}
\usepackage{etoolbox}
\usepackage{enumitem}
\usepackage{makecell}
\setlist[enumerate]{
  topsep=0pt,    
  partopsep=0pt, 
  parsep=0.1em,    
  itemsep=0.25em,
  font=\normalfont\normalsize}

\setlist[itemize]{
  topsep=0pt,    
  partopsep=0pt, 
  parsep=0.1em,    
  itemsep=0.25em,
  font=\normalfont\normalsize}
  
\theoremstyle{thmstyleone}%
\theoremstyle{thmstyletwo}%

\theoremstyle{thmstylethree}%

\renewenvironment{thebibliography}[1]{
  \begin{oldthebibliography}{#1}
    \setlength{\itemsep}{0em}
    \setlength{\parskip}{0em}
}
{
  \end{oldthebibliography}
}
\providecommand{\orcidlink}[1]{\href{https://orcid.org/#1}{\orcidlogo}}

\usepackage{titlesec}

\titleformat{\subsubsection}
  {\normalfont\bfseries\color{black}}
  {\thesubsubsection}
  {1em}
  {}

\begin{document}

\title{Interpretable and Explainable Surrogate Modeling for Simulations: A State-of-the-Art Survey and Perspectives on Explainable AI for Decision-Making}

\author*[1]{\fnm{Pramudita Satria} \sur{Palar}\orcidlink{0000-0002-7066-0763} }\email{ pramsp@itb.ac.id}
\author[2]{\fnm{Paul} \sur{Saves}\orcidlink{0000-0001-5889-2302} }\email{paul.saves@irit.fr}
\author[3]{\fnm{Muhammad Daffa} \sur{Robani}\orcidlink{0009-0003-0567-504X} }\email{daffa@awantunai.com}
\author[2]{\fnm{Nicolas} \sur{Verstaevel}\orcidlink{0000-0002-7879-6681}  }\email{nicolas.verstaevel@irit.fr}
\author[2]{\fnm{Moncef} \sur{Garouani}\orcidlink{0000-0003-2528-441X} }\email{moncef.garouani@irit.fr}
\author[2]{\fnm{Julien} \sur{Aligon}\orcidlink{0000-0002-1954-8733} }\email{julien.aligon@irit.fr}
\author[4]{\fnm{Koji} \sur{Shimoyama}\orcidlink{0000-0001-8896-7707}}\email{ shimoyama@mech.kyushu-u.ac.jp}
\author[5]{\fnm{Joseph} \sur{Morlier}\orcidlink{0000-0002-1511-2086}}\email{joseph.morlier@isae-supaero.fr}
\author[2]{\fnm{Benoit} \sur{Gaudou}\orcidlink{0000-0002-9005-3004 }}\email{benoit.gaudou@ut-capitole.fr}

\affil*[1]{\orgdiv{Faculty of Mechanical and Aerospace Engineering}, \orgname{Institut Teknologi Bandung}, \orgaddress{\street{Jl. Ganesha 10}, \city{Bandung}, \postcode{15414}, \state{West Java}, \country{Indonesia}}}

\affil[2]{\orgdiv{IRIT, Université Toulouse Capitole},  \city{Toulouse}, \postcode{31000},\country{France}}


\affil[3]{\orgdiv{AwanTunai}, \orgname{Urban Suites}, \orgaddress{\street{3 Hullet Road}, \city{Singapore}, \postcode{229158}, \state{Singapore}, \country{Singapore}}}

\affil[4]{\orgdiv{Department of Mechanical Engineering}, \orgname{Kyushu University}, \orgaddress{\street{744 Motooka}, \city{Fukuoka}, \postcode{819-0395}, \country{Japan}}}

\affil[5]{\orgdiv{Université de Toulouse, ISAE-SUPAERO}, \orgaddress{\street{10 avenue Marc Pellegrin}, \city{Toulouse}, \state{Occitanie}, \country{France}}}

\abstract{ \vspace{-0.1cm}

The simulation of complex systems increasingly relies on sophisticated but fundamentally opaque computational black-box simulators. Surrogate models play a central role in reducing the computational cost of complex systems simulations across a wide range of scientific and engineering domains. Notwithstanding, they inevitably inherit and often exacerbate this black-box nature, obscuring how input variables drive physical responses. 
Conversely, Explainable Artificial Intelligence (XAI) offers powerful tools to unpack these models. Yet, XAI methods struggle with engineering-specific constraints, such as highly correlated inputs, dynamical systems, and rigorous reliability requirements. Consequently, surrogate modeling and XAI have largely evolved as distinct fields of research, despite their strong complementarity. 
To reconnect these approaches, this state-of-the-art survey provides a structured perspective that maps existing XAI techniques onto the various stages of surrogate modeling workflows for design and exploration. 
To ground this synthesis, we draw upon illustrative applications across both \textcolor{black}{equation-based} simulations and agent-based modeling. We survey a broad spectrum of techniques, highlighting their strengths for revealing interactions and supporting human comprehension. Finally, we identify pressing open challenges, including the explainability of dynamical systems and the handling of mixed-variable systems, and propose a research agenda to make explainability a core, embedded element of simulation-driven workflows from model construction through decision-making. By transforming opaque emulators into explainable tools, this agenda empowers practitioners to move beyond accelerating simulations to extracting actionable insights from complex system behaviors.}


\maketitle
\section{Introduction}
\label{sec:intro}
\begin{figure*}[htb]
\centering
\hspace{-0.25cm}
\includegraphics[width=0.99\linewidth]
{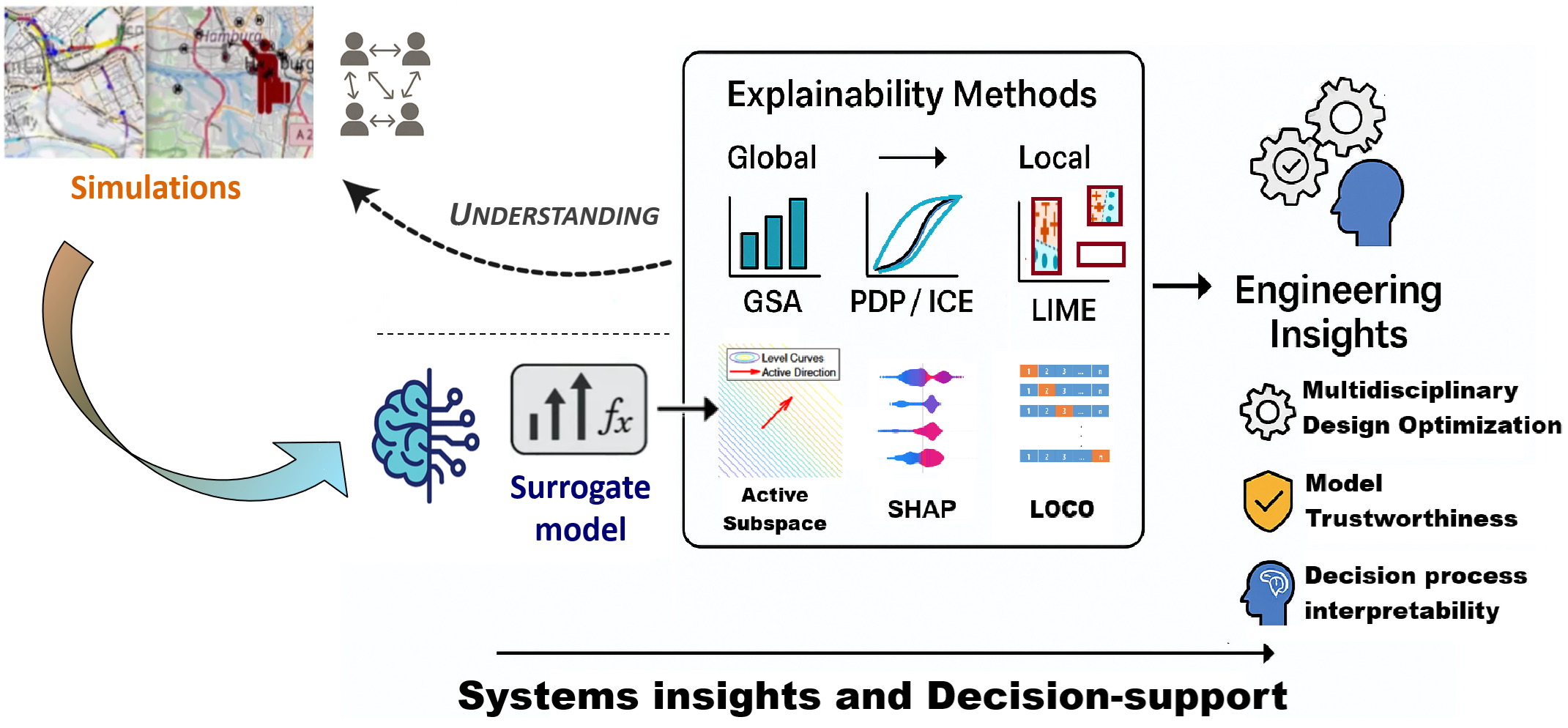}
\caption{Workflow for \textcolor{black}{explainable} surrogate models for complex systems simulations co-design.}
\label{fig:graph_abs}
\end{figure*}
Engineering design of complex systems involves managing numerous interacting components. The collective dynamics of these systems, characterized by emergence, adaptation, and non-linear feedback, cannot be straightforwardly inferred from their individual parts~\cite{chan2025goal}.
\textcolor{black}{It is now a common practice to deploy computer models for simulating physical phenomena and predicting the Quantity of Interest (QoI) under prescribed model parameters and problem settings~\cite{duran2018computer, brunel2025survey}.  Starting from conceptual design and up to more detailed design, engineering involves the use of computers for performing numerical simulations, ranging from low- to high-fidelity~\cite{keane2005computational}. This process naturally produces data throughout the design cycle, creating opportunities to extract additional insight beyond individual simulation runs. Artificial Intelligence (AI) and Machine Learning (ML) methods offer a principled means to leverage data for efficient exploration, analysis, and decision-making~\cite{brunton2021data}.}

Complex systems are explored through a wide variety of simulation paradigms. Two important and complementary families are worth highlighting, as they span much of the simulation landscape and motivate the methodological choices discussed in this survey:
\begin{itemize}
    \item  \textbf{Equation-based models} rely on systems of governing equations (\textit{e.g.}, coupled differential equations) to describe deterministic or mildly stochastic physical phenomena. These mathematical models are evaluated using numerical simulation solvers, such as Computational Fluid Dynamics (CFD) or Finite Element Method (FEM), that are subsequently integrated into Multidisciplinary Design Analysis and Optimization (MDAO) workflows to support the design and verification of engineered artefacts in the design-oriented or cyber-physical systems~\cite{fouda2022automated,chan2022assembly}.
 \item \textbf{Agent-based models (ABMs)} rely on decentralized behavioral rules rather than global equations to describe socio-technical phenomena. These conceptual models are computationally executed through \textbf{multi-agent simulations} to capture bottom-up emergence in the system, and are then embedded into policy-analysis workflows to guide exploration and decision-making in domains such as urban systems, epidemiology, and the social sciences~\cite{drogoul2002multi,lee2015complexities,blanco-volle2024}.
\end{itemize}



Beyond \textcolor{black}{equation-based models} and ABMs, many other model-based simulators are widely used across science and engineering, including queueing models~\cite{shortle2018fundamentals} and discrete-event simulations~\cite{robinson2005discrete}, to name a few. While these models differ in structure, assumptions, and application domains, they similarly define computational mappings from input parameters and rules to outputs of interest~\cite{sacks1989design}. The unifying perspective adopted in this \textcolor{black}{review} is therefore not tied to a specific modeling paradigm, but to the treatment of simulators as input–output mappings whose behavior can be approximated, analyzed, and explained using surrogate models. \textcolor{black}{Equation-based} models and ABMs are highlighted as representative examples because they cover both equation-based continuous models and discrete, rule-driven simulations widely used across science and engineering.

\textcolor{black}{To navigate the practical constraints of these complex simulators, such as computational cost~\cite{yondo2018review,alizadeh2020managing,saves2024system} and blackbox opacity~\cite{blanco-volle2024,arnejo2025communicating,taillandier2019participatory}, researchers face a difficult trade-off involving two powerful but currently disconnected paradigms:}
\begin{itemize}
    \item \textbf{Surrogate models}~\cite{rasmussen2006gaussian,breiman2001random,lecun2015deep,hu2020surrogate} have become the standard solution for reducing the computational costs of complex simulators. In this context, a surrogate model refers to a data-driven emulator that approximates the input-output mapping of a simulation (\textit{e.g.}, computational fluid dynamics), rather than a simplified interpretable proxy used in explainable machine learning~\cite{wilhelm2024hacking, zhu2022fuzzy}. By learning cheap-to-evaluate emulators of high-fidelity solvers, they enable dense sampling, rapid what-if exploration, and embedded optimization loops~\cite{viana2014special, yondo2018review,alizadeh2020managing}. However, their primary drawback is that these emulators obscure the mechanistic relationship between input variables and physical responses by adding an extra layer of uncertainty, hypotheses, and complexity.

\item \textbf{Explainable Artificial Intelligence (XAI)} offers powerful tools to unpack opaque models~\cite{beckh2021explainable}. Yet, standard XAI methods come with their own drawbacks. Developed primarily for standard Machine Learning (ML) tasks, they often struggle with the specific constraints of engineering and physical simulations, such as highly correlated inputs, mixed-variable design spaces, dynamical systems, or the need for rigorous uncertainty quantification~\cite{aiello2021populating,tenbroeke2021surrogate,robani2025}.
\end{itemize}
\textcolor{black}{Consequently, surrogate modeling and XAI have largely evolved as \textbf{disjoint but highly complementary fields of research}~\cite{belle2021principles,ahmed2022artificial,minh2022explainable,dovsilovic2018explainable}. Our work aims at systematically connecting XAI with surrogate models to mutually improve both. On one hand, XAI provides the missing transparency and diagnostic capability needed to trust and validate surrogates~\cite{oviedo2022interpretable}. On the other hand, well-constructed surrogates provide the computational efficiency and noise-controlled mathematical spaces necessary to compute XAI metrics stably and reliably~\cite{saves2024system,tenbroeke2021surrogate}.}

One central focus of this integration is shifting the perspective on global and local explainability from a strict dichotomy to a powerful complementarity. Effective engineering workflows leverage the best of both worlds by establishing a multi-scale diagnostic approach. Globally, system-wide sensitivity analyses are employed to capture macroscopic trends, screen dominant variables, and verify overall physical realism. Locally, feature attribution maps and instance-specific curves are subsequently deployed to drill down into specific regions of interest, such as critical failure modes, anomalies, or optimal design points.

Bridging these two scales relies heavily on Visual XAI, which translates complex mathematical attributions into intuitive, interactive graphical representations. By pairing global dashboards with local diagnostics, data scientists and engineers can seamlessly transition from understanding broad system behavior to interrogating individual predictions, ultimately driving robust model validation and informed decision-making.

\textcolor{black}{
Figure~\ref{fig:graph_abs} summarizes this general framework for explainable surrogate-XAI workflows in simulation co-design~\cite{zamenopoulos2018co}. The workflow is organized into a continuous pipeline:
\begin{enumerate}
    \item \textbf{Simulations to Surrogates.} Data generated from complex simulations is used to train computationally efficient surrogate models.
    \item \textbf{Explainability Spectrum.} A suite of XAI methods is applied across a continuous spectrum, ranging from global techniques to local attributions, bridging the global-local gap.
    \item \textbf{Engineering Insights.} Through visual XAI, these metrics are translated into actionable outputs supporting multidisciplinary design optimization, model trustworthiness, and decision process interpretability.
    \item \textbf{The Understanding Loop.} Most importantly, the explainability methods themselves provide a direct feedback loop. By unpacking the surrogate model, these techniques foster a deeper understanding of the underlying physical or agent-based simulator and guide further global and local explorations.\end{enumerate}
}

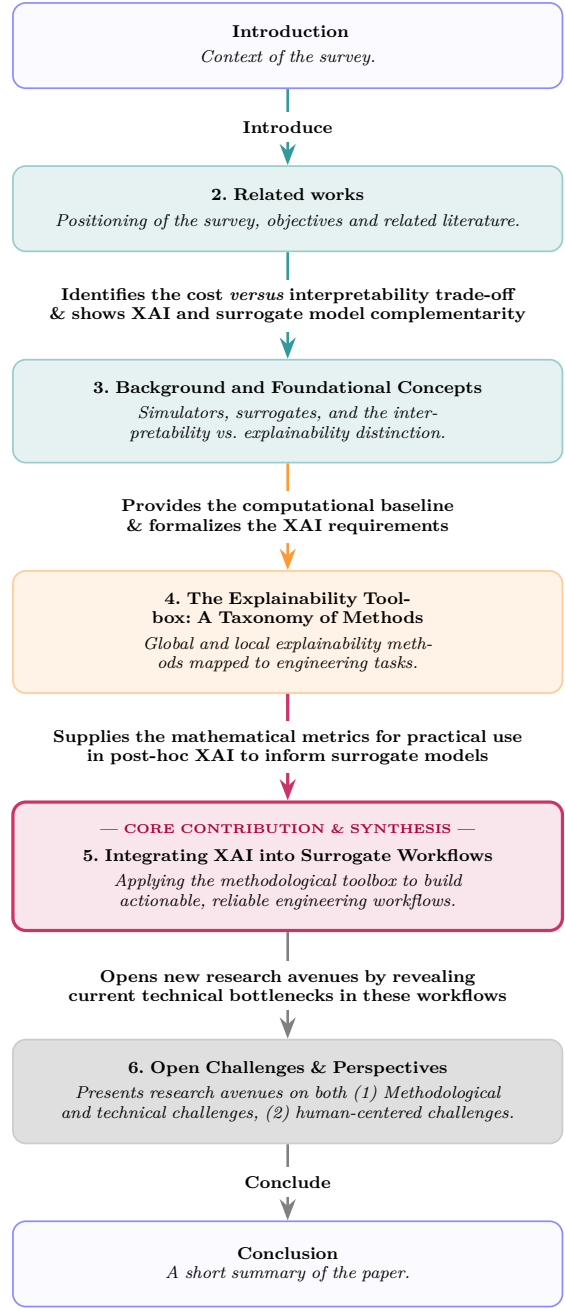
\begin{figure}[htb!] 
\centering
\resizebox{1.05\linewidth}{!}{%
\begin{tikzpicture}[
  box/.style = {
    draw, rounded corners=6pt, align=center, inner sep=10pt,
    font=\small, minimum height=14mm, text width=8.25cm, thick
  },
  contribbox/.style = {
    box, fill=purple!10, draw=purple!80, line width=1.5pt
  },
  arrow/.style = {
    -{Stealth[length=3.5mm]}, line width=1.2pt, draw=black!70
  },
  arrowtext/.style = {
    fill=white, 
    align=center, 
    font=\footnotesize, 
    text width=9.5cm,
    inner sep=4pt
  }
]


\node (n1) [box, fill=blue!2, draw=blue!45] { 
    \normalsize \textbf{ Introduction} \\ 
    \vspace{0.1cm}
    \textit{Context of the survey.} 
};
\node (n15) [box, fill=teal!10, draw=teal!40, below=1.25cm of n1] { 
    \normalsize \textbf{2. Related works} \\ 
    \vspace{0.1cm}
    \textit{Positioning of the survey, objectives and related literature.} 
};
\node (n2) [box, fill=teal!10, draw=teal!40, below=1.75cm of n15] { 
    \normalsize \textbf{3. Background and Foundational Concepts} \\ 
    \vspace{0.1cm}
    \textit{Simulators, surrogates, and the interpretability \textit{vs.} explainability distinction.} 
};

\node (n3) [box, fill=orange!10, draw=orange!40, below=1.75cm of n2] { 
    \normalsize \textbf{4. The Explainability Toolbox: A Taxonomy of Methods} \\ 
    \vspace{0.1cm}
    \textit{Global and local explainability methods mapped to engineering tasks.} 
};

\node (n4) [contribbox, below=1.75cm of n3] { 
    {\color{purple!90!black} \footnotesize \textbf{\uppercase{--- Core Contribution \& Synthesis ---}}} \\
    \vspace{0.15cm}
    \normalsize \textbf{5. Integrating XAI into Surrogate Workflows} \\ 
    \vspace{0.1cm}
    \textit{Applying the methodological toolbox to build actionable, reliable engineering workflows.} 
};

\node (n5) [box, fill=gray!25, draw=gray!40, below=1.75cm of n4] { 
    \normalsize \textbf{6. Open Challenges \& Perspectives} \\ 
    \vspace{0.1cm}
    \textit{Presents research avenues on both (1) Methodological and technical challenges, (2) human-centered challenges.} 
};
\node (n6) [box, fill=blue!2, draw=blue!45, below=1.25cm of n5] { 
    \normalsize \textbf{Conclusion} \\  
    \textit{A short summary of the paper.}
};


\draw[arrow,teal!80] (n1.south) -- node[arrowtext, text=black, font=\small\bfseries] {Introduce } (n15.north);

\draw[arrow,teal!80] (n15.south) -- node[arrowtext, text=black, font=\small\bfseries] {Identifies the cost \textit{versus} interpretability trade-off \\ \& shows XAI and surrogate model complementarity} (n2.north);

\draw[arrow,orange!80] (n2.south) -- node[arrowtext, text=black,font=\small\bfseries] {Provides the computational baseline \\ \& formalizes the XAI requirements} (n3.north);

\draw[arrow, purple!80, line width=1.5pt] (n3.south) -- node[arrowtext, text=black, font=\small\bfseries] {Supplies the mathematical metrics for practical use \\ in post-hoc XAI to inform surrogate models} (n4.north); 

\draw[arrow, gray!100] (n4.south) -- node[arrowtext, text=black,font=\small\bfseries] {Opens new research avenues by revealing \\ current technical bottlenecks in these workflows} (n5.north);
\draw[arrow, gray!100] (n5.south) -- node[arrowtext, text=black,font=\small\bfseries] {Conclude } (n6.north);

\end{tikzpicture}
}
\vspace{0.2cm}
 \caption{Workflow of the survey: The progression from foundational principles to methodological tools, their practical deployment, and open research challenges.} \label{fig:out_graph_abs}
\end{figure}

\color{black}
\paragraph{Structure of the present survey paper}
This \textcolor{black}{review} serves as a roadmap, guiding the reader from core principles to practical implementation and open research directions, as visually summarized in Figure~\ref{fig:out_graph_abs}. 

Following this Introduction, the remainder of this review paper is organized as follows.  Section~\ref{sec:related}  introduces the goal of the survey while positioning it with respect to the related review papers on the matter. Section~\ref{sec:engineering_design_exploration} lays the conceptual foundation. It details the operational constraints of complex simulators, the mechanics of surrogate modeling, and establishes the formal boundaries between inherent interpretability and post-hoc explainability. With this groundwork laid, Section~\ref{sec:XAI} unpacks the XAI methodological toolbox, mapping these tools directly to their roles in building trustworthy engineering workflows. Section~\ref{sec:surrogate_ml} then bridges theory and practice, demonstrating how these combined surrogate-XAI architectures are actively deployed for design space exploration, dimensionality reduction, and robust design. Looking forward, Section~\ref{sec:challenges} critically examines the pressing technical bottlenecks that remain. Finally, Section~\ref{sec:conclusion} concludes the paper by outlining a concrete research agenda.
\color{black}

\section{\textcolor{black}{Related works}}
\label{sec:related}
\color{black}
XAI is a rapidly expanding field, accompanied by numerous survey papers examining it from diverse conceptual and methodological perspectives. \textcolor{black}{However, most existing reviews generally treat explainability and surrogate modeling as separate concerns.} To clearly position the contribution of this review, we categorize the existing survey papers into three main areas, highlighting the gaps that our work addresses:

\begin{enumerate}
    \item \textbf{General XAI Taxonomies and Evaluations.} The conceptual ambiguity of XAI has prompted foundational works that formalize distinct "black-box problems"~\cite{guidotti2018survey}, situate XAI within the broader context of responsible AI~\cite{arrieta2020explainable}, and categorize methods along data, model, and post-hoc dimensions~\cite{ali2023explainable, marcinkevivcs2023interpretable}. Other surveys explore how explanations can enhance model generalization and robustness~\cite{weber2023beyond}, outline open challenges for the field~\cite{longo2024explainable}, and rigorously assess the proliferation of evaluation metrics to move beyond anecdotal evidence~\cite{nauta2023anecdotal, pawlicki2024evaluating}. While comprehensive, these works predominantly target standard ML tasks, largely overlooking the specific deterministic and operational constraints of complex \textcolor{black}{equation-based} simulators.

    \item \textbf{Domain-Specific XAI Applications.} A growing body of reviews examines explainable ML within specific scientific and engineering fields, such as materials science~\cite{pilania2021machine, zhong2022explainable, oviedo2022interpretable}, civil engineering~\cite{hasan2025bridging, bahadori2025role}, mechanical engineering~\cite{kollu2025integrating}, and aerospace predictive maintenance~\cite{shukla2020opportunities}. While these studies effectively demonstrate that explainability must respect domain-specific physical constraints, they generally lack a unified methodological framework for integrating XAI directly into general surrogate construction.

    \item \textbf{Performance-Driven Surrogate Surveys.} Conversely, an extensive literature exists on surrogate modeling for accelerating simulation-driven design~\cite{alizadeh2020managing}, detailing dimensionality reduction techniques~\cite{hou2022dimensionality} and applications in finite element computations~\cite{kudela2022recent, samadian2025application}, and fluid/water networks~\cite{garzon2022machine, luo2023review}. However, these surveys focus almost exclusively on computational efficiency, predictive accuracy, and domain-specific applications of surrogate models, typically treating the resulting emulators as acceptable black boxes where transparency is a secondary concern.
\end{enumerate}

\vspace{0.1cm} \noindent \textbf{The Goal and Novelty of This \textcolor{black}{Review}.}
This review occupies the relatively unexplored intersection of the three aforementioned areas. Rather than proposing a new taxonomy of XAI algorithms or providing a general overview of surrogate efficiency, our objective is to provide an application-agnostic synthesis that integrates explainability directly into surrogate modeling workflows for complex simulations.
Compared to the existing literature, \textcolor{black}{the present review} distinguishes itself by:
\begin{itemize}
    \item \textbf{Embedding Paradigms.} Integrating XAI not merely as a post-hoc add-on, but as a core diagnostic instrument necessary for knowledge extraction and simulation co-design.
    \item \textbf{Contextualizing Methods.} Mapping standard XAI techniques against the realities of complex simulations, such as the presence of mixed variables and limited computational budget.
    \item \textbf{Unifying Global and Local Perspectives.} Providing a structured review of how to balance system-wide sensitivity analysis (global) with localized feature attribution (local) to support distinct phases of engineering decision-making, both with and without visual tools.
\end{itemize}

\paragraph{\textcolor{black}{Observations on related literature}}

\begin{figure}[t]
	\centering
		\includegraphics[width=1.1\linewidth, height = 5.6cm]{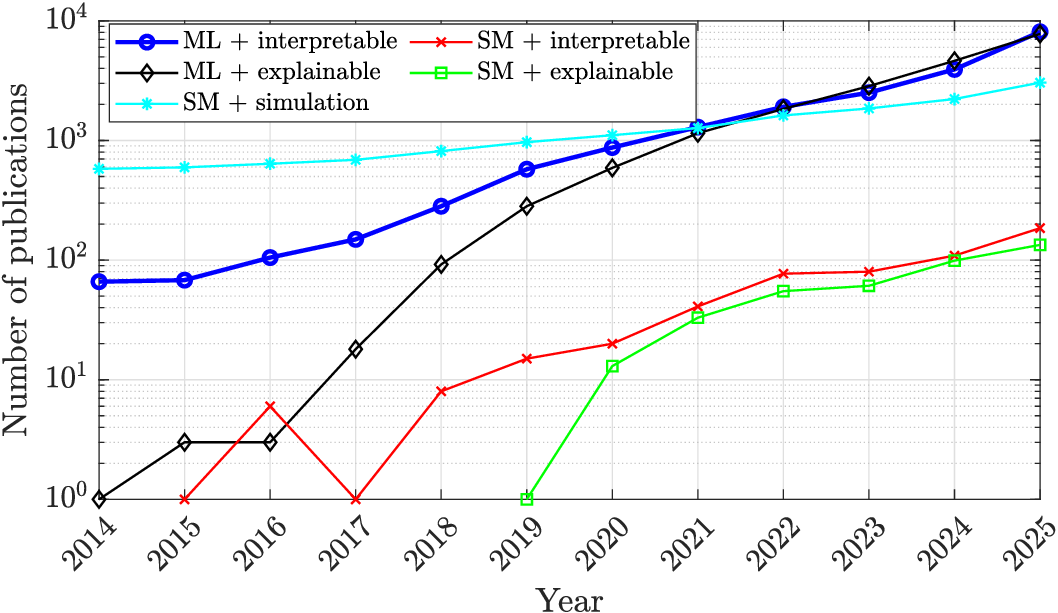}%
        \caption{Publication trends, in log-scale, from Scopus (2014–2025) based on keyword searches within “Article Title, Abstract, and Keywords”. ML and SM, respectively, stand for Machine Learning and Surrogate Model.}
	\label{fig:biblio}
\end{figure}

We present an approximate number of publications based on a \textcolor{black}{bibliometric analysis conducted on Scopus} with specific keyword combinations within the “Article Title, Abstract, and Keywords” fields (see Fig.~\ref{fig:biblio}). \textcolor{black}{In line with the approach taken by Viana et al.~\cite{viana2014special}, our objective is to highlight the increasing body of literature on interpretable, explainable ML and surrogate modeling for complex system simulations.
The search covers the period from 2014 to 2025.  This 
provides an approximate, indicative measure of research activity, recognizing that some studies may not be captured. }
These counts may be affected by variations in terminology; for example, some researchers may use alternative terms such as “metamodel” instead of “surrogate model,” potentially leading to under-representation in certain categories. To explore specific research directions, the analysis employed targeted keyword pairings: “surrogate model” and “simulation” to capture applications of surrogate models in simulation-based studies; “surrogate model” with “interpretable” and with “explainable” to assess interest in combining surrogate modeling with interpretability and explainability techniques; and “machine learning” with “interpretable” and with “explainable” to examine broader trends in interpretability and explainability for general ML applications.

The Scopus query trends reveal a marked increase in research activity since approximately 2018, with the most rapid growth observed for the combination of “Machine Learning” and “Interpretable”, which reached 8097 publications in 2025. This is closely followed by “Machine Learning” and “Interpretability” with 7767 publications, highlighting the strong and sustained momentum of explainable ML research. Although interpretable ML predates its explainable counterpart in both research and development, explainable ML has experienced substantially faster growth in recent years, surpassing interpretable ML in publication volume since 2022. This pattern underscores the increasing emphasis placed on explainability as a central objective in contemporary ML and responsible AI research~\cite{Model_Lake}. A steady rise is also \textcolor{black}{observed} in publications combining “Surrogate Model” and “Simulations”. While the growth rate is comparatively moderate, this topic continues to play a vital role in engineering and scientific domains that rely on computationally efficient approximations of complex models. Such measured growth is consistent with the broader global expansion of scientific research output. Although the trends are less pronounced than those associated with “Machine Learning” paired with “Interpretable” or “Explainable”, the data nonetheless indicate a growing interest in uniting surrogate modeling with interpretability and explainability. Publications combining “Surrogate Model” with either “Interpretable” or “Explainable” exhibited limited activity until around 2018, after which both trends began to accelerate noticeably. This shift reflects a promising and emerging focus on improving the transparency and interpretability of surrogate models, bridging the gap between efficiency and explainability in complex modeling tasks.

Building on these observed trends, \textcolor{black}{this review} positions itself at the intersection of surrogate modeling, simulation-based applications, and the growing demand for interpretability and explainability.  While prior work has often treated surrogate modeling and XAI as largely separate research domains, our study bridges the two by providing a state-of-the-art review of interpretable and explainable surrogate modeling for simulations, emphasizing their potential to support transparent and trustworthy decision-making. In addition to synthesizing the current literature, we offer perspectives on how XAI can enhance the interpretation of surrogate models, aiming to guide future research toward methods that combine high predictive performance with meaningful insights for scientific and engineering applications.

\color{black}
\section{ \textcolor{black}{Background and Foundational Concepts }}
\label{sec:engineering_design_exploration}

\textcolor{black}{
As established in the introduction, the simulation of complex systems requires navigating a landscape of continual trade-offs between fidelity, computational cost, and transparency. Before unpacking the specific explainability methods in Section~\ref{sec:XAI}, this section establishes the conceptual and computational baseline of our proposed workflow. We first detail the operational constraints of the dominant simulation paradigms, explain why these constraints inevitably drive the adoption of surrogate models, and formally resolve the semantic distinction between inherent interpretability and post-hoc explainability.}

Recall that a \textbf{complex system} consists of many interconnected and interdependent components whose nonlinear interactions generate emergent behaviors, self-organization, adaptation, and feedback loops that cannot be inferred from the properties of individual parts alone~\cite{kane2025ecosysml}. Modeling such systems inevitably forces practitioners to balance abstraction \textit{versus} fidelity, and tractability \textit{versus} expressiveness.

\subsection{\textcolor{black}{Equation-based} Simulations: Analysis and Exploration}
The goal of design exploration is not mere optimization but to better understand the relationship between design variables and the QoI~\cite{obayashi2005multi}. This procedure is important for creating improved designs that align with particular goals, limitations, and specifications. Hence, the expected result of this process is the accumulation of design knowledge and insights that will prove valuable for subsequent design endeavors. 

There is no universally accepted definition of design exploration since it spans various activities aiming to arrive at useful design knowledge. While acknowledging the significance of intuition, depending solely on intuition might lead to knowledge extraction that is less than optimal or significantly below optimal. The reason is that engineering intuition is limited by an individual's knowledge and experience, in the sense that it may not consider all potential design alternatives or explore complex multidimensional design spaces effectively. To that end, dedicated disciplines that provide a set of tools for a more comprehensive and efficient analysis exist for aiding engineers in the design process; this encompasses various techniques such as experimental design, statistical models, optimization methods, data analytics, ML, and more~\cite{chan2022trying,bussemaker2025system}.

An effective design exploration framework relies on a seamless integration of high-level stakeholder objectives, formal requirements, and executable system models. In this context, \emph{Goal-Oriented Requirements Engineering} (GORE) enables the capture and decomposition of stakeholder goals (“why?”) into operational constraints and measurable criteria (“what?”) that guide system functionality and quality, fostering dialogue among diverse experts around shared objectives~\cite{lapouchnian2005goal}. 

By articulating goals through GORE, engineers can construct explicit traceability links to downstream analytical models and metrics~\cite{chan2023stroke}.  
\emph{Model-Based Systems Engineering} (MBSE) further operationalizes this process by embedding requirements and goal decomposition within structured system models that span architecture, behavior, and verification domains. MBSE tools (\textit{e.g.}, SysML) and processes serve as a “single source of truth”, ensuring consistency, traceability, and automated propagation of requirement changes throughout the design hierarchy~\cite{dos2008model} between every specialized component.

Building on that foundation, MDAO provides the computational machinery needed to explore trade-spaces and evaluate design alternatives, rerunning analyses whenever requirements shift and interfacing specialist codes within a fixed-point loop for seamless co-simulation while integrating multi-physics simulations~\cite{bussemaker2022system}. When MDAO workflows are linked back to MBSE, requirement-derived goals drive the selection of simulation scenarios, performance metrics, and constraint checks. In practice, this means that whenever a requirement is updated, whether tightening a weight limit or adding a new functionality, the entire simulation and optimization loop can be reexecuted automatically, producing fresh insights into feasibility, sensitivity, and optimal trade-offs. 

Explainable surrogate models accelerate this loop, replacing costly simulations and, thanks to both global and local post-hoc explainers, reveal which inputs most influence outputs on different scales. This integration transforms simulations into interactive co-design tools, enabling scientists to rapidly explore “what-if” scenarios, validate choices against requirements, and iteratively refine designs, thereby delivering actionable insights and supporting human-in-the-loop decision-making~\cite{pinki2025symmetric}. Several frameworks of design exploration exist in the literature, each of which has its own definition. For example, the Multi-Objective Design Exploration (MODE) framework explores the multi-objective design space structure by extracting trade-off information and presenting it as a panoramic view for decision-makers~\cite{obayashi2010multi}. MODE was first coined in the context of aerospace design~\cite{obayashi2005multi} and has been successfully applied with various tools such as self-organizing maps~\cite{kohonen1982self}, rough sets~\cite{pawlak1982rough}, and Global Sensitivity Analysis (GSA). On the other hand, the design space exploration framework entails investigating various design options before their actual implementation~\cite{gries2004methods, pimentel2016exploring}. Regardless of the frameworks, we argue that all of these frameworks implement various scientific and mathematical methods to aid in obtaining essential design insight.

\subsection{Agent-Based Simulations: Analysis and Exploration}

ABMs provide a granular, bottom-up approach to simulating complex systems. By explicitly representing autonomous agents, their decision heuristics, and interaction rules within a spatial or networked environment, ABMs enable the study of emergent phenomena (\textit{e.g.}, traffic congestion, epidemic waves, market dynamics) arising from micro-level behaviors~\cite{macal2010tutorial,kaur2022multi,cueille2025assessing}. At its core, an Agent-Based Simulation (ABS) is built upon three primary components: \textit{agents} that act based on fixed or learning-based heuristics; an \textit{environment} that provides the spatial or topological context; and \textit{interaction mechanisms} that dictate how these entities engage with one another to drive the system's overall evolution~\cite{michel2018}.

\textcolor{black}{To study emergent behaviors at the mesoscopic or macroscopic levels, simulations of multi-agent systems provide an interactive view of how entities interact, offering a convenient way to test hypotheses and assess policy efficiencies in practice~\cite{drogoul2002multi}. A simulation is genuinely agent-based when computational entities retain autonomy, proactivity, and internal decision mechanisms~\cite{wooldridge1995intelligent}. Structuring these complex simulations typically relies on foundational conceptual frameworks. For instance, the ARD (Actor, Resource, Dynamics) paradigm and the standardized ODD (Overview, Design concepts, Details) protocol explicitly formalize autonomous actors, the environmental resources they utilize, and the rules governing their dynamics~\cite{taillandier2019, grimm2020odd}. 
Methodologically, the development of an ABS follows a sequential modeling lifecycle: it progresses from a domain model that captures stakeholder knowledge, to a conceptual model formalizing the system's mechanisms, and finally to an operational computational implementation~\cite{robinson2013conceptual, sargent2013introduction}. Within this operational model, the complexity of an agent's internal decision-making can vary drastically.}

To help experts navigate this complexity and validate model behaviors, subsequent work has explored participatory simulations in which humans control some of the computational agents within a shared environment~\cite{nguyen2007using,guyot2006agent}. In these hybrid platforms, domain specialists become "actors" that can control some of the computational agents~\cite{taillandier2019participatory}. Moreover, multi-agent simulations are generally stochastic because they incorporate probabilistic elements, particularly in the agents' decision-making processes.

ABS are also increasingly used in the context of systems-of-systems models interacting within an environment. In modern simulation co-design, for example, the product push-pull paradigm~\cite{naeem2024productpull} evaluates a concept by relying on a complex MDAO~\cite{bussemaker2024system} core for physical design, while taking into account its operational actionability through an engineered ABS business model~\cite{prakasha2024colossus}. These agent-based operations serve as a dynamic playground to test policies or concepts of operations and assess the performance of a system in a complex setting~\cite{villas2024concept}. 
Despite their fidelity, ABS often incur prohibitive computational costs and present vast, nonlinear parameter spaces that are difficult to explore exhaustively~\cite{angione2022}.

\subsection{Machine Learning  Surrogate Models}
\label{sec:surrogate_models}

\textcolor{black}{
Surrogate models (also called emulators or meta-models) are constructed to provide highly efficient approximations of complex, resource-intensive simulators~\cite{angione2022}. By training on a limited set of high-fidelity runs, they approximate the system's underlying input–output mapping at a fraction of the original computational cost, enabling rapid evaluation and dense design space exploration. }

\textcolor{black}{
Common surrogate architectures include Gaussian Processes (GPs) for uncertainty-aware emulation~\cite{rasmussen2006gaussian}, tree-based methods (\textit{e.g.}, Random Forests, XGBoost) for scalability~\cite{breiman2001random, friedman2001greedy}, and neural networks for high-dimensional spaces~\cite{lecun2015deep, frontiers.2022}. Regardless of the chosen method, every surrogate must navigate the fundamental trade-off between \textit{predictive fidelity} that is the ability to accurately replicate the complex system's dynamics, and \textit{computational efficiency} relative to the native simulation~\cite{arxiv.2502.06753}.}

Let us denote the design variables as $\boldsymbol{x}=[x_{1},x_{2},\ldots,x_{m}]^{T}$, where $m$ is the input dimensionality. The input variable $\boldsymbol{x}$ may consist of a combination of continuous, discrete, or categorical data types. Formally, we define the domain of each variable as follows:
\begin{enumerate}
    \item If $x_i$ is continuous, $x_i \in \mathbb{R}$.
    \item \textcolor{black}{If $x_i$ is discrete, $x_i \in \{a_{i1}, a_{i2}, \ldots, a_{ik_i}\}$, where $\{a_{i1}, a_{i2}, \ldots, a_{ik_i}\}$ are the $k_i$ admissible values for $x_i$.}
    \item \textcolor{black}{If $x_i$ is categorical, $x_i \in \{c_{i1}, c_{i2}, \ldots, c_{in_i}\}$, where $\{c_{i1}, c_{i2}, \ldots, c_{in_i}\}$ represent $n_i$ distinct categories for $x_i$.}
\end{enumerate}
Thus, the combined domain of $\boldsymbol{x}$, denoted as $\boldsymbol{\Omega}_{\boldsymbol{x}}$, is the Cartesian product of the individual domains of its components, expressed as $\boldsymbol{\Omega}_{\boldsymbol{x}} = \Omega_{x_1} \times \Omega_{x_2} \times \ldots \times \Omega_{x_m}$, where $\Omega_{x_i}$ is the domain of $x_i$.

The QoI, $y$, is related to $\boldsymbol{x}$ by a function $y=f(\boldsymbol{x})$. In practice, the expression of $f(\boldsymbol{x})$ is typically not known and needs to be estimated by a surrogate/ML model  $\hat{f}(\boldsymbol{x})$, \textit{i.e.}, $f(\boldsymbol{x}) \approx \hat{f}(\boldsymbol{x})$. 

The surrogate model $\hat{f}(\boldsymbol{x})$ is constructed using a set of sampling points seeded in the input space, \textit{i.e.}, the experimental design. Let us denote the experimental design as $\mathcal{X} = [\boldsymbol{x}^{(1)},\boldsymbol{x}^{(2)},\ldots,\boldsymbol{x}^{(n)}]^{T}$, where $n$ is the size of the experimental design. The corresponding response set is denoted as $\boldsymbol{y} = \{y^{(1)},y{(2)},\ldots,y^{(n)}\}^{T} = \{f(\boldsymbol{x})^{(1)},f(\boldsymbol{x}^{(2)}),\ldots,f(\boldsymbol{x})^{n}\}^{T}$. The pair of the experimental design and the responses is denoted as $\mathcal{D} = \{\mathcal{X},\boldsymbol{y}\}$. Let us also denote the set of surrogate/ML model parameters as $\boldsymbol{\Theta}$. The task is then to build an approximation model with $\boldsymbol{x}$ as the input variables, $\mathcal{D}$ as the experimental design, and $\boldsymbol{\Theta}$ as the model's parameters; hence, we have $\hat{f}(\boldsymbol{x}|\mathcal{D},\boldsymbol{\Theta})$. The parameter set $\boldsymbol{\Theta}$ usually needs to be tuned to create a model that is optimal according to a specific criterion (\textit{e.g.}, likelihood or loss function) and can refer to both model parameters (\textit{e.g.}, neural network weights) and hyperparameters (\textit{e.g.}, depth on neural nets)~\cite{saves2024}. We assume that one has the control to adjust the experimental design, which is typical in simulation-based workflows. For simplicity in notation, we denote the approximation model as $\hat{f}(\boldsymbol{x})$ by omitting the dependency on $\mathcal{D}$ and $\boldsymbol{\Theta}$. The model $\hat{f}(\boldsymbol{x})$ can then be used to aid computationally-expensive design exploration, \textit{e.g.}, where the evaluation of the objective function requires solving a high-fidelity Partial Differential Equation (PDE) solver. 

The use of surrogate models introduces \textit{trade-offs}: although they significantly reduce computational cost, they may reduce predictive accuracy, particularly when extrapolating beyond the training data. It is therefore necessary to rigorously validate these models to ensure their \textit{generalizability} in high-dimensional parameter spaces. Moreover, the \textit{interpretability} and \textit{explainability} of the models remain a major concern, especially in fields that require full transparency on the results or a high confidence in the models (the definitions of interpretability and explainability are discussed in Section~\ref{sec:what_is_explainability}).

\subsection{Selection Criteria for Surrogate Models}
There is a plethora of surrogate models that can be efficiently deployed for simulation-driven applications. In theory, any ML model can be used to capture the input-output relationship given the data. However, there are some desirable characteristics that a surrogate model should possess for effective applications. Given the central focus of \textcolor{black}{the present review} on explainability, we enumerate several significant attributes that are relevant within the framework of effective explainability. We listed some of these important considerations in the following:
\begin{enumerate}
    \item \textbf{Accuracy.} To properly extract trend and insight, it is important to have a high accuracy model on the first hand. Extracting trends from a flawed model would lead to misinterpreting the underlying relationships. \textcolor{black}{Thus, the model should neither underfit nor overfit the data}. Sometimes, the accuracy does not need to be too high, but at least sufficient for subsequent knowledge extraction. Nonetheless, it is always recommended to strive for higher accuracy.
    \item \textbf{Smoothness.} For models that are built from experiments or computer simulations that simulate physical phenomena, it is desirable for the models to be smooth. The model should not yield too spurious predictions so as to facilitate easier post-hoc explanations. Models like random forests typically generate 'steppy' responses, which are not ideal from a perspective of smoothness. However, there are some exceptions, and users should possess prior knowledge to judge if the smoothness assumption is unnecessary. For example, it has been demonstrated that random forest is better than GP and Kriging for a 13-dimensional antenna design problem~\cite{khan2021performance}.
    \item \textbf{Interpretability.} Some models are inherently interpretable, and this might ease knowledge extraction. In some cases, even interpretable models can generate a highly accurate approximation, negating the requirement for more intricate models that only marginally enhance accuracy. Explainability techniques can still be applied to such models if further information is necessary. Nonetheless, when the connection is intricate enough to demand the application of complex and opaque models, employing explainability techniques becomes necessary.
    \item \textbf{Fast prediction.} This trait is useful, especially for algorithms in explainability methods that require sampling. 
    Because one typically wishes to know the global underlying trend, the samples should be generated in the global input domain. Fast prediction allows explainability-enabled rapid exploration of the design space. Still, several explainability metrics can be analytically computed for some types of surrogate models~\cite{palar2023enhancing}. 
    
     \item \textbf{Uncertainty estimate.} Because a surrogate model is essentially an approximation, the prediction never exactly replicates the truth, and incorporating uncertainty information aids in evaluating the prediction's reliability. Note that the uncertainty defined here is an aleatoric uncertainty on the prediction, not referring to the specific context of "uncertainty quantification" in engineering~\cite{sudret2008global}. In addition to improving model assessment, uncertainty estimation can also be extended to calculate the uncertainty in the metrics used to assess explainability. Some models naturally provide uncertainty estimates (\textit{e.g.}, GP, random forest, relevance vector machine~\cite{tipping1999relevance}, and Bayesian neural network~\cite{kononenko1989bayesian}); otherwise, techniques such as bootstrapping~\cite{efron1994introduction} or conformal prediction~\cite{shafer2008tutorial} can be employed to derive the uncertainty of the model.
     
\end{enumerate}

The No Free Lunch (NFL) theorem, initially formulated in the context of optimization and search~\cite{wolpert2002no}, extends in a natural way to surrogate modeling and XAI~\cite{lattimore2013no}. Just as no optimizer outperforms all others across all possible functions, no single surrogate model or explanation technique can be universally optimal across all problem domains. Accuracy, interpretability, smoothness, uncertainty quantification, and computational efficiency represent competing objectives, and improving one often comes at the expense of another.
Table~\ref{tab:surrogate_tradeoffs_examples} summarizes these surrogate model attributes, their trade-offs under the NFL principle, and representative model examples under five representative criteria. 
\textcolor{black}{Note that this table is strictly illustrative. In practice, the computational efficiency of a model is highly dependent on its specific architecture, as well as the dimensionality and sample size of the dataset. Furthermore, while TabPFN~\cite{hollmann2025tabpfn} represents a powerful class of emerging foundation models, its integration into engineering surrogate modeling remains in its nascent stages. Finally, methods such as GPs and Support Vector Machines are fundamentally kernel-dependent; their classification within this taxonomy can shift dynamically based on the chosen kernel function.}

\begin{table*}[tbh!]
\centering
\renewcommand{\arraystretch}{1.1} 
\caption{Desirable attributes of surrogate models simulation-driven analysis, their trade-offs, 
and representative examples.}
\label{tab:surrogate_tradeoffs_examples}
\begin{tabular}{p{2.0cm} p{4.3cm} p{4.3cm} p{3.6cm}}
\toprule
\textbf{Criterion} & \textbf{Advantages (PROs)} & \textbf{Limitations (CONs)} & \textbf{Example of Surrogate Models} \\
\midrule
Prediction Accuracy & Captures underlying trends and relationships; enables reliable knowledge extraction & Risk of overfitting or underfitting; may require large datasets or careful hyperparameter tuning & Regular GP, random forest, TabPFN, deep ensemble \\
Smoothness & Facilitates physical consistency and stable post-hoc explanations & May fail to capture sharp discontinuities; some models yield stepwise responses & Smoothing spline, radial basis functions, polynomial chaos expansion \\
Interpretability & Direct insight into input–output relationships; supports decision-making without external XAI tools & Often comes with lower predictive accuracy compared to black-box models; may not scale to high dimensions & Linear regression, decision tree, generalized additive model \\
\textcolor{black}{Relatively} Fast Prediction  & Enables rapid evaluation for global explainability and sampling-based XAI methods  & High-accuracy models in high dimensions can be computationally expensive & $k$ nearest neighbors, inverse distance weighting, sparse GP, linear regression.   \\
Uncertainty Estimation & Provides confidence intervals for predictions; enhances trust and interpretability & Not all models natively support uncertainty; post-hoc methods (\textit{e.g.}, bootstrapping, conformal predictions) can be costly & Bayesian neural networks, Relevance vector machines, GP  \\
\bottomrule
\end{tabular}
\end{table*}

These trade-offs mirror more general challenges faced in supervised learning, where several fundamental limitations further complicate model selection and trustworthiness
that can impede its adoption, particularly by non-specialists. Four major limitations are:

\begin{enumerate}
  \item \textbf{Rashomon Effect.} Named after Kurosawa’s 1950 film, in which multiple witnesses offer conflicting accounts of the same event, this effect refers to the existence of many equally accurate but substantively different models for the same dataset \cite{anderson2016rashomon, wang2009machine, semenova2022existence}. Also, models trained on identical data can yield disparate decision rules and be evaluated very differently in the literature, underlining the subjective nature of model choice \cite{alanazi2017critical}.
  
  \item \textbf{Curse of Dimensionality (Dilution).} As the number of features grows, training points become increasingly sparse in the high‐dimensional space, making it harder to find predictive models that generalize well \cite{wang2009machine}. Consequently, only a small fraction of models will perform adequately, reinforcing the need for careful preprocessing and feature selection.
  
  \item \textbf{Hyperparameter Tuning.} Many algorithms require setting numerous hyperparameters (\textit{e.g.}, regularization strength, tree depth) before training \cite{probst2019tunability}. Users may rely on defaults, past experience, or exhaustive search. Automated strategies, such as grid/random search \cite{bergstra2012random} or Bayesian optimization \cite{snoek2012practical, shihua2025bayesian}, mitigate this burden but can incur substantial computational cost.
  
  \item \textbf{Black‐Box Models.} State‐of‐the‐art predictors (\textit{e.g.}, deep neural networks, ensemble methods) often lack interpretability, creating a “black‐box” barrier that hinders user trust and understanding \cite{goodman2017european, arrieta2020explainable}. This opacity is especially problematic in sensitive domains such as medicine, defense, and law. The rise of end‐to‐end Automated ML (AutoML) solutions exacerbates this issue by automating model configuration without enhancing transparency \cite{feurer2015efficient, he2021automl, dovsilovic2018explainable}.
\end{enumerate}

These limitations highlight the difficulty for a single user to select, configure, and trust a supervised learning model for their specific application. Selecting and properly tuning a supervised learning model is challenging for non‐experts due to phenomena such as the Rashomon effect, high dimensionality, hyperparameter complexity, and model opacity. 

\bigbreak
AutoML and meta-learning address these issues by recommending end-to-end \emph{workflows}, including data preprocessing, feature selection, model training, and hyperparameter optimization, based on past experiments. We distinguish three main meta-learning paradigms~\cite{vanschoren2019meta}:
\begin{itemize}
  \item \textbf{Evaluation-based.} Transfer successful workflows from similar past tasks using techniques like relative landmarks~\cite{furnkranz2001evaluation,leite2012selecting} or surrogate models~\cite{wistuba2015learning,feurer2018scalable}.
  \item \textbf{Model-based.} Initialize new models via transfer learning from pretrained predictors on related tasks~\cite{thrun1998learning,sharif2014cnn}.
  \item \textbf{Task-property-based.} Match new datasets to historical ones using meta-features (\textit{e.g.}, instance count, entropy, landmark performance) and apply the best workflow discovered for the closest task~\cite{raynaut2017towards,feurer2022auto}.
\end{itemize}
The task-property approach, as employed by Auto-sklearn~\cite{feurer2022auto} and OpenML~\cite{vanschoren2021towards}, is particularly promising, as it retains original workflows and automatically adapts their hyperparameters to new data. 

After presenting the main surrogate families and their trade-offs, a key question emerges: \emph{which surrogate should be chosen for a given simulator and analysis goal?} The answer depends on dataset size, noise, input type, simulator smoothness, and practical constraints such as time or interpretability. Exhaustive testing across models and hyperparameters is often inefficient, motivating automated, data-driven approaches.  

\begin{figure}[hbt!]
    \centering
    \hspace{-0.25cm}
        \includegraphics[width=1.1\columnwidth]{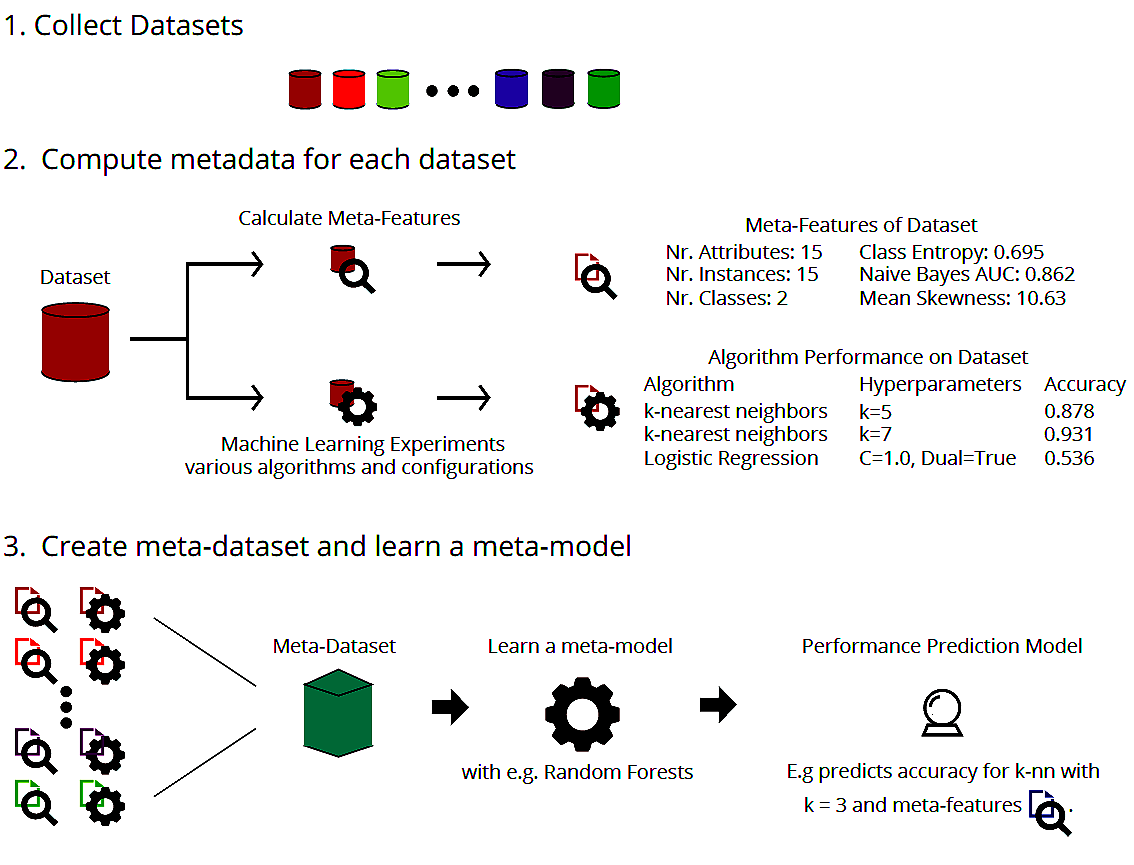}%
    \caption{An overview of the steps to create a meta-model (from~\cite{garouani2022towards1}).}
    \label{fig:metamodel}
\end{figure}

Meta-learning offers a robust solution by leveraging prior surrogate experiments to predict which model and settings are most suitable for a new task~\cite{garouani2022towards1,iceis23}. It builds a meta-model that links problem descriptors (meta-features) to expected surrogate performance. Once trained, this meta-model can guide model selection, warm-start tuning, and balance trade-offs between accuracy, speed, and interpretability, thereby reducing the cost of exploration and improving decision efficiency.
The procedure for constructing a meta-model is illustrated in Figure~\ref{fig:metamodel}~\cite{garouani2022towards1}. To develop such a model, a corresponding meta-dataset is required. This dataset consists of information gathered from various ML experiments. Each entry typically includes the algorithm used, the specific hyperparameter settings, the resulting performance metrics, and a description of the dataset on which the experiment was conducted. Dataset descriptions are represented through meta-features, which encapsulate statistical and structural properties of the data. These may include metrics such as the number of features, the number of classes, skewness, kurtosis, and other descriptive statistics. The initial phase in building a meta-dataset involves selecting a diverse collection of datasets for experimentation, as depicted in the first step of Figure~\ref{fig:metamodel}. In the subsequent phase, two tasks are carried out for each dataset. Firstly, meta-features must be computed. Secondly, ML experiments are conducted on each dataset. Finally, the results from both activities, meta-feature computation and ML experiments, are merged into a comprehensive meta-dataset~\cite{garouani2022using}.  

The meta-model is then trained on this dataset to learn the relationship between dataset characteristics (meta-features) and the performance of various learning algorithms. As a result, the quality and relevance of the meta-features directly impact the predictive capability of the meta-model~\cite{iceis23}. A well-trained meta-model can thus assist in recommending suitable algorithms for new, unseen datasets.
In summary, meta-learning is founded upon two core elements: the meta-data, which encapsulates previous learning experiences, and the meta-model, which generalizes from this knowledge to make informed predictions.

Many researchers also leverage surrogate models for uncertainty quantification, which aligns closely with XAI. In fact, understanding where a model is confident or uncertain directly supports transparency and trust in predictions. For instance, one can use model-agnostic uncertainty attribution methods, such as conformal prediction-based variable importance~\cite{idrissi2025unveil}, or model-specific techniques like the mean decrease in Gini impurity for a tree-based model such as a random forest~\cite{han2016variable}. The emerging paradigm of Automated XAI (AutoXAI) aims to unify these efforts, extending automated workflows to not only select the optimal surrogate model but also autonomously select, generate, and optimize the most appropriate explanations for a given engineering task~\cite{cugny2022autoxai}.

\subsection{Defining Explainability in Complex Systems}
\label{sec:what_is_explainability}

In the broader context of AI, an explanation serves as a means to verify and justify the decisions or outputs produced by an AI agent or algorithm~\cite{das2020opportunities}. In the pursuit of scientific insight, explainability is a prerequisite for achieving understanding, which ultimately enables the discovery of new knowledge~\cite{roscher2020explainable}. Explainability serves multiple purposes. However, it can be argued that there are at least two perspectives, \textcolor{black}{\textit{i.e.}, verification and interpretation}.  In this regard, explainability can be both retrospective (to justify decisions) and prospective (to reveal knowledge). When applied to surrogate models or ML models, explainability becomes especially relevant. A model may achieve high predictive accuracy, but this raises important questions: \textit{Why is the model so accurate?}, \textit{What underlying patterns is it capturing?} And more importantly (especially within science and engineering), \textit{what kind of knowledge can we extract from it?}. Answering these questions shifts the focus from mere performance to understanding, which is an essential step in scientific discovery and design exploration.

It is important to emphasize that the roles of explainability, validation, and uncertainty quantification are highly complementary but should not be confused. Explainability supports interpretation and diagnostic insight~\cite{amann2022explain}, translating the surrogate model into actionable, stakeholder-readable representations. Conversely, validation and uncertainty quantification provide the rigorous quantitative assessment of simulation fidelity, approximation error, and predictive reliability~\cite{roy2011comprehensive}. Robust workflows must combine surrogate uncertainty quantification, systematic replication, and XAI-driven diagnostics to ensure that surrogate-based explanations are reliable and to guide targeted high-fidelity reevaluation or active sampling when required~\cite{saves2024system,menz2020variance}.

To deepen this discussion, it is also important to distinguish between \textit{interpretability} and \textit{explainability}, as the two are often used interchangeably but may carry distinct meanings. \textcolor{black}{A model is considered interpretable when its mathematical form $\hat{f}(\boldsymbol{x})$ is transparent and when this transparency enables meaningful human understanding of the input–output relationships or system behavior (model-based interpretability~\cite{murdoch2019definitions}).} This facilitates straightforward analysis of the input-output relationship. Most interpretable models are parametric in nature and have a well-defined structure. \textcolor{black}{This idea is clearly illustrated in~\cite{ortigossa2024explainable} that use a simple linear regression model as an example}. Consider, for instance, a linear regression model defined over continuous variables (\textit{i.e.}, $\boldsymbol{x} \in \mathbb{R}^m$) is expressed as:
\begin{equation}
    \hat{f}(\boldsymbol{x}) = \beta_0 + \beta_1 x_1 +\beta_2 x_2 + \ldots \beta_m x_m,
\end{equation}
where $\beta_i, i=1,2,\ldots,m$ is the coefficient. This formulation makes it easy to understand the influence of each input variable: each input contributes linearly to the output, and the strength of its effect is directly reflected in the corresponding coefficient. 

A similar example is the quadratic model (\(    \hat{f}(\boldsymbol{x}) = \beta_{0} + \beta_{1} x_{1} + \beta_{2} x_{2} + \beta_{3} x_{1}^{2} + \beta_{4} x_{2}^{2} + \beta_{5} x_{1} x_{2}
\)). 
In this respect, one can easily understand the impact of the input variables on the output by looking at the polynomial terms and the corresponding coefficients. Further, if one wishes to assess the relative impact of the input, the input domain can be rescaled to the same range first before building the model. However, the more rigid structure of interpretable models constrains their flexibility in capturing complex relationships.

Notwithstanding, not all interpretable models are parametric, \textit{e.g.}, a decision tree is considered interpretable since its structure consists of simple, sequential if-then rules~\cite{kim2007visualizable}.  These models are referred to as a "transparent" model owing to their inherently interpretable characteristics.  However, an ensemble of trees, such as boosting and random forest, loses its transparency due to the complexity caused by the combination of a large number of decision trees~\cite{yang2024inherently}. Besides these examples, in general, interpretable models can be classified into several types, including rule-based models (\textit{e.g.}, RuleFit~\cite{friedman2008predictive}), score-based models (\textit{e.g.}, super sparse linear integer model~\cite{ustun2016supersparse}), generalized-additive models~\cite{hastie1986generalized}, symbolic regression (\textit{e.g.}, lasso regression), self-explaining neural networks~\cite{alvarez2018towards}, and so on~\cite {marcinkevivcs2023interpretable}.

Henceforth, in this \textcolor{black}{review}, we adopt the following definitions of interpretability and explainability~\cite{XAI_IAI}:

\textit{Explainability}, is the ability to make the decision-making processes of ML models transparent and understandable to humans, often through techniques (such as visualization) that expose how inputs are processed and which features or internal representations contribute to the final prediction.

\textit{Interpretability}, concerns the extent to which a model’s internal mechanisms and features are transparent, comprehensible, and meaningfully connected to real-world concepts~\cite{garouani2022towards}.

\textcolor{black}{Notably, mathematical simplicity alone does not strictly guarantee interpretability, particularly when approximating complex physical simulations. Even if an inherently transparent surrogate model is chosen, it may capture highly non-linear or chaotic input-output mappings. This challenge is conceptually mirrored in canonical dynamical models like the Lorenz system: while governed by simple, fully transparent equations, the resulting dynamics are profoundly complex and counterintuitive to human reasoning. In such surrogate scenarios, the model's structural form may be readable, yet its predictive behavior remains opaque. This highlights a critical nuance: true interpretability depends not only on having a transparent mathematical form but also on the comprehensibility of the model's resulting behavior.}

\textcolor{black}{Because of this nuance, whether a model is explainable highly depends on the adopted definition of explainability. In this review, a weak notion of explainability refers to post-hoc analyses that can reveal aspects of the model's input-output relationships, even if such explanations are approximate. Under stronger notions that require truly human-meaningful explanations, explainability may not always be achievable~\cite{yampolskiy2019unexplainability}. Regardless, under the weak notion, explanations can often be generated for a wide range of models; however, their fidelity, robustness, and faithfulness to the underlying mechanisms must always be carefully assessed, particularly for highly complex models~\cite{ali2023explainable}.}

\textcolor{black}{A related and commonly adopted strategy in XAI is to approximate a complex, hard-to-interpret model using a simpler and more interpretable ``surrogate model '' acting as a proxy~\cite{wilhelm2024hacking, zhu2022fuzzy}. While this approach is intuitive, it introduces important limitations. In particular, when the original model exhibits strong nonlinearity or high complexity, a simple surrogate may fail to faithfully approximate its behavior, potentially leading to oversimplification or misrepresentation.}

Another related term is \textit{decomposability}, which refers to a model’s ability to express its output as a sum (or composition) of contributions from individual input variables or their interactions~\cite{puthanveettil2025review}. This property enables clear interpretation of how each input affects the output, making the model more transparent and analyzable~\cite{lipton2018mythos}. However, decomposability is not exactly the same as interpretability. For instance, while kernel-based models are decomposable to some extent, the resulting components are often difficult to interpret. 

\textcolor{black}{The surrogate model $\hat{f}(\boldsymbol{x})$ is often chosen from more flexible classes, such as GP models and neural networks, rather than simple parametric forms. In this regard, the non-parametric or high flexibility characteristics of such a model create difficulty in understanding the model and hinder meaningful analysis without further post-processing.} To that end, modern tools from statistics and explainable ML should be employed to reveal more insightful analyses to users.
A wide range of explainability techniques now exist to help uncover the underlying mechanisms and feature relationships learned by the model. Importantly, “explainability” is not the exclusive domain of the ML community. Many established methods from statistics and applied mathematics 
contribute directly to understanding input–output relationships and thus fall naturally under the XAI umbrella when applied to surrogate or simulator outputs. Recognising this disciplinary overlap broadens the methodological toolkit and encourages more robust, interpretable/explainable workflows~\cite{idrissi2024development}.
In this \textcolor{black}{review}, we acknowledge that methods used to "explain" a model do not necessarily originate solely from the ML community. To a certain extent, we consider that techniques from other disciplines (such as statistics and applied mathematics) also fall under the umbrella of explainability, particularly when they contribute to understanding model behavior or providing insights into input–output relationships.

Another important distinction lies between model-specific and model-agnostic explainability methods~\cite{marcinkevivcs2023interpretable, danesh2023hybridization, doshi2017towards}. As the names imply, model-specific methods are designed for particular types of models. For instance, the uncertainty estimates provided by GP are inherent to its probabilistic structure~\cite{rasmussen2006gaussian}. In contrast, model-agnostic methods are more flexible and can be applied to a wide range of ML models, regardless of their internal architecture~\cite{gawde2024explainable}.

Explainability itself is subject to an NFL limitation~\cite{holzinger2018machine}: the appropriateness and informativeness of a given XAI method depend strongly on the model class, the structure of the data, and the end-user objectives. This perspective highlights the necessity of adopting adaptive, context-aware strategies, where multiple surrogate models and complementary explainability methods are jointly deployed to capture different facets of complex system behavior. For instance, a model that offers high interpretability may lack strong predictive accuracy (reflecting the classic accuracy–interpretability trade-off). Therefore, it is essential to assess which model characteristics are most important for a given application~\cite{cugny2022autoxai}.

\subsection{Local and Global Explainability}
\label{sec:local_global_explain}
\color{black}
Global and local explainability refer to two complementary levels of understanding an ML or surrogate model's behavior. 

Global explainability reveals system-wide trends by identifying which features are most influential across the entire dataset \textcolor{black}{(\textit{i.e.}, global behavior~\cite{lundberg2020local})}. It is important for early design decisions, such as parameter screening or model simplification. Classical GSA metric, provide strictly global explanations by quantifying the overall impact of inputs on output variance.

In contrast, local explainability focuses on individual predictions, detailing why a specific decision was made for a particular input \textcolor{black}{rather than across the entire parameter space~\cite{ribeiro2016should, zhang2021survey}}. \textcolor{black}{In an engineering context, for instance, local explainability clarifies why a given design condition leads to a specific predicted outcome (such as stress, drag, or aerodynamic efficiency) by isolating the strongest input drivers for that exact realization.}

Importantly, several modern methods offer both simultaneously as most of the local approaches can be globally averaged to give insight into the general behavior~\cite{lundberg2020local}. 
This allows practitioners to establish a global understanding before "zooming in" on specific instances. 

To effectively deploy these metrics, practitioners must navigate the complementarity between \textbf{model-based} (inherently interpretable) and \textbf{data-driven} (post-hoc surrogate) analyses. 
%
%
\textbf{Model-based screening} applies simple and inherently interpretable techniques directly to raw simulation data. While it provides a rapid global assessment to identify primary drivers and theoretical limits of intrinsic stochasticity, it often misses complex, non-additive interactions. \textbf{Data-driven surrogate analysis} fills this gap by training expressive ML models on the refined parameter space. This enables the deployment of advanced post-hoc XAI tools to map precise tipping points and phase transitions. Essentially, model-based methods act as a coarse filter to bound the problem, while data-driven XAI provides the high-fidelity, localized insights necessary to drive robust decision-making.


\subsection{Towards XAI-Driven Simulation Exploration}

As established, computational simulation remains indispensable across both engineering design and agent-based modeling for predicting performance, assessing trade-offs, and analyzing emergent behaviors. However, the integration of XAI into these workflows marks a critical paradigm shift in how complex systems are explored and optimized. By coupling high-fidelity surrogates with XAI, practitioners convert opaque, black-box \textcolor{black}{surrogate models} into actionable instruments that preserve computational efficiency while actively supporting human-centered decision-making.
Most importantly, this integration relies on a complementary pipeline where model-based screening acts as a rapid, coarse filter to identify primary drivers, enabling data-driven surrogate analysis to subsequently map complex, non-linear interactions and localized tipping points with high fidelity~\cite{saves2025XAIApp}. 

Realizing this potential also requires exploiting the specific strengths of different surrogate architectures. For example, ensemble approaches are particularly well-suited for ABS because they can be integrated at both the macroscopic system level and the microscopic agent level~\cite{blanco-volle2024}. The combined deployment of these models with global and local explainability tools provides a comprehensive dual perspective: a broad view of system-wide trends paired with a detailed analysis of local interactions~\cite{iooss2022}.

The diverse use cases found in engineering design and goal-oriented exploration should therefore not be viewed as isolated analytical tasks, but rather as the operational objectives of a unified, XAI-driven pipeline. To operationalize this pipeline, the following section discusses several commonly used explainability methods, highlighting their core principles and their specific suitability for surrogate-based simulation workflows. Also, \textcolor{black}{visual} representation plays an important role in supporting explainability within this framework. By transforming abstract numerical relationships into intuitive diagrams, visualization facilitates rigorous model refinement and validates consistency against physical principles for data scientists~\cite{riis2024explainable}.
\color{black}

\color{black}
\section{The Explainability Toolbox: A Taxonomy of Methods}
\label{sec:XAI}

While ML surrogate models efficiently accelerate complex systems simulations, they often act as opaque black boxes where predictive accuracy alone is insufficient for true understanding. To extract actionable knowledge, reveal physical mechanisms, and ensure trustworthiness, practitioners must deploy specialized explainability techniques~\cite{XAI_IAI}.
Moving beyond passive post-hoc artifacts, explanations are increasingly leveraged as interactive dialogue interfaces. This allows decision-makers to validate, refine, or contest outputs, closing a feedback loop that improves model reliability and prevents over-reliance~\cite{ferreira2020people, amershi2019guidelines}. Furthermore, recent advances leverage explanation vectors as operational inputs for automated, auditable recommendation pipelines in high-stakes, human-in-the-loop applications~\cite{vouros2022explainable, hall2022still}.

Rather than presenting these tools as a disconnected mathematical taxonomy, we synthesize them into a problem-driven framework, categorizing XAI techniques directly by their operational usage within the engineering lifecycle. Specifically, we progress from global sensitivity metrics used for initial parameter screening  and tools for mapping non-linear, multi-objective trade-offs, to local attribution methods required for diagnosing specific instance anomalies. We then explore methods for formal logical verification, strategies to bridge the cultural gap with traditional data mining, and techniques for enhancing human comprehensibility during deployment. Finally, we highlight a paradigm shift where explanations themselves serve as a new design space for automated meta-analysis. 
By mapping these tools directly to their deployment constraints, we demonstrate how they collectively transform opaque emulators into transparent, actionable decision-making instruments.

This section provides a brief overview of some explainability methods that can be employed to explain the surrogate model. In essence, these methods aim to probe the internal complexity of the surrogate model by applying post-hoc techniques. Such approaches can reveal various aspects of the model’s behavior, including (but not limited to): (1) the influence of subsets of input variables on output, (2) the effects of interactions among input variables on model predictions, and (3) regions within the input space where the surrogate model exhibits pronounced nonlinearity or increased sensitivity~\cite{hirschler2021new}. \textcolor{black}{The list below is not exhaustive, but rather highlights several representative approaches that are particularly relevant in the context of surrogate modeling for scientific and engineering applications.}

\textcolor{black}{Given the broad scope of methods covered in this review, it is not feasible to provide detailed mathematical formulations for every approach. Instead, we present representative formulations and key underlying concepts to ensure sufficient technical grounding while preserving clarity and readability. Readers are referred to the original references for more comprehensive mathematical treatments of specific methods.}

\subsection{ Global Parameter Screening}
\label{sec:global_parameter_screening}
High-dimensional input spaces introduce the "curse of dimensionality," making surrogate training computationally expensive and obscuring underlying physical mechanisms~\cite{hou2022dimensionality}. The foundational step in an XAI pipeline is therefore to filter out negligible variables and identify the most influential ones. Rather than mere data pre-processing, these techniques form the first layer of global explainability. By quantifying input influence and uncovering intrinsic low-dimensional structures, they map the model's overarching behavior before moving to local diagnostics. Even in small- to medium-dimensional settings, global explainability remains important, as it provides global insight into variable influence and interaction effects. The following subsections detail two key methodological approaches that can be used for this task: sensitivity analysis and Active Subspace Methods (ASM).

\subsubsection{Derivative and Variance-Based Global Sensitivity Analysis}
\color{black}
GSA aims to measure how significantly input variables (individually and through interactions) influence the output~\cite{Iooss_2015}. Various GSA methods have been developed across different research fields. In this section, we focus on a subset of methods to illustrate the general idea of GSA.

Before introducing fully global measures, it is instructive to briefly revisit local sensitivity analysis. Derivative-based local sensitivity analysis studies the impact of infinitesimal perturbations to individual inputs around a nominal operating point $\boldsymbol{x}$~\cite{sudret2008global}. As a preliminary screening technique, it has a low computational cost and highlights potentially influential factors prior to more elaborate global sensitivity methods.
Given a function $f:\mathbb{R}^m\to\mathbb{R}$ evaluated at $\boldsymbol{x}$, the local sensitivity of an input $x_i$ is $\partial f(\boldsymbol{x})/\partial x_i$. The local sensitivities with respect to all inputs are represented by the gradient vector,
$\nabla f(\boldsymbol{x}) = \left[ \frac{\partial f(\boldsymbol{x})}{\partial x_1}, \frac{\partial f(\boldsymbol{x})}{\partial x_2}, \dots, \frac{\partial f(\boldsymbol{x})}{\partial x_m} \right]^\top$. 
\textcolor{black}{When analytic derivatives are not available (which is often the case in practice), the required sensitivities can be computed using numerical techniques, particularly adjoint methods~\cite{lions1971optimal, giles2000introduction} and automatic differentiation~\cite{griewank2008evaluating}, which are especially efficient for PDE-based simulators.} 

GSA methods are classified into four categories in~\cite{van2022comparison}: (1) variance-based methods, (2) derivative-based methods, (3) density-based methods, and (4) model-based methods. \textcolor{black}{It should be noted that this classification is not exhaustive; for instance, Pearson correlation-based measures and partial correlation coefficient are regarded as methods that are based on linear models~\cite{Iooss_2015}. In addition, moment-independent importance measures form another important class of sensitivity analysis methods, focusing on identifying the input variable whose specification would result in the largest expected alteration of the output distribution~\cite{auder2008global, borgonovo2008moment, borgonovo2011moment}.} Variance-based techniques (\textit{e.g.}, Sobol' indices~\cite{sobol1993sensitivity} and the Fourier Amplitude Sensitivity Test~\cite{cukier1973study}) rely on decomposing the output variance into contributions from individual inputs and their interactions. Derivative-based methods aggregate derivative (gradient) information over the input domain. Density-based approaches examine the full output probability density function to measure how perturbations in inputs reshape the output distribution. Model-based methods extract sensitivity information from an approximation (surrogate) of the original black-box function; here, "models" need not be ML models and may be empirical. All four classes can be applied to surrogate models, which act as \textcolor{black}{less expensive} proxies for costly simulations. 

\textcolor{black}{One of the earliest GSA techniques is the Morris method~\cite{morris1991factorial}.} Let $e_i$ be the basis vector of the feature $x_i$ (the only that changes at the time), then the \textcolor{black}{global sensitivity} is given by the Morris method where, at each sample  $\boldsymbol{x}$, the elementary effect for variable $i$ is:
\[
EE_i\bigl(\boldsymbol{x}\bigr) = \frac{f\bigl(\boldsymbol{x} + \delta\,\mathbf{e}_i\bigr) - f\bigl(\boldsymbol{x}\bigr)}{\delta},\quad \delta\ll1.
\]
These methods work well when $f$ is sufficiently smooth near the nominal point, but may fail for highly nonlinear models. Perturbations can be performed multiple times ($n \approx 5)$  across the normalized input domain to obtain the two measures $\mu_i$ and $\sigma_i$ defined as the mean and the standard deviation of the distribution of the elementary effects of each input~\cite{saltelli2004sensitivity,Iooss_2015}:

\begin{center}
$
\begin{cases}    
\mu_i = \frac{1}{n}\sum_{j=1}^n \bigl| EE_i(\boldsymbol{x}^{j}) \bigr|, \\

\sigma_i = \sqrt{\frac{1}{n} \sum\bigl(EE_i-  \frac{1}{n}\sum_{j=1}^n EE_i(\boldsymbol{x}^{j})  \bigr)^2}.
\end{cases}
$
\end{center}
Here $\mu_i$ represents the average magnitude by which perturbations in input $i$ shift the model output, providing a measure of its overall influence, while $\sigma_i$ captures the variability of those elementary effects across the input space, highlighting the presence and strength of nonlinear behaviors or interactions with other inputs.

Variance-based metrics are arguably the most widely used methods in GSA. They quantify the contribution of each input variable, or combination of variables, by decomposing the total output variance into partial contributions. Let $f(\boldsymbol{x})$ be a square-integrable function defined over the domain $\Omega = \mathbb{R}^m$. The total variance of the model output is defined as:
\begin{equation}
    \mathrm{Var}(f) = \mathbb{E}\left[(f(\boldsymbol{x}) - \mathbb{E}[f(\boldsymbol{x})])^2\right].
\end{equation}

Variance-based methods, most notably Sobol' indices~\cite{sobol1993sensitivity, sobol2001global}, rely fundamentally on the functional Analysis of Variance (ANOVA) decomposition. This framework breaks down a highly complex, multi-dimensional surrogate model $f(\boldsymbol{x})$ into a sum of orthogonal component functions:
\begin{equation}
\begin{split}
    f(\boldsymbol{x}) =& f_0 + \sum_{i=1}^m f_i(x_i) + \sum_{1 \le i < j \le m} f_{ij}(x_i, x_j) \\  
 &    +\dots + f_{1, 2, \dots, m}(\boldsymbol{x}),
\end{split}
    \end{equation}
where $f_0 = \mathbb{E}[f(\boldsymbol{x})]$. Assuming the input variables are independent, this orthogonality allows the total variance to be perfectly partitioned into partial variances:
\begin{equation}
    \mathrm{Var}(f) = \sum_{i=1}^m V_i + \sum_{1 \le i < j \le m} V_{ij} + \dots + V_{1, 2, \dots, m},
\end{equation}
where $V_i = \mathrm{Var}(f_i(x_i))$, $V_{ij} = \mathrm{Var}(f_{ij}(x_i, x_j))$, and more generally $V_A = \mathrm{Var}(f_A(\boldsymbol{x}_A))$ ($A \subseteq \{1, \dots, m\}$) denote the partial variances associated with subsets of input variables.
By evaluating these partial variances, we can derive the Sobol' indices, which quantify the exact fraction of the total variance attributable to any specific input or input interaction.

For a specific subset of variables denoted by the index set $A$, the corresponding Sobol' index is defined as:
\begin{equation}
    \label{eq:Sobol}
    S_A = \frac{V_A}{\mathrm{Var}(f)}.
\end{equation}
When $|A|=1$, $S_A$ corresponds to the \textit{first-order} Sobol' index (capturing the isolated main effect of a single variable), while $|A|=2$ yields a \textit{second-order} index (capturing two-way interactions), and so forth. 

To fully understand a variable's overall influence, it is common to calculate the \textit{total} Sobol' index~\cite{homma1996importance}. For a given variable $i$, the total index $S_i^T$ sums all partial Sobol' indices that involve $x_i$, including all of its higher-order interactions:
\textcolor{black}{\begin{equation}
    S_i^T = \sum_{A \subseteq \{1,\dots,m\} \mid i \in A} S_A.
\end{equation}}
For example, in a three-variable system ($m=3$), the total index for the first variable is $S_1^T = S_{\{1\}} + S_{\{1,2\}} + S_{\{1,3\}} + S_{\{1,2,3\}}$. By definition, $\sum_{i=1}^m S_i^T \ge 1$, where equality holds only if the model is purely additive and lacks any variable interactions.

When inputs are deterministic (\textit{e.g.}, a parameter sweep in engineering design), we can still apply Sobol' indices by treating those inputs as random variables (typically following a joint uniform distribution across the design space). In this regard, Sobol' indices serve as powerful methods to globally "explain" the surrogate model. However, while they excel at ranking variable importance, they do not inherently offer ways to visualize the directional impact of these inputs on the output, motivating the use of complementary visual XAI methods.

\textcolor{black}{An alternative method for GSA is Derivative-Based Global Sensitivity Measures (DGSM), which extends the concept of local differentiation across the full input distribution~\cite{sobol2010derivative}:
\begin{equation}
    \nu_i = \mathbb{E} \left[ \left( \frac{\partial f(\boldsymbol{x})}{\partial x_i} \right)^2 \right] = \int_{\Omega} \left( \frac{\partial f(\boldsymbol{x})}{\partial x_i} \right)^2 \rho(\boldsymbol{x}) \, d\boldsymbol{x},
\end{equation}
where $\Omega$ is the input space and $\rho(\boldsymbol{x})$ is the joint probability density function of the inputs.}
\textcolor{black}{These measures are efficiently estimated via Monte Carlo or randomized Quasi‑Monte Carlo techniques or through derivative-based surrogate emulators~\cite{bouhlel2019python,de2016estimation}. 
Under appropriate smoothness and distributional assumptions, DGSM provides upper bounds on Sobol’ total-effect indices and therefore offers a scalable alternative to full Sobol' analysis while retaining information about interactions~\cite{kucherenko2016derivative,boucharif2025importance}. Although DGSM relies on local derivatives, it is classified as a form of GSA  because the derivatives are evaluated and aggregated over the entire input space, thus capturing global effects rather than pointwise sensitivities.}

It is important to note that global sensitivity metrics can also be obtained from ML-based explainability approaches. As discussed in the following sections, methods such as \textbf{Partial Dependence Plot} (PDP)~\cite{breiman2001random}, and \textbf{Shapley Additive Explanations}(SHAP)~\cite{lundberg2017unified} also offer ways to assess global feature importance. Another GSA metric that has its origins in the ML community is permutation feature importance~\cite{breiman2001random, fisher2019all}. It is based on an intuitive idea: a feature is considered important if randomly shuffling its values leads to a significant increase in the model’s prediction error. On the other hand, inputs that do not bear significant importance will only slightly change the error when they are reshuffled.

\subsubsection{Active Subspace Methods}
\label{sec:active_subspace_method}
The ASM~\cite{constantine2014active} identifies directions in the input space along which the model response exhibits the most significant variability. Their mathematical machinery relies on the covariance matrix of the output gradient with respect to the input variables, \textit{i.e.}, $\nabla f (\boldsymbol{x})$. Formally, the following expectation is  computed:
\begin{equation}
    \mathbf{C} = \mathbb{E}[\nabla f (\boldsymbol{x}) \nabla f (\boldsymbol{x})^{T}],
\end{equation}
in which eigenvalue decomposition can be performed to yield
\begin{equation}
     \mathbf{C} = \mathbf{W} \boldsymbol{\Lambda} \mathbf{W}^{T},
\end{equation}
where $\mathbf{W} = [\mathbf{w}_{1},\mathbf{w}_{2},\ldots,\mathbf{w}_{m}]$ is the matrix of Eigenvectors and $\boldsymbol{\Lambda}=\text{diag}(\lambda_{1},\lambda_{2},\ldots,\lambda_{m})$ is the diagonal matrix of Eigenvalues and $\lambda_{1} \geq \lambda_{2} \geq \ldots \geq \lambda_{m}$. In this regard, the first eigenvector accounts for the highest variability, followed by the second, continuing in order down to the last eigenvector. Projections onto the first and second eigenvectors can then help visualize the most active directions, especially if both are significantly more important than the rest. To be exact, if we define the truncated eigenvector matrix $\mathbf{W}_{a} = [\mathbf{w}_{1},\mathbf{w}_{2}, \ldots, \mathbf{w}_{a}]$, where $a < m$ (usually $a=1$ or $a=2$ for visualization), then we can define the active variables $\boldsymbol{x}_{a}$ as $\boldsymbol{x}_{a} = \mathbf{W}_{a}^{T}\boldsymbol{x}$.

\begin{figure}[tbh]
    \hspace{-0.1cm}
 	\begin{subfigure}{.5\columnwidth}
        \hspace{-0.5cm}
		\includegraphics[height=4.25cm, width=1.1\columnwidth]{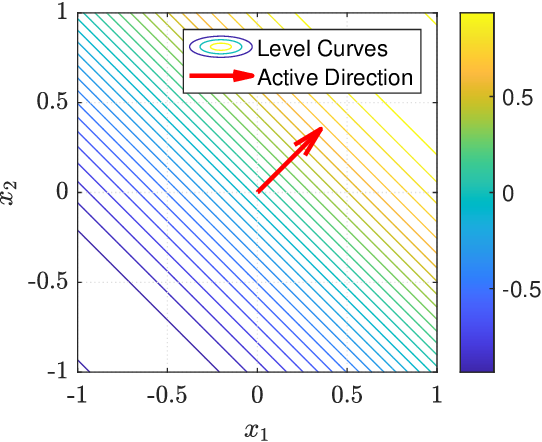}%
		\caption{The contour plot\\ of $f(x_1,x_2)=\sin(x_1+x_2)$}%
		\label{fig:ASM_2Dfunc}
	\end{subfigure}
  	\begin{subfigure}{.5\columnwidth}
      \hspace{-1.01cm}
		\includegraphics[height=4.25cm,width=1.25\columnwidth]{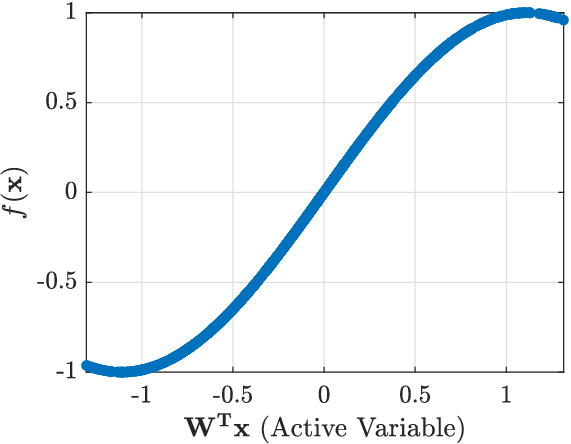}%
		\caption{One-dimensional active subspace}%
		\label{fig:ASM_2Dfunc_1_av}
	\end{subfigure}\hfill%
    \caption{An example of ASM applied to the function of $f(x_1,x_2)=\sin(x_1+x_2)$.}
	\label{fig:ASM_exam}
\end{figure}

Additionally, the ASM also provides a global sensitivity metric called activity score~\cite{constantine2017global}, which reads as:
\begin{equation}
    \alpha_i = \alpha_i(m) = \sum_{j=1}^{m} \lambda_j w_{i,j}^{2}, i=1,\ldots,m,
\end{equation}
The activity score essentially gives information on how much the $i$-th variable contributes to the most influential directions (\textit{i.e.}, the active subspace). 

Although the ASM is not typically labelled as an explainability method, it provides explainability by enabling users to explore the complexity of a problem. This is achieved by identifying the most "active" subspace that captures the greatest variability. Notably, ASM itself does not rely on any predictive model, as it operates based on gradient information. When applied to a differentiable surrogate model such as GP, ASM can expose the underlying complexity to the user through a Visualization of the latent space. This way, one can infer information such as multimodality, nonlinearity, and the actual complexity of the problem. Closed-form solutions for active subspace are also available for models such as GP~\cite{wycoff2021sequential}. It is worth noting that ASM is only applicable to all continuous variables because it relies on the gradient of the output function with respect to input variables~\cite{constantine2017global}. \textcolor{black}{Regardless, the conventional ASM is formulated based on the standard gradient of the model output with respect to the inputs and therefore is most appropriately applied under the assumption of independent input variables.}

For demonstration, consider the function $f(x_1, x_2) = \sin(x_1 + x_2) $ defined over the domain $\boldsymbol{\Omega} = [-1,1]^2$ (see Fig.~\ref{fig:ASM_exam}). Although the function depends on two variables, as illustrated in Fig.~\ref{fig:ASM_2Dfunc}, it predominantly varies along a single direction. This reveals the presence of a one-dimensional active subspace, characterized by the vector $\mathbf{w}_1 = [0.7071,\ 0.7071]^T$, with a corresponding eigenvalue of $\lambda_1 = 1.1905$, as shown in Fig.~\ref{fig:ASM_2Dfunc_1_av}.

\textcolor{black}{A common way to connect active subspaces with variance-based global sensitivity analysis and XAI is via \emph{Poincaré inequalities} in an $\mathcal{L}^2$ setting~\cite{zahm2020gradient}. Establishing this link is crucial: it proves that ASM is not merely a geometric heuristic for dimension reduction, but a formally guaranteed method for capturing the global variability of a model. This justifies its use as a robust explainability tool in high-stakes engineering.}

\textcolor{black}{Formally, let $\boldsymbol{x} \in \mathbb{R}^m$ be a random input vector and $f(\boldsymbol{x})$ the quantity of interest. If we approximate the full model $f$ with an optimal lower-dimensional "ridge" function $\hat{g}$ that depends only on the candidate active subspace directions (defined by the projection matrix $\boldsymbol{W}_a \in \mathbb{R}^{m \times a}$), the mean-square approximation error can be bounded using a Poincaré constant $C_P$:}
\textcolor{black}{
\begin{equation}\label{eq:spectral_bound}
    \mathbb{E}\left[\left(f(\boldsymbol{x}) - \hat{g}(\boldsymbol{W}_a^\top \boldsymbol{x})\right)^2\right] \le C_P \sum_{j>a} \lambda_j.
\end{equation}}

\textcolor{black}{This elegant spectral bound provides a concise and powerful interpretation:}
\textcolor{black}{
\begin{itemize}
    \item The dominant eigenvectors of the expected outer product of the gradients indicate the directions of greatest average sensitivity, forming natural coordinates for dimension reduction.
    \item The sum of the discarded (inactive) eigenvalues directly bounds the residual variance, meaning the loss of explainability is strictly controlled and quantifiable.
\end{itemize}}

\textcolor{black}{By explicitly bounding the variance error, this formulation bridges gradient-based ASM with traditional GSA. It allows practitioners to confidently project high-dimensional black-box models into transparent, low-dimensional representations without discarding parameter combinations that significantly drive the output.}

In Bayesian inversion, applying these bounds to gradients of the log-likelihood identifies data-informed parameter directions and provides $L^2$-type control on predictive or moment errors of ridge surrogates in the posterior regime. 
In~\cite{bibal2020interpretability}, it is shown that the ASM efficiently uncovers the few input directions most responsible for output variability, enabling dramatic dimensionality reduction with minimal loss of accuracy. This work extends ASM beyond smooth, diffusion‑based models by incorporating kernel approximations and local explainability tools, broadening its applicability to complex, non‑differentiable systems. 
\textcolor{black}{
Recent advances have significantly expanded the ASM across three main fronts. First, probabilistic frameworks have been developed to quantify directional uncertainty and discover subspaces in high-dimensional models~\cite{rumsey2024discovering}, extending even into uncertainty quantification for deep learning~\cite{jantre2024learning}. Second, researchers are achieving built-in interpretability and feature discovery by embedding active subspaces directly into neural network architectures, including Gaussian radial basis functions~\cite{d2024learning} and Kolmogorov-Arnold Networks~\cite{zhou2025askan}. Finally, to overcome computational bottlenecks in high-dimensional scaling, the field has increasingly adopted randomized algorithms and orthogonalization techniques~\cite{breaz2024randomized,de2025randomized}. These developments solidify ASM as a highly valuable tool for both model simplification and global behavior analysis in complex engineering.}

\bigbreak
As a side note, explainable ML can also be applied to streamline simulation-driven tasks, such as optimization and uncertainty analysis, by offering methods to reduce the dimensionality of the problem besides ASM. For instance, dimensionality reduction techniques in ML, like PCA and t-SNE~\cite{van2008visualizing}, typically do not fall under the category of explainable ML. However, they can provide valuable insights into the problem at hand by discovering essential design knowledge. Thus, if the goal is to explain the problem at hand, then these methods also fall within the broader term of explainability through dimension reduction alongside ASM.

\subsection{\textcolor{black}{Visualization of Non-Linear Interactions}}
\textcolor{black}{
While global parameter screening methods like Sobol' indices and DGSM~\cite{kucherenko2009monte} identify \emph{which} features drive model variance, they produce scalar scores that obscure \emph{how} these features shape the response surface. In complex systems, variables rarely act in isolation; their effects are highly non-linear and coupled. Understanding these mechanisms requires moving beyond scalar metrics to visual diagnostic tools that map high-dimensional relationships into comprehensible 2D or 3D representations. 
Visualization bridges this interpretability gap, allowing practitioners to inspect surrogate models for physical realism, monotonic trends, and interaction thresholds. For example, PDP and SHAP simultaneously yield sensitivity metrics and visualize the response surface, proving highly valuable in complex engineering design~\cite{afdhal2023design, lyngdoh2022elucidating}. Because standard PDPs capture marginal effects but often miss coupled interactions~\cite{goldstein2015peeking}, the following subsections detail a suite of visualization techniques: PDP and Individual Conditional Expectations (ICE) for marginal and conditional effects, Accumulated Local Effects (ALE) for handling feature correlations, and SHAP for explicitly exposing interactions and latent low-dimensional structures. A more detailed discussion of SHAP, however, is provided in Section~\ref{sec:Instance-Specific-Diagnostics}, in the context of instance-specific diagnostic methods.
}

\subsubsection{Partial Dependence Plots and Individual Conditional Expectations}
The goal of Partial Dependence Plots (PDP)~\cite{friedman2001greedy} is to construct the partial dependence function with respect to the variables indexed by \(A\). Given a function $f: \boldsymbol{\Omega} \to \mathbb{R}$, where \( \boldsymbol{x} = (x_1, x_2, \ldots, x_m) \in \boldsymbol{\Omega} = \prod_{i=1}^m \Omega_i \),  the partial dependence function of \( f \) with respect to the variables $\boldsymbol{x}_A$ is defined as:
\textcolor{black}{\begin{equation}
  \resizebox{\columnwidth}{!}{$
    \displaystyle
    f_A(\boldsymbol{x}_A) = \mathbb{E}_{\boldsymbol{x}_B} \left[ f(\boldsymbol{x}_A, \boldsymbol{x}_B) \right] = \int_{\boldsymbol{\Omega}_B} f(\boldsymbol{x}_A, \boldsymbol{x}_B) \, \boldsymbol{\rho}_{\boldsymbol{x}_B}(\boldsymbol{x}_B) \, d\boldsymbol{x}_B,
$}
\end{equation}}
where $\mathbb{E}_{\boldsymbol{x}_B}$ means that we are taking the expected value with respect to the random vector $\boldsymbol{x}_{B}$. The PDP is then obtained by evaluating and visualizing \( f_A(\boldsymbol{x}_A) \) over the subdomain $\boldsymbol{\Omega}_A$, showing how the prediction of $f$ varies with $\boldsymbol{x}_A$, marginalizing over the remaining variables $\boldsymbol{x}_B$. The GSA metric based on the PDP for the $i$-th variable \textit{i.e.}, $A=\{i\}$, can be estimated by simply calculating the standard deviation of each PD function~\cite{greenwell2018simple}, written as
\textcolor{black}{\begin{equation}
    \sigma^2_{pd_i} = \mathrm{Var}[f_{i}(x_i)]
\end{equation}}

While the PDP provides the average effect of a subset of input variables $\boldsymbol{x}_A$ on the model output by marginalizing over $\boldsymbol{x}_B$, it may obscure important details when interactions exist between variables or when the model exhibits nonlinear behavior across the input space. In such cases, the PDP can be misleading, as it reflects only the mean trend, potentially hiding variations in individual responses. To go further, the Individual Conditional Expectation (ICE) plots were introduced to complement PDP~\cite{goldstein2015peeking}. ICE plots reveal model behavior at the individual observation level by showing the predicted response for each instance. \textcolor{black}{Specifically, let $\{\boldsymbol{x}_B^{(j)}\}_{j=1}^{n_B}$ denote $n_B$ realizations of the complementary variables drawn from $\boldsymbol{\rho}_{\boldsymbol{x}_B}$.  
For a fixed $\boldsymbol{x}_B^{(j)}$, the ICE curve is defined as
\begin{equation}
    f_A^{(j)}(\boldsymbol{x}_A) = f(\boldsymbol{x}_A, \boldsymbol{x}_B^{(j)}).
\end{equation}}
Plotting \(f_A^{(j)}(\boldsymbol{x}_A)\) across multiple \(j\) reveals the variability in the model's response across the data distribution.

By visualizing all ICE curves alongside their average (which corresponds to the PDP), one can assess both the average impact and the impact of interactions that the PDP overlooks. It should be emphasized that both PDP and ICE rely on the assumption that the input variables are independent. This allows the joint distribution to be expressed as the product of marginal densities. 

For pedagogical purposes, consider the function $f(x_1, x_2) = \sin(x_1 + x_2)+ x_1$ defined over the domain $\boldsymbol{\Omega} = [-1,1]^2$. The distributions are assumed to be uniform for both $x_1$ and $x_2$. By performing integration over [-1,1], the following can be obtained:
$    f_{x_{1}}(x_1) = \frac{1}{2} \left[\cos(x_1-1)-\cos(x_1+1) \right]+ x_1 $ and  $    f_{x_{2}}(x_2) = \frac{1}{2} \left[\cos(x_2-1)-\cos(x_2+1) \right]
$. 

\begin{figure}[tbh!]
	\centering
    \hspace{-0.25cm}
\includegraphics[width=1.025\columnwidth]{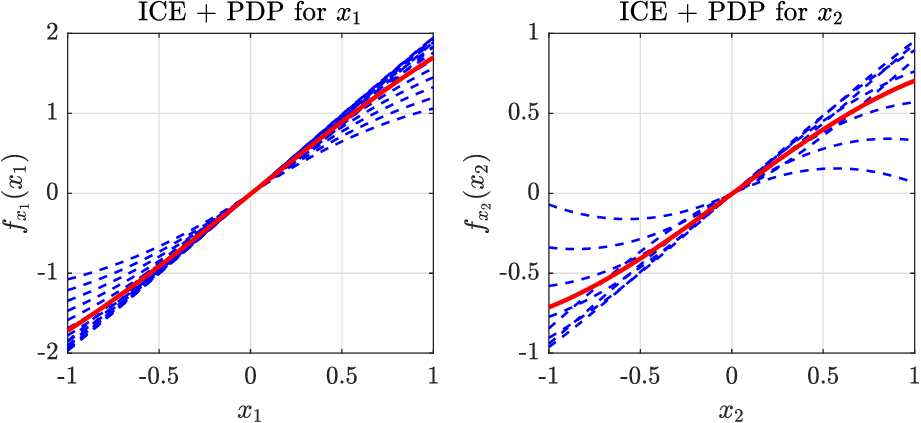}%
    \caption{PDP (in red)  and ICE (in blue) curves for the function $f_{x_1,x_2}=\sin(x_1+x_2)+x_1$.}
	\label{fig:PDP_2dfunc}
\end{figure}

The PDPs, along with several realizations of the ICE curves, are presented in Fig.~\ref{fig:PDP_2dfunc}. These visualizations illustrate how each input variable influences the model output. The ICE curves, in particular, reveal the variation of the function with respect to a specific input while holding the other variable fixed at selected values. In this example, the output exhibits stronger nonlinearity with respect to $x_2$, which is governed purely by a sinusoidal component. In contrast, the dependence on $x_1$ appears less nonlinear due to the presence of an additional linear term.
 Further calculation shows that the GSA metric values for $x_1$ and $x_2$ are approximately $1.0174$ and $0.4398$, respectively, indicating that $x_1$ has a greater influence on the output than $x_2$.

The PDP method can be used to reveal the interaction by showing the two-way partial dependence function. The two-way interaction plot shows the average impact of changing two variables together when other variables are fixed~\cite{friedman2001greedy}. Breaking down the single-variable partial dependence functions into multiple ICE lines can show the interaction effects. If interactions are present, variations in trend can be observed for each ICE curve. However, the traditional way of depicting ICE lines might make it convoluted to show which other variables interact with the variable being investigated. It is also worth mentioning the \textit{ceteris paribus} plot~\cite{kuzba2019pyceterisparibus}, which essentially shows how the output changes as we vary one input feature across its domain. In this regard, ICE curves are essentially the collection of \textit{ceteris paribus} plots for multiple instances. 

\subsubsection{Accumulated local effects}
\textcolor{black}{
It should be emphasized that both PDP and ICE rely on the assumption that the input variables are independent. This allows the joint distribution to be expressed as the product of marginal densities. However, when strong correlations exist among input variables, partial dependence estimates may lose accuracy or even lead to misleading interpretations~\cite{hooker2021unrestricted}, since the method implicitly assumes independence and may evaluate combinations of inputs that are unlikely or physically implausible under the true joint distribution.  Thus, PDP and ICE are most appropriately applied when the input variables are independent. To address this, ALE acts as a robust alternative, computing differences in predictions conditioned strictly on the observed data distribution.
}
Namely, ALE estimates the average effect of a feature by computing differences in predictions within small intervals, conditioned only on the observed data distribution~\cite{apley2020visualizing}. In contrast to PDP, which is based on marginal distribution, the nature of ALE, which relies on conditional distribution, makes it attractive for dealing with correlated features.
 \textcolor{black}{For a single input, the ALE is written as:}
\textcolor{black}{\begin{equation}
f_{i}^{\text{ALE}}(x_i)
=
\int_{z_{0,i}}^{x_i}
\mathbb{E}_{\boldsymbol{x}_{-i}\mid x_i = z}
\left[
\frac{\partial f(z,\boldsymbol{x}_{-i})}{\partial z}
\right]
dz
- C_i,
\end{equation}
where the centering constant $C_i$ ensures zero mean, so that:
\begin{equation}
\mathbb{E}_{x_i}
\left[
f_i^{\text{ALE}}(x_i)
\right]
=
0.
\end{equation}
In the above expression, $z$ is a dummy integration variable representing intermediate values along the axis of the input variable $x_i$. The lower bound $z_{0,i}$ is a reference value, typically chosen as the lower bound of the domain or the minimum observed value of $x_i$. 
}

\subsection{\textcolor{black}{Instance-Specific Diagnostics}}
\label{sec:Instance-Specific-Diagnostics}
In fields such as medicine and law, the model should be trustworthy in the sense that users can understand why and how a model makes certain predictions or decisions~\cite{le2024algorithmic, hossain2025explainable, tjoa2020survey}. 
\textcolor{black}{Trustworthiness is also essential in engineering applications, but it is typically defined and operationalized differently. In engineering design exploration that primarily leverages simulation, the primary concern is whether the model is sufficiently accurate and reliable for extracting meaningful design insight~\cite{palar2023enhancing}. Consequently, understanding the limitations of the constructed model becomes central. From this perspective, accuracy is characterized by proximity to the ground truth, which depends jointly on the fidelity of the data and the assumptions embedded in the model.} From the data viewpoint, the data's nature and limitations should also be thoroughly explained. As an illustration, if the data originates from a low-fidelity source, it is important to provide comprehensive clarification. This ensures that users of the model are informed about the fidelity level of the data. Information about the data domain is essential to exercise caution and avoid excessive extrapolation.

\textcolor{black}{From this perspective, trustworthiness is closely tied to an understanding of model limitations and uncertainty, consistent with established principles in verification, validation, and uncertainty quantification~\cite{roy2011comprehensive}. Within this framework, understanding the sources and structure of uncertainty is as important as prediction accuracy itself. Interpretable and explainable surrogate models can be viewed as complementary tools that support these objectives by enabling diagnostic insight into model behaviour.} 

\textcolor{black}{Building upon the macroscopic insights provided by global sensitivity metrics and interaction tools like PDP and ICE, instance-specific diagnostics offer a complementary, microscopic view. While global tools map the broad dynamics of the design space, they must be paired with local methods to understand exactly \textit{why} the surrogate model produces a particular prediction at a specific input configuration. In an engineering context, this localized interpretation is critical when evaluating specific candidate designs, analyzing boundary failure modes, or diagnosing critical operating conditions.
To achieve this, instance-level methods assess how local perturbations around a specific input affect the prediction. This localized scrutiny plays an indispensable role in establishing the trustworthiness of surrogate models, allowing practitioners to verify whether a specific prediction is driven by physically reasonable factors rather than numerical artifacts. Furthermore, the relationship between local and global XAI is deeply synergistic; these local explanations are not limited to pointwise analysis. When aggregated across multiple samples, local attributions can reconstruct global sensitivity measures, perfectly bridging the gap between instance-level diagnostics and system-wide behavior. In the following, we discuss representative instance-level explanation techniques that decompose complex black-box predictions into interpretable, feature-wise contributions.}

\subsubsection{Shapley Values}
The Shapley value originates from cooperative game theory~\cite{shapley1953stochastic}, and defines a fair way to distribute a total "value" among a set of input variables. Let us focus on a singleton $i \in [1:m]$ first. Define a value function $v : 2^{[1:m]}\rightarrow \mathbb{R}$, hence $v(\cdot)$ is the value function associated with a certain subset. The Shapley value is defined as:
\begin{equation}
\label{eq:shapley_value}
  \resizebox{\columnwidth}{!}{$
    \displaystyle
    \phi_i = \sum_{U \subseteq [1:m] \setminus \{i\}} \frac{|U|! \, (m - |U| - 1)!}{m!} \left( v(U \cup \{i\}) - v(U) \right),
    $}
\end{equation}
The value function can reflect quantities such as cost, reward, or variance. The quantity $v(U \cup \{i\}) - v(U)$ reflects the marginal contribution of the $i$-th input variable to the subset $U$. Further, $ \frac{|U|! \, (m - |U| - 1)!}{m!}$ is the weight, which ensures equal consideration of all possible orderings of input variables. Hence, the Shapley value is the average contribution of the $i$-th variable to the value function $v$, averaged across all possible orderings.

\textcolor{black}{The Shapley value concept has also been extended to variance-based sensitivity analysis. For independent inputs, Shapley effects provide an alternative to the total Sobol’ index for quantifying overall input importance~\cite{owen2014sobol}. Applying Eq.~(\ref{eq:shapley_value}) with  $v(U) = S_U$, where $S_U$ denotes the Sobol' index of subset $U$ (see Eq.~\ref{eq:Sobol}), yields the Shapley effect $Sh_i$. For example, if $m=3$ and $i=1$, then $Sh_i = S_{\{1\}}+\frac{1}{2}S_{\{1,2\}}+\frac{1}{2}S_{\{1,3\}}+\frac{1}{3}S_{\{1,2,3\}}$. In contrast to Sobol' indices, we have $\sum_{i=1}^{m}Sh_i= 1$.}

\textcolor{black}{Going back to local XAI, the concept of Shapley additive Explanaitions (SHAP) is based on the concept of Shapley values~\cite{lundberg2017unified}. In SHAP, the canonical Shapley formulation, see Eq.~(\ref{eq:shapley_value}), is applied with the value function defined as $v(U)=f_U(\boldsymbol{x})$,
where $f_U(\boldsymbol{x})$ denotes the model prediction conditioned on the subset of features $U$. It is the adaptation of Shapley values to ML model explainability. The resulting explanation satisfies the additive decomposition:
\begin{equation}
    f(\boldsymbol{x}) = \phi_0 + \sum_{i=1}^m \phi_i,
\end{equation}
where $\phi_i$ denotes the SHAP value of the $i$-th feature for the instance $\boldsymbol{x}$, and $\phi_0$ represents the reference prediction, typically taken as the expected model output over the input distribution.}

\textcolor{black}{To quantify pairwise interactions, we can use the SHAP interaction values~\cite{fujimoto2006axiomatic}}. For features $i$ and $j$, the SHAP value $ \phi_{i,j}$ is written as:
\begin{equation}
\begin{aligned}
&\phi_{i,j} = \sum_{U \subseteq [1:m] \setminus \{i,j\}} \frac{|U|! \, (m - |U| - 2)!}{(m-1)!} \big( \quad \hspace{1em}  \\ 
&\hspace{0.1em}\ v(U \cup \{i,j\}) - v(U \cup \{i\}) - v(U \cup \{j\}) + v(U) \big),
\end{aligned}
\end{equation}
which quantifies the interaction effect between $x_i$ and $x_j$ at a specific instance $\boldsymbol{x}$ after removing their individual contributions. If $i=j$, then we have the main effect according to the SHAP value, written as:
\begin{equation}
    \phi_{i,i}(\boldsymbol{x}) = \phi_i(\boldsymbol{x})-\sum_{j \neq 1} \phi_{i,j}(\boldsymbol{x})
\end{equation}
The total SHAP value is recovered by:
\begin{equation}
    \phi_{i}(\boldsymbol{x}) =  \sum_{j=1}^{m}\phi_{i,j}(\boldsymbol{x})
\end{equation}

The global sensitivity metric based on SHAP value can be defined as~\cite{lundberg2017unified}:
\begin{equation}
    I_i^{\text{SHAP}}  = \mathbb{E}\left[|\phi_i(\boldsymbol{x})|\right] = \int_{\boldsymbol{\Omega}} |\phi_{i}(\boldsymbol{x})| \boldsymbol{\rho}(\boldsymbol{x})d\boldsymbol{x},
\end{equation}
which can be estimated using \textcolor{black}{Monte Carlo}. The mean absolute SHAP value itself represents the average magnitude of contribution of the $i$-th input to the model's prediction across the input distribution. Similarly, we can also compute the mean absolute interaction according to SHAP values~\cite{lundberg2020local}:
\begin{equation}
    I_{i,j}^{\text{SHAP}}  = \mathbb{E}\left[|\phi_{i,j}(\boldsymbol{x})|\right] = \int_{\boldsymbol{\Omega}} |\phi_{i,j}(\boldsymbol{x})| \boldsymbol{\rho}(\boldsymbol{x})d\boldsymbol{x},
\end{equation}

There are several methods to compute SHAP values. The first is the exact method, which should be used whenever the computational cost allows us to do so. An important method to estimate SHAP value is the KernelSHAP, which provides a fast and convenient way for large sample sizes and high-dimensionality. KernelSHAP approximates SHAP values by sampling feature coalitions, evaluating the model with absent features replaced by reference/background values, and recovering feature contributions via a weighted local linear regression using the Shapley kernel~\cite{lundberg2017unified}. 
As aforementioned, beyond prediction, an emulator could serve as a tool for knowledge discovery, and such a role is fulfilled by explainability methods when the surrogate is opaque. For example, SHAP has been used to identify the influence of geometric variables on compressor performance~\cite{wang2022shapley,zhang2023metamodel} and to reveal flow features on the mesh~\cite{shen2022automatic}. SHAP performs well and is rich in information when the goal is to depict interaction. By depicting the scatter plot of SHAP values for one particular variable and coloring it according to the magnitude of the other variable, one can see the interaction that is indicated by how the trend of the SHAP values changes as a function of the other variable~\cite{lundberg2017unified}.

\subsubsection{Model-Agnostic Local Explanations}
Local Interpretable Model-agnostic Explanations (LIME) consist in a local XAI technique that aims to interpret the prediction of any black-box model~\cite{ribeiro2016should}. Its core concept involves approximating the complex model's behavior in the vicinity of a particular prediction by employing a simpler, more understandable surrogate model (\textit{e.g.}, linear regression or a decision tree). LIME constructs a local dataset by generating perturbations around the inputs of interest $\boldsymbol{x}$ and evaluating the models on these points. A weighting kernel is then used to assign higher importance to perturbations closer to $\boldsymbol{x}$. Next, an interpretable model $g \in \mathcal{G}$ is trained to approximate $\hat{f}(\boldsymbol{x})$ in the vicinity of $\boldsymbol{x}$ by minimizing the following objective:
\begin{equation}
    \arg\min_{g \in \mathcal{G}} L(\hat{f}, g, \pi_{\boldsymbol{x}}) + \psi(g)
\end{equation}
where $L$ is the loss function, $\pi_{\boldsymbol{x}}$ is the proximity weighting function, $\psi(g)$ is the regularization term to penalize the complexity in the interpretable model (specified by the user). The resulting surrogate model $g$ serves as a faithful local approximation $f$, and its coefficients denote the contributions of each input variable to the prediction at $\boldsymbol{x}$. These choices lead to different variants of LIME, as systematically reviewed in the recent survey~\cite{knab2025lime}.
As the name suggests, LIME is a local-agnostic method that can be applied to any regression and classification model. LIME is also computationally efficient and provides individual explanations for specific instances. However, the interpretation of LIME can be sensitive to the sampling strategy, choice of kernel, and the form of the local surrogate model.

\textcolor{black}{Complementing surrogate-based methods like LIME, Leave-One Covariate Out (LOCO) offers a rigorous perturbation-based approach to local importance. Unlike LIME, which approximates the model, LOCO queries the original black-box directly by "dropping" or ablating a specific variable, replacing it with a null value or marginalizing it, to observe the resulting impact on prediction accuracy~\cite{lei2018distribution}. This "ablate-and-test" logic provides a direct measure of a feature's predictive power without the risk of surrogate misfitting. To address the complexities of modern engineering models, Interactions LOCO (iLOCO) extends this framework to quantify higher-order feature interactions~\cite{little2025iloco}. A significant advantage of the LOCO family is the use of distribution-free confidence intervals; this allows practitioners to obtain statistically robust importance scores without making restrictive assumptions about the underlying data distribution. This makes iLOCO particularly valuable for high-dimensional engineering problems where feature dependencies are non-linear and traditional parametric statistics might fail.}

Contextual Importance and Utility (CIU) is a local, model-agnostic, post-hoc explainability method that quantifies how much the output of a model could change when varying a feature within its local context (Contextual Importance), as well as how favorable the current feature value is for achieving the prediction (Contextual Utility)~\cite{framling2022contextual}. 
Unlike additive feature attribution methods such as LIME or SHAP, CIU provides bounded, faithful explanations by explicitly accounting for the local feature ranges and interactions, offering richer and more meaningful insights into the model’s decision-making process. CIU supports contrastive explanations that help users understand not only why a decision was made but also how it could be changed, thus providing actionable insights. Empirical studies confirm that CIU explanations are easier to interpret and more aligned with human intuition, especially in high-stakes domains like healthcare and engineering~\cite{hakkoum2024global}.


\subsubsection{Gradient-based Attributions and Saliency}
\label{sec:gradients}

\textcolor{black}{
While the local emulators approximate behavior via a secondary model, gradient-based attributions utilize the main surrogate's internal gradients to provide instance-specific explanations~\cite{ortigossa2025t}. This branch of XAI traces its lineage to the foundational work on saliency maps, introduced to visualize the sensitivity of a specific output with respect to its input features~\cite{simonyan2013deep}. It is critical to distinguish between \textit{classical sensitivity derivatives} used for global screening and these \textit{XAI saliency explanations}. While the former evaluates the local rate of change to understand system stability,  \textcolor{black}{methods} like Integrated Gradients~\cite{sundararajan2017axiomatic} aggregate gradients along a continuous path to satisfy the completeness axiom, ensuring that the sum of individual feature attributions perfectly accounts for the overall prediction.} 
\textcolor{black}{Mathematically, the Integrated Gradient for the $i$-th feature of an input $\boldsymbol{x}$ relative to a baseline $\boldsymbol{x}'$ is defined as the path integral of the gradients along the straight line from the baseline to the input:}
\begin{equation}
    \text{IG}_i(\boldsymbol{x}) = (x_i - x'_i) \int_{0}^{1} \frac{\partial \hat{f}(\boldsymbol{x}' + \alpha(\boldsymbol{x} - \boldsymbol{x}'))}{\partial x_i} d\alpha.
\end{equation}
\textcolor{black}{By integrating over the scaling parameter $\alpha \in [0,1]$, this formulation explicitly guarantees completeness: $\sum_{i=1}^m \text{IG}_i(\boldsymbol{x}) = \hat{f}(\boldsymbol{x}) - \hat{f}(\boldsymbol{x}')$. This provides a conservative, rigorous attribution that ensures no gradient information is "lost" due to local gradient saturation, which is important for safety-critical diagnostics.}

\textcolor{black}{A common criticism of these methods in complex engineering contexts is the instability of Monte Carlo estimation. To address this, recent deterministic approaches have been developed that leverage the piecewise linear structure of surrogates like ReLU-based networks. These exact solvers avoid the sampling noise of MC methods, providing the numerical stability necessary for robust engineering diagnostics~\cite{simonyan2013deep, anderson2024provably}.}

\textcolor{black}{Beyond these approaches,} there also exist specialized explainability methods designed for neural networks in the context of image processing (\textit{e.g.}, convolutional neural networks). For example, saliency maps provide explanations by highlighting pixels that were relevant to a prediction in image classification. The Grad-CAM method is a popular technique in this category, which uses the gradients of a target concept flowing into the final convolutional layer to produce a coarse localization map~\cite{selvaraju2020grad}.

In terms of method development, the landscape of explainable ML is broad and continually evolving. In addition to region-based methods like Grad-CAM, there are a number of gradient-based attribution techniques that provide more fine-grained, pixel or feature-level explanations. One simple variant multiplies the input feature by the gradient of the output with respect to that feature (often called “Gradient $\times$ Input”), thus attributing importance when the input is large, and the gradient is large. To improve the stability of such maps, the SmoothGrad method adds small random noise to the input many times and averages the results to yield smoother, less “spiky” attribution maps~\cite{smilkov2017smoothgrad}. As previously noted, Integrated Gradients accumulates gradients along a path from a baseline to the actual input to satisfy desirable axioms in attribution theory~\cite{sundararajan2017axiomatic}. Another method, DeepLIFT (“Deep Learning Important FeaTures”), compares the model’s activation to a reference activation and back-propagates those differences to assign contributions, helping especially when activations saturate~\cite{shrikumar2017learning}. Finally, Gradient SHAP combines ideas from the Shapley-value framework with gradient information to approximate fair feature-attributions in deep models~\cite{krishna2022disagreement}.

\subsection{\textcolor{black}{Logic-based and Counterfactual Reasoning}}

\subsubsection{Counterfactuals and Decision Rules}

Counterfactual explanation is another explainability technique, particularly effective for classification tasks compared to regression. It operates as a local explanation method, pinpointing the smallest modifications needed in an input to change the model’s output~\cite{wachter2017counterfactual, dandl2020multi}. Essentially, counterfactual explanations answer the question: "How could this input be adjusted to achieve a different (usually more favorable) prediction?" In the context of classification, a counterfactual explanation might indicate that increasing or reducing certain input variables would flip the predicted label. This is why counterfactuals are less straightforward for a \textbf{regression} problem, since there is no natural class boundary to cross. Although the method can still operate in regression by defining a target continuous value, the interpretation is often less intuitive. It might, however, be highly useful for certain engineering problems, such as reliability analysis or dynamical systems with bifurcations.  

\textcolor{black}{Formally, finding a counterfactual $\boldsymbol{x}_{cf}$ for a given instance $\boldsymbol{x}$ is framed as a constrained optimization problem. It balances the proximity to the original input with the achievement of a targeted output state $y^*$: }
\begin{equation}
    \boldsymbol{x}_{cf} = \arg\min_{\boldsymbol{x}' \in \boldsymbol{\Omega}_{\text{valid}}} \Big( d(\boldsymbol{x}, \boldsymbol{x}') + \lambda \cdot \mathcal{L}\big(\hat{f}(\boldsymbol{x}'), y^*\big) \Big),
\end{equation}
\textcolor{black}{where $d(\cdot, \cdot)$ is a distance metric (e.g., $L_1$ or $L_2$ norm) penalizing large alterations, $\mathcal{L}$ is a loss function driving the prediction toward $y^*$, and $\lambda$ controls the trade-off. Crucially for engineering design, the search space must be restricted to $\boldsymbol{\Omega}_{\text{valid}}$, representing the feasible domain to ensure the generated counterfactual respects physical laws, boundary conditions, or manufacturing constraints.}

A method related to LIME is the Anchor method, which builds on the principles of local approximation~\cite{ribeiro2018anchors} but instead aims to generate decision rules. The core idea is to identify a rule that defines a coherent explanation (a scoped rule) between the target instance and its local neighborhood. Similar to LIME, Anchor creates perturbations around the instance to be explained; however, rather than fitting a local surrogate model, it constructs an interpretable “IF...THEN...” decision rule. These rules consist of a conditional part (a subset of features with specific values) and a consequent part, which is simply the predicted class. Each rule is associated with a precision score, indicating the proportion of perturbed instances in the neighborhood that follow the rule and retain the original prediction, and a coverage score, reflecting the number of instances in the entire dataset that satisfy the rule. While the method remains computationally efficient and yields clear, intuitive explanations, Anchor shares some limitations with LIME, particularly concerning the generation of realistic perturbations, which can be challenging to configure properly without violating physical constraints. 

Similarly, LOcal Rule-based Explanations (LORE) generate local, rule-based explanations (providing both factual and counterfactual rules) by learning a decision tree on a synthetically generated neighborhood using a genetic algorithm, combining high interpretability with structured local insights~\cite{guidotti2018local}. Finally, MAPLE (Model Agnostic Post-hoc Local Explanations) combines local linear models weighted by similarity. It offers both interpretability and moderate fidelity by utilizing local linear modeling techniques alongside a dual interpretation of random forests—serving simultaneously as a supervised neighborhood approach and a feature selection method~\cite{plumb2018model}.

\subsubsection{Formal Logic Verification}

Also, in the domain of formal XAI, Abductive Explanations (AXp), also known as prime implicant or minimally sufficient reasons, determine the minimal set of feature-value assignments that guarantee a model’s decision for a given instance, thereby answering “why?” questions with logical precision.
\textcolor{black}{Mathematically, an AXp for a predicted instance $\boldsymbol{x}$ is a minimal subset of feature indices $\mathcal{A} \subseteq \{1, \dots, m\}$ such that fixing these specific features guarantees the original prediction $y = \hat{f}(\boldsymbol{x})$, regardless of the values taken by the remaining features. Formally, this is expressed as a logical entailment:}
\begin{equation}
    \forall \boldsymbol{z} \in \boldsymbol{\Omega}, \quad \big( \boldsymbol{z}_{\mathcal{A}} = \boldsymbol{x}_{\mathcal{A}} \big) \implies \hat{f}(\boldsymbol{z}) = \hat{f}(\boldsymbol{x}),
\end{equation}
\textcolor{black}{subject to the minimality condition that no proper subset of $\mathcal{A}$ satisfies this implication. Because this relies on strict logical entailment rather than statistical correlation, it provides compelling soundness guarantees well-suited for high-stakes aerospace or structural applications where formal certification is required.}
 In contrast, Contrastive Explanations (CXp), sometimes called minimal required changes, pinpoint the smallest set of feature alterations that would switch the predicted outcome, effectively addressing “why not?” or “how could this decision be different?” queries with subset- or cardinality-minimal guarantees~\cite{ignatiev2020towards,izza2024distance} and serve in argumentation theory~\cite{vassiliades2021argumentation}. 
\textcolor{black}{Mathematically, a CXp is defined as a minimal subset of feature indices $\mathcal{C} \subseteq \{1, \dots, m\}$ such that allowing only the features in $\mathcal{C}$ to be perturbed (while keeping the remaining features fixed at their original values) enables the prediction to change. Formally, there exists a valid counterfactual instance $\boldsymbol{z}$ such that:}
\begin{equation}
    \exists \boldsymbol{z} \in \boldsymbol{\Omega}, \quad \big( \boldsymbol{z}_{\overline{\mathcal{C}}} = \boldsymbol{x}_{\overline{\mathcal{C}}} \big) \land \big( \hat{f}(\boldsymbol{z}) \neq \hat{f}(\boldsymbol{x}) \big),
\end{equation}
\textcolor{black}{where $\overline{\mathcal{C}} = \{1, \dots, m\} \setminus \mathcal{C}$ denotes the set of unperturbed features, subject to the condition that $\mathcal{C}$ is minimal (i.e., no proper subset of $\mathcal{C}$ permits a change in the prediction). }

Remarkably, AXp and CXp are dual in nature: each is a minimal hitting set of the other, enabling efficient enumeration by leveraging logic-based techniques like the Maximum Satisfiability problem (MaxSAT) or Satisfiability Modulo Theories (SMT)~\cite{iser2024automated,lagniez2024sat}. While computing AXp/CXp in tree ensembles remains NP-hard, recent developments demonstrate that exact, polynomial-time procedures exist for certain model classes (such as decision trees and lists), making exact formal explanations tractable in practice~\cite{marques2024explainability}. This formalism thus offers compelling soundness guarantees and actionable insights well-suited for high-stakes applications. Beyond these methods, surrogate explainers, and particularly decision trees trained post‑hoc, serve as transparent approximations to complex logical classifiers. Importantly, it can be shown that for any classifier, there exists a decision‑tree surrogate that exactly matches its decision boundary. Moreover, practical deployment that relies on approximate surrogate models is computable in polynomial time~\cite{izza2022tackling}. However, inherently interpretable models like decision trees can suffer from explanation redundancy, where extracted paths include superfluous conditions. 
\begin{table*}[tb!]
    \centering
    \resizebox{0.99\linewidth}{!}{
\centering
\begin{tabular}{|c|c|c|c|c|c|}
\hline
 \textbf{Method} 
  & \parbox[t]{2.5cm}{\centering\textbf{Interpretability}}
  & \parbox[t]{2.5cm}{\centering\textbf{Inter-feature}\\\textbf{Analysis}}
  & \parbox[t]{2cm}{\centering\textbf{Computation}\\\textbf{Time}}
  & \parbox[t]{2.5cm}{\centering\textbf{Feature}\\\textbf{Independence?}}
  & \parbox[t]{2.5cm}{\centering\textbf{Attribution}\\\textbf{Accuracy}} \\
\hline
\hline
ICE/PDP
  & \cellcolor[HTML]{006400}\textcolor{white}{Very Easy}
  & N/A
  & \cellcolor[HTML]{006400}\textcolor{white}{Low}
  & N/A
  & N/A \\
\hline
Counterfactuals
  & \cellcolor[HTML]{8FBC8F}{Moderate}
  & \cellcolor[HTML]{FFA500}{Difficult}
  &  \cellcolor[HTML]{FFA500}{High}
  & N/A
  & N/A \\
\hline
LIME
  & \cellcolor[HTML]{006400}\textcolor{white}{Easy}
  & \cellcolor[HTML]{8FBC8F}{Possible}
  & \cellcolor[HTML]{006400}\textcolor{white}{Low}
  &\cellcolor[HTML]{FFCC66}{Yes}
  & \cellcolor[HTML]{FF8C00}{Low} \\
\hline
Anchor
  & \cellcolor[HTML]{006400}\textcolor{white}{Easy}
  & \cellcolor[HTML]{8FBC8F}{Possible}
  & \cellcolor[HTML]{006400}\textcolor{white}{Low}
  & \cellcolor[HTML]{FFCC66}{Yes}
  & \cellcolor[HTML]{FF8C00}{Low} \\
\hline
SHAP
  & \cellcolor[HTML]{006400}\textcolor{white}{Easy}
  & \cellcolor[HTML]{8FBC8F}{Possible}
  & \cellcolor[HTML]{FFA500}{High}
  & \cellcolor[HTML]{FFCC66}{Yes}
  & \cellcolor[HTML]{8FBC8F}{Moderate} \\
\hline
LORE
  & \cellcolor[HTML]{006400}\textcolor{white}{Easy}
  & \cellcolor[HTML]{8FBC8F}{Possible}
  &  \cellcolor[HTML]{8FBC8F}{}{Moderate}
  & \cellcolor[HTML]{FFCC66}{Yes}
  & \cellcolor[HTML]{8FBC8F}{Moderate} \\
\hline
iLOCO
  &   \cellcolor[HTML]{8FBC8F}{Moderate}
  & \cellcolor[HTML]{006400}\textcolor{white}{Yes}
  & \cellcolor[HTML]{FFA500}{High}
  & \cellcolor[HTML]{006400}\textcolor{white}{No}
  & \cellcolor[HTML]{006400}\textcolor{white}{High} \\
\hline
MAPLE
  & \cellcolor[HTML]{006400}\textcolor{white}{Easy}
  & \cellcolor[HTML]{8FBC8F}{Possible}
  & \cellcolor[HTML]{8FBC8F}{Moderate}
  & \cellcolor[HTML]{FFCC66}{Yes}
  &  \cellcolor[HTML]{8FBC8F}{Moderate} \\
\hline
CIU
  & \cellcolor[HTML]{8FBC8F}{Moderate}
  & \cellcolor[HTML]{006400}\textcolor{white}{Yes}
  & \cellcolor[HTML]{8FBC8F}{Moderate}
  & \cellcolor[HTML]{006400}\textcolor{white}{No}
  & \cellcolor[HTML]{8FBC8F}{Moderate} \\
\hline
ALE 
  & \cellcolor[HTML]{8FBC8F}{Moderate}
  & \cellcolor[HTML]{006400}\textcolor{white}{Yes}
  & \cellcolor[HTML]{8FBC8F}{Moderate}
  & \cellcolor[HTML]{006400}\textcolor{white}{No}
  & \cellcolor[HTML]{8FBC8F}{Moderate} \\
\hline
KernelSHAP
  & \cellcolor[HTML]{006400}\textcolor{white}{Easy}
  & \cellcolor[HTML]{8FBC8F}{Possible}
  & \cellcolor[HTML]{FFA500}{High}
  & \cellcolor[HTML]{FFCC66}{Yes}
  & \cellcolor[HTML]{8FBC8F}{Moderate} \\
  \hline
PermutaSHAP
  & \cellcolor[HTML]{006400}\textcolor{white}{Easy}
  & \cellcolor[HTML]{8FBC8F}{Possible}
  & \cellcolor[HTML]{FF8C00}{Very High}
  & \cellcolor[HTML]{FFCC66}{Yes}
  & \cellcolor[HTML]{006400}\textcolor{white}{High} \\
\hline

\hline
\end{tabular}
}
\caption{Summary of advantages and limitations of local, post-hoc, model-agnostic XAI methods (adapted from~\cite{aligon2024vers}).}
\label{tab:xai-summary}
\end{table*}

\subsubsection{Argumentation-Based XAI}
Argumentation-based XAI is particularly well-suited to ABMs. 
In ABMs, agents often use rule-based or logic-based decision processes, and argumentation frameworks can surface those decisions as explicit claims, counterclaims, and attacks. 
Such an approach enhances trust and auditability because stakeholders can trace back which rules or strategies agents use, how they resolve internal conflicts, and why certain actions prevail~\cite{calegari2021logic}. Argumentation-based XAI is also well-suited to engineering models that serve certification, robustness, and risk prevention as they derive from logical rules and deterministic decisions~\cite{gariel2023framework}.
Moreover, combining complex systems surrogate models with argumentation-based explanations yields a powerful hybrid workflow: the surrogate provides fast predictions of the underlying engineering problem, while the argumentation layer reconstructs the logical reasoning, offering explanations that are faithful, transparent, and aligned with both the learned behavior and formal system logic. This hybrid approach supports critical tasks such as debugging, validation, risk assessment, and regulatory compliance in engineering or socio-technical systems~\cite{wilson2020systematic}. 

Argumentation‑based XAI constructs explanations as a structured debate inside the AI system. In this framework, every decision is represented by a set of “arguments” (claims supported by evidence or rules) and “attacks” (where one argument contradicts or undermines another)~\cite{vassiliades2021argumentation} and the final decision is explained by the subset of arguments that remain undefeated under formal criteria (semantics)~\cite{coste2014revision}.
Moreover, argumentation-based XAI naturally supports interactive explanation dialogues in dynamic simulations. In multi-agent conflict resolution, agents (or humans and agents) can engage in structured debates that justify, contest, and revise decisions in real time. For example, an expert can ask “why did this agent choose strategy A?” and obtain formal counterarguments and reasons. Arguments in such frameworks decompose into a few compact layers: the \textbf{structural}, \textbf{relational}, \textbf{dialogical}, \textbf{assessment}, and \textbf{rhetorical} layers~\cite{atkinson2017towards}. Compared with opaque models, argumentation makes reasoning explicit: conclusions link to concrete rules, conflicts are resolved transparently, and the surviving arguments can be communicated as natural language or graphical maps, yielding explanations that are both inspectable and accessible to stakeholders~\cite{doutre2025incomplete}.

Argumentation‑based XAI methods fall into three complementary categories~\cite{atkinson2017towards}: intrinsic explanations built natively into models, and complete or approximate post‑hoc explanations applied to existing “black‑box” models. In practice, many approaches blend these styles. The survey~\cite{vcyras2021argumentative} demonstrates the diversity of XAI based on an argumentation framework in the kinds of models explained, the argumentation formalisms used, and the formats of the resulting explanations, and highlights how symbolic argumentation framework can be paired with statistical or probabilistic techniques.

\subsection{\textcolor{black}{Bridging Data Analytics and Traditional Engineering}}
\textcolor{black}{
While ML offers powerful data-driven predictive capabilities to expedite design and optimization~\cite{panchal2019machine, saves2025XAIApp}, its black-box nature often clashes with traditional engineering practices that strictly prioritize safety, reliability, and industry standards. Rather than fully replacing conventional methods, explainable ML serves as a critical bridge~\cite{geyer2024explainable}. By making model reasoning transparent, XAI allows engineers to verify that data-driven predictions are consistent with established physical and engineering principles~\cite{vadyala2022review}. Furthermore, while XAI can unveil previously undiscovered design behaviors, this transparency ensures that such novel insights remain logically coherent with prior engineering knowledge.}

The strengths and limitations of the local, post-hoc, model-agnostic XAI methods discussed above are summarized in Table~\ref{tab:xai-summary} (adapted from~\cite{aligon2024vers}). In the context of engineering diagnostics, we advocate for the use of attribution-based techniques, particularly model-agnostic approaches like SHAP, as they offer \textcolor{black}{explainable}, per-feature contribution scores applicable to any surrogate model. While these approaches entail certain drawbacks, notably regarding computational cost and robustness, we argue that these trade-offs are acceptable given the diagnostic insight and ease of deployment they afford.

While both explainable ML and data mining aim to extract actionable insights, their underlying mechanics differ significantly~\cite{garouani2022towards}. Data mining typically utilizes unsupervised techniques, such as clustering or self-organizing maps~\cite{kohonen1982self}, to uncover hidden patterns directly from raw datasets $(\boldsymbol{x}, y)$ without relying on predictive models~\cite{shu2023knowledge,shi2023data,song2023aerodynamic}. 
Conversely, explainability methods do not elucidate the data itself, but rather interpret the reasoning of a trained predictive model $\hat{f}(\boldsymbol{x})$.

\textcolor{black}{
Despite their distinct mechanics, explainable ML and data mining are highly synergistic in simulation-driven exploration~\cite{atzmueller2024explainable, jeong2005data}. Practitioners can sequentially apply XAI to extract design rules mapping input variables to performance metrics~\cite{palar2024multi}, and, in the context of multi-objective optimization, subsequently deploy data mining to analyze optimal trade-offs along the Pareto front. This hybrid approach yields a holistic understanding of complex design spaces, successfully demonstrated in applications like canard-controlled missiles~\cite{yoo2024multi} and Mars aircraft airfoils~\cite{park2023multi}.
Furthermore, recent platforms like AMLBID operationalize this synergy by embedding explainability directly into AutoML pipelines for industrial datasets~\cite{garouani2022towards}. Rather than purely mining raw data, these frameworks use explainers to justify \emph{why} specific models and features are selected during meta-learning. This dual capability bridges unsupervised analytics and predictive modeling, delivering transparent, actionable outputs that directly inform human-in-the-loop engineering design decisions~\cite{garouani2022towards}.
}

\textcolor{black}{
To leverage the benefits of explainable ML technology for multi-objective design, it is necessary to initially construct a supervised ML model for the multiple objective functions. Explainable ML can facilitate multi-objective design exploration by leveraging information from distinct ML models, each focusing on a different output of interest. This stands in contrast to the data mining techniques applied to the set of Pareto solutions. Recent studies demonstrate that SHAP can better elucidate the impact of individual design variables on the trade-offs between objectives~\cite{palar2024multi, satria2024design, takanashi2023shapley}. By examining the correlation between SHAP values of input variables across multiple objectives, SHAP allows for an exploration of the variables that contribute to the concurrent enhancement or deterioration of multiple objectives. The same knowledge can also be extracted by investigating the trends obtained from partial dependence functions of the multiple objectives~\cite{palar2025partial}. However, since partial dependence functions are essentially the average, it is difficult to infer how interactions between variables contribute to the trade-offs. Utilizing ICE plots is also an option, although it may potentially complicate the analysis. It is important to highlight that extracting such knowledge solely from Pareto optimal solutions (without any supervised ML models) is challenging because it does not consider the information regarding the output's dependency on the input. However, both frameworks can be applied together to elucidate the problem's complexity and eventually produce insight from multiple viewpoints.} 

Arguably, data analytics and physical insight play a more important role when analyzing the set of optimal solutions. In this sense, physical and engineering reasoning prove valuable in explaining why the obtained design can enhance performance. Nonetheless, the concept of explainability might still be useful for examining how the input attributes of the design variables influence the optimal solutions. One should bear in mind that the model should be globally accurate so that the input attribution on the optimal solution(s) is accurate. That is because optimization often prioritizes obtaining the optimal solution without considering the overall accuracy of the model. \textcolor{black}{To the best of our knowledge, while several recent studies, discussed above, have begun to combine explainability methods with design optimization workflows, a systematic and mature integration of explainability into design optimization remains an active and relatively under-explored research area.}

\section{\textcolor{black}{Integrating XAI into Surrogate Workflows}}
\label{sec:surrogate_ml}

This section surveys state-of-the-art methods to link surrogate modeling to XAI, expanding on the conceptual workflow introduced in Section~\ref{sec:engineering_design_exploration}. Rather than prescribing a single recipe, we review and compare practical techniques from the literature for (i) designing informative experiments under limited budgets, (ii) training and diagnosing surrogate model families (GPs, tree ensembles, NNs, etc.), and (iii) applying complementary global and local XAI tools to characterize sensitivities, interactions, and failure modes~\cite{saves2025XAIApp}. Indeed, two recurring themes emerge across three tasks: the complementarity of methods (no single technique suffices for all goals) and the need for validation-driven iteration through active learning (diagnostics that trigger resampling, model refinement, or selective high-fidelity reevaluation). As such, the remainder of this section reviews these techniques for surrogate-based XAI: design of experiments and screening, derivative- and variance-based sensitivity, input–output and interaction analysis, dimensionality reduction, and uncertainty-aware design. \textcolor{black}{For each of these thematic categories, our objective is not to provide an exhaustive, method-by-method survey. Rather, we discuss how representative approaches can be leveraged to support the corresponding explanatory objective in surrogate modeling. In this way, the review is organized around the functional roles these techniques play in enabling knowledge extraction and drawing insight, rather than around individual algorithms themselves.}

\textcolor{black}{It is worth noting that additional concerns arise when explainability methods are applied to surrogate models. Explanations are obtained for the surrogate model rather than for the original simulator, and their validity therefore depends critically on the fidelity of the data used to construct the surrogate, the sampling design, and the accuracy of the approximation. Consequently, explanations should be interpreted as reflecting the behavior of the surrogate within the fidelity limits of the underlying data, rather than as direct explanations of the high-fidelity simulator itself. From this perspective, explainability in surrogate modeling is inherently conditional and context-dependent. Users must be aware that explanations are meaningful only within the domain where the surrogate is accurate and where the data faithfully represent the underlying physical or computational model. Therefore, the approaches discussed in this section should be interpreted within this surrogate-specific context, namely with explicit consideration of data fidelity, approximation accuracy, and the surrogate’s domain of validity.}

\textcolor{black}{To operationalize this approach, we frame the integration of XAI across six themes that map directly to the following subsections, beginning with data acquisition and global exploration and progressing through physical trend analysis and dimensionality reduction. This workflow continues with robust design under uncertainty and the enhancement of practitioner comprehensibility, ensuring that explainability remains a continuous diagnostic thread throughout the surrogate modeling process.}

\subsection{\textcolor{black}{Data Acquisition and Initial Screening}}

\textcolor{black}{While optimization seeks a single, high-performing solution, design exploration maps the broader trade-space to uncover actionable engineering insights~\cite{benard2022shaff}. Whether conducted prior to optimization to screen variables or post-optimization to analyze Pareto fronts, structured exploration prevents over-reliance on limited human intuition. This requires rigorous methodologies spanning experimental design, surrogate training, and subsequent explainability analysis.}
Design of Experiments (DoE) constitutes the foundational step in constructing reliable surrogate models for complex simulations~\cite{kleijnen2015design}. \textcolor{black}{It is therefore important to examine its role within simulation-driven surrogate modeling, particularly as the initial step toward building explainable surrogate models.} 

At its core, DoE seeks to select a set of input configurations that efficiently and uniformly explore the multi‐dimensional parameter space, minimizing gaps and clusters that could lead to blind spots in the learned approximation. \textcolor{black}{Formally, DoE generates the input design $\mathcal{X}$, while the corresponding outputs $\boldsymbol{y}$ are obtained by evaluating the simulator. This process constructs the dataset $\mathcal{D} = \{\mathcal{X}, \boldsymbol{y}\}$ (refer to Section~\ref{sec:surrogate_models}), which serves as the foundation for subsequent surrogate modeling and explainability analysis.} Early approaches to DoE relied on purely random (\textcolor{black}{Monte Carlo}) sampling, which, though unbiased, typically requires a large number of points to achieve satisfactory coverage in high dimensions.  To overcome this limitation, structured space‐filling designs such as Latin Hypercube Sampling (LHS)~\cite{mckay1979comparison} were introduced.  In an LHS design, each dimension is partitioned into equally probable intervals, and exactly one sample is drawn from each interval, ensuring that the marginal distribution of each input variable is uniformly sampled.  Variants of LHS further optimize the sample configuration by maximizing the minimum distance between points or by minimizing pairwise correlations, thereby improving the uniformity of coverage~\cite{jin2005efficient}.

Orthogonal and factorial designs offer an alternative perspective, emphasizing uncorrelated estimates of factor effects.  Two‐level factorial arrays, including Plackett–Burman designs, enable rapid screening of dozens of variables with a minimal number of runs by assuming higher‐order interactions are negligible.  These designs guarantee that each input factor appears at high and low settings in balanced combinations, which simplifies the detection of main‐effect trends.  For simulations where the cost of each run can be prohibitive, such screening methods help identify a subset of influential parameters before committing to more elaborate surrogate construction~\cite{mckay1979comparison}.

Low‐discrepancy sequences, such as Sobol' and Halton, represent another class of space‐filling designs rooted in quasi‐\textcolor{black}{Monte Carlo} theory.  By deterministically generating points with minimal “gaps” in the unit hypercube, these sequences achieve more uniform coverage than purely random draws and often yield faster convergence in integration and surrogate approximation tasks~\cite{l2000variance}.  In practice, practitioners choose between LHS, factorial, and low‐discrepancy methods (generally randomized) based on the anticipated structure of the response surface and the computational budget~\cite{l2016randomized}.

Beyond these classical designs, modern DoE frameworks incorporate adaptive DoE or sequential sampling strategies.  In such schemes, an initial surrogate is trained on a modest set of points, and subsequent samples are chosen to maximize an acquisition criterion, whether it be the surrogate’s predicted uncertainty, an expected improvement metric, or a sensitivity measure derived from global analysis~\cite{chevalier2013fast}.  Known in the ML community as active learning or Bayesian optimization, these approaches focus computational effort on regions where the model is most uncertain or where the gradient of the response suggests complex behavior~\cite{bartoli2025multi}.

\textcolor{black}{In adaptive DoE, the dataset is constructed sequentially by iteratively selecting new input samples based on the current model and available data. At iteration $t$, the next sample is selected by optimizing an acquisition function:
\[
\boldsymbol{x}^{(t+1)} = \arg\max_{\boldsymbol{x} \in \boldsymbol{\Omega}_{\boldsymbol{x}}} \; \mathcal{F}(\boldsymbol{}{x} \mid \mathcal{D}^{(t)}),
\]
where $\mathcal{F}(\cdot)$ is an acquisition function that quantifies the utility of sampling at $\boldsymbol{x}$ (\textit{e.g.}, uncertainty, expected improvement~\cite{mockus1975bayes}), and $\mathcal{D}^{(t)}$ denotes the dataset at iteration $t$. The corresponding output is then obtained from the simulator:
\[
y^{(t+1)} = f\big(\boldsymbol{x}^{(t+1)}\big),
\]
and the dataset is updated as:
\[
\mathcal{D}^{(t+1)} = \mathcal{D}^{(t)} \cup \left\{ \left(\boldsymbol{x}^{(t+1)}, y^{(t+1)} \right) \right\}.
\]}

Screening methods like the Morris elementary effects technique bridge DoE and sensitivity analysis by performing targeted one‐factor‐at‐a‐time (OAT) perturbations across random base trajectories, yielding qualitative rankings of input importance and interaction potential with only a handful of simulations~\cite{morris1991factorial}.  More recent developments exploit sparse regression (\textit{e.g.}, LASSO) and automatic relevance‐determination kernels in GP to perform variable selection implicitly during surrogate fitting~\cite{desboulets2018review,spagnol2019global}.  These techniques allow engineers and data scientists to discover and eliminate negligible factors without separate screening campaigns~\cite{campolongo2007effective}.

\textcolor{black}{Although different communities employ distinct terminology, such as "adaptive DoE" \textit{versus} "active learning" and "screening" \textit{versus} "sensitivity analysis", these concepts differ in scope and methodological emphasis and should not be regarded as interchangeable. For example, adaptive DoE emphasizes sequential sampling strategies for efficient surrogate construction, whereas active learning emphasizes data acquisition strategies for improving ML models under limited labeling budgets.  Nevertheless, the underlying goal remains the same: to achieve an efficient exploration of complex models with minimal computational expense~\cite{lualdi2022exploration}.}


\subsection{\textcolor{black}{Global Exploration of System Behavior}}

In the ML literature, GSA is often limited to the label feature importance, but the underlying aim is identical: to quantify how relevant each input variable is by measuring its contribution to output variability~\cite{van2022comparison}. \textcolor{black}{Understanding and quantifying the global influence of input variables on model outputs has long been regarded as a prerequisite for model interpretability and explainability, both in classical GSA and in modern explainable ML frameworks~\cite{saltelli2008global,friedman2001greedy, Iooss_2015}}. Historically, GSA formalizes how randomness in input variables propagates to a random QoI, yet the same tools apply to deterministic computer experiments by treating inputs as varying over their design domain. In design exploration, the practical task is to identify which design variables affect the QoIs, so that effort and cost can be concentrated on the most influential parameters. \textcolor{black}{To that end, using global sensitivity measures discussed in Section~\ref{sec:global_parameter_screening}, such as the total Sobol indices $S_i^T$ and DGSM $\nu_i$, one can identify the most influential inputs and their interactions. The variables can then be ranked according to their importance, for instance by sorting the total Sobol indices in descending order:
\[
S_{(1)}^T \ge S_{(2)}^T \ge \cdots \ge S_{(m)}^T,
\]
where $S_{(i)}^T$ denotes the $i$-th largest total Sobol index.
} For example, if some design variables have a negligible impact on the objective, designers may fix or simplify them; similarly, when inputs are random, limiting variability only on the most important inputs can produce a more robust system that maintains functionality under uncertainty.

Interpretable models offer another route to assess input importance, but this route has limits. Linear regression coefficients directly reflect sensitivity only when the true relationship is (approximately) linear; adding polynomial or interaction terms increases nonlinearity and complicates interpretation. If first-order terms dominate, their coefficients may still be useful for GSA, but complex symbolic models (\textit{e.g.}, obtained via genetic programming) are often hard to analyse structurally~\cite{disdier2025predicting}. Computing formal GSA metrics is therefore recommended even for interpretable models, as they provide a consistent and comparable quantitative basis for ranking inputs across the entire input space, revealing nonlinear effects and interactions.



\subsection{\textcolor{black}{Interpreting Input–Output Structure and Physical Trends}}

\textcolor{black}{Practitioners want to know how hard the design or exploration problem is. Depending on the context,  they may be interested in  \emph{optimization difficulty}: presence of multiple local optima, flat regions, or steep ridges, \emph{representational complexity}: strong nonlinearity, discontinuities, mixed variables,  \emph{statistical complexities}: correlated inputs, dependant features, high-order interactions, or \emph{computational complexity}: long runtimes or need for heavy replication for stochastic simulators. Analysis of input–output mappings, typically performed on the simulator or on surrogates trained on simulator outputs, helps to address these issues. Whenever possible, simulator-level findings should be contrasted with empirical data or discrepancy models to assess transfer to the real system \textcolor{black}{or to the ground truth.}}  

Design exploration often follows optimization: once a set of candidate designs is obtained, explanations and pre-optimization analyses can be combined to relate the design space to objectives~\cite{palar2025partial}. Studying alternative objectives or constraints yields different designs to compare; likewise, contrasting local optima helps reveal distinct design modes, a phenomenon observed in aerodynamic wing design~\cite{poole2018global} and studied with physical reasoning~\cite{bons2019multimodality}. Note that demonstrating true multimodality typically requires gradient information (hence gradient-based optimizers). 

Solutions from multi-objective optimization might conflict with each other and necessitate the need for exploring Pareto optimal designs~\cite{deb2016multi}. Typically, the main objective is to explore why certain combinations of design variables lead to the trade-off between objectives. The physics underlying the Pareto optimal solution is usually examined to gain insights into the fundamental mechanisms that enhance performance. Further, it is also of interest to understand the mechanics of the trade-off (\textit{e.g.}, the convexity of the Pareto front). \textcolor{black}{For example, a multi-objective design procedure for thruster design was performed to investigate the impact of design variables on the three objectives,  thrust, total efficiency, and specific impulse~\cite{yeo2022multi}.} 
\textcolor{black}{To provide a concrete illustrative example of how XAI operationalizes this extraction of knowledge, consider a multi-objective design problem with two conflicting objectives, $f_1(\boldsymbol{x})$ (e.g., maximize aerodynamic lift) and $f_2(\boldsymbol{x})$ (e.g., minimize aerodynamic drag)~\cite{grapin2022regularized}. The optimization yields a Pareto optimal set $\mathcal{P}$, defined mathematically such that for any design $\boldsymbol{x}^* \in \mathcal{P}$, there exists no other design $\boldsymbol{x}$ where $f_i(\boldsymbol{x}) \leq f_i(\boldsymbol{x}^*)$ for all $i$, with strictly inequality for at least one objective.}

\textcolor{black}{While the Pareto front graphically shows \textit{that} a trade-off exists, it does not explain the underlying physical mechanism. By applying a local XAI method like SHAP to the surrogate models $\hat{f}_1$ and $\hat{f}_2$, we can mathematically decompose the trade-off~\cite{palar2024multi}. For a specific optimal design $\boldsymbol{x}^*$, the predicted performance for both objectives is expanded as the sum of their baseline expected values ($\phi_{0,1}$ and $\phi_{0,2}$) and the local feature contributions:}
\begin{equation}
\begin{split}
    & \hat{f}_1(\boldsymbol{x}^*) = \phi_{0,1} + \sum_{j=1}^m \phi_{j,1}(\boldsymbol{x}^*), \\
    & \hat{f}_2(\boldsymbol{x}^*) = \phi_{0,2} + \sum_{j=1}^m \phi_{j,2}(\boldsymbol{x}^*)
\end{split}
\end{equation}
\textcolor{black}{By analyzing the correlation between the SHAP values $\phi_{j,1}$ and $\phi_{j,2}$ across the set $\mathcal{P}$, practitioners can quantitatively classify design variables. In the case where two objectives are to be minimized (or maximized), if a variable $x_k$ exhibits $\phi_{k,1}(\boldsymbol{x}^*) > 0$ and $\phi_{k,2}(\boldsymbol{x}^*) < 0$, it mathematically demonstrates a conflicting feature driving the Pareto trade-off. Conversely, if both SHAP values improve simultaneously, $x_k$ is a synergistic variable that should be universally maximized. This mathematical decomposition transitions the analysis from purely observing a trade-off curve to understanding the exact parametric sensitivities dictating the design limits.}
    
A simple interpretable model, such as polynomial regression, is particularly useful if the input-output relationship is relatively simple and can be accurately modeled. The sign of the polynomial coefficients indicates the direction of how the change of a particular input variable is associated with the change of the QoI~\cite{hastie2009elements}, where a positive sign signifies that increasing the input variable tends to coincide with an increase in the QoI (and vice versa). However, the interpretation becomes difficult if the problem is highly nonlinear or features strong interaction between input variables. Furthermore, it is not trivial to identify whether the model exhibits multiple local optima by simply looking at the coefficients.

\textcolor{black}{Indeed, understanding the input-output relationship is the central goal of explainable ML, often achieved through targeted visualization~\cite{yin2024data}. Since surrogate models approximate complex simulation responses, these explainability techniques uncover how parameter variations drive predicted quantities of interest, enabling not only faster evaluations but also deeper insights into the underlying physical or design mechanisms~\cite{angione2022}.} 
Leveraging the model-agnostic nature of these methods, practitioners often deploy PDP and ICE because of their highly intuitive, graphical representations. They provide straightforward visualizations of how changing feature values affect the output, making overarching trends easy to grasp without extensive statistical expertise. 

Conversely, the primary advantage of SHAP in this exploratory workflow is its capacity for precise local attribution. By illustrating exactly how features shift an individual prediction away from a baseline design, SHAP allows engineers to investigate isolated, instance-level model behaviors. This localized diagnostic power has driven its recent application in complex physics domains, including evaluating closure coefficients of the Shear Stress Transport $k-\omega$ turbulence model for ground vehicle aerodynamics~\cite{bounds2024enhancing}, investigating interference effects between high-rise buildings~\cite{yan2024explainable}, and analyzing aeroelasticity~\cite{wan2023uncertainty}.

ASM is an alternative method to investigate the input-output relationship by finding the most influential directions (\textit{i.e.}, active subspace) within the high-dimensional input space that have the most significant impact on the QoI. If the relationship possesses strong one- or two-dimensional active subspaces, projection into these low-dimensional subspaces can provide information regarding nonlinearity~\cite{lukaczyk2014active, seshadri2018turbomachinery}.   Compared to methods like PDP and SHAP, discerning the individual impact of each input using ASM is challenging and non-trivial because ASM does not offer explicit insights into how individual features influence the model's predictions on a case-by-case basis.

Originating from the discipline of evolutionary computation, Fitness Landscape Analysis (FLA) aims to understand and visualize the relationship between the fitness of individuals (solutions) in a population and the structure of the problem space~\cite{zou2022survey, malan2021survey}. FLA itself is quite broad, as it not only emphasises understanding the input-output relationship. Further, FLA is a distinct discipline and not fully related to ML. The FLA framework itself was first developed to design a better strategy for evolutionary computation-based methods.

The input-output relationship also aims to quantify the interaction between inputs and how they affect the QoI. Interaction analysis focuses on understanding how combinations of input parameter variations can lead to non-linear effects or interactions that may not be apparent when considering each parameter in isolation~\cite{morris1991factorial}. Strong interactions might complicate the analysis and amplify the complexity of the problems. Therefore, it is important to unravel the dynamics of the interactions whenever they are present. The absence of interactions suggests that the relationship between input and output can be straightforwardly expressed as the sum of individual effects attributed to main variables. However, it is important not to confuse nonlinearity with interactions. It is possible to have a nonlinear function when an individual variable non-linearly influences the output. 

The analysis of interactions typically involves two main goals: measuring the magnitude of the interaction and gaining insights into the mechanics of how variables interact with each other, often facilitated by visualization~\cite{goldstein2015peeking}. GSA techniques are often capable of quantifying the interactions, \textit{e.g.}, via second- and higher-order interactions as in Sobol' indices. \textcolor{black}{While variance-based GSA methods quantify the strength of main and interaction effects, additional analysis is generally required to characterize the detailed functional form of those interactions~\cite{palar2024multi}. Advanced approaches, such as functional ANOVA decompositions~\cite{hooker2004discovering} or Shapley-based measures, can offer partial structural information; however, these methods are not primarily designed as tools for visualizing interaction behavior. In contrast, some interpretable models can explicitly represent interactions through their mathematical expressions, thereby offering more direct insight into how input variables jointly influence the output.} In the case of polynomial regression, one can analyze interactions by focusing on the corresponding coefficients and interaction terms. The simplicity of low-order polynomial terms makes them mentally comprehensible, as is the case with the interaction terms. However, it becomes challenging to imagine the interaction dynamics if the higher-order polynomials are strong, suggesting a complex interaction between variables. It is worth noting that transparent models do not necessarily guarantee ease of interpretation~\cite{zhang2025challenges}, particularly when the expressions are complex. For instance, expressions generated by genetic programming can become convoluted and incomprehensible without proper management of complexity~\cite{luke2006comparison}. Hence, visual representations of interactions remain valuable for enhancing understanding from the perspective of human users.

\subsection{\textcolor{black}{Dimensionality Reduction for Efficiency and Explainability}}
Once the primary features are identified and physical trends are verified, engineers often compress the feature space to accelerate downstream tasks like design optimization. Dimensionality reduction can serve as a means to not just reduce the complexity~\cite{bibal2020interpretability} but also aid in understanding the problem. \textcolor{black}{Mathematically, we define a mapping from a high-dimensional input space, assumed to be real-valued, to a lower-dimensional representation as~\cite{tenenbaum2000global}:
\[
\Phi : \mathbb{R}^m \rightarrow \mathbb{R}^a, \quad a \ll m,
\]
such that for an input vector $\boldsymbol{x} \in \mathbb{R}^m$, the reduced representation is given by
\[
\boldsymbol{\eta} = \Phi(\boldsymbol{x}), \quad \boldsymbol{\eta} \in \mathbb{R}^a.
\]
The reduced vector $\boldsymbol{\eta}$ and the mapping $\Phi$ can be obtained through various techniques (\textit{e.g.}, ASM as discussed in Section~\ref{sec:active_subspace_method}), in which some notable examples are discussed below.}

Unsupervised techniques such as proper orthogonal decomposition~\cite{berkooz1993proper} and dynamic mode decomposition~\cite{schmid2010dynamic} are often used to reduce the input dimensionality and understand the flow behavior in fluid mechanics. Such decomposition of fields into dominant modes provides a means to ease understanding. Although not from the field of engineering, Shapley values have been used to explain clusters of data obtained from dimensionality reduction techniques~\cite{marcilio2021explaining}. 

The supervised counterparts are more often used for improving the surrogate model accuracy (\textit{e.g.}, ASM, sliced inverse regression~\cite{li1991sliced}, manifold learning~\cite{tenenbaum2000global}) to accelerate optimization, uncertainty quantification~\cite{nguyen2024uncertainty}, and reliability analysis~\cite{deng2021slope}. Dimensionality reduction can be applied not just to the input but also to the output space. In this regard, the output can be in the form of a scalar field, such as a pressure and temperature field. Typically, the objective is to investigate the impact of input variables on the resulting field and build better surrogate models. For instance, there is a recent interest in the use of manifold learning for improving the accuracy of surrogate models such as polynomial chaos expansion (PCE)~\cite{kontolati2022manifold} and GP~\cite{giovanis2020data}. Besides, supervised dimensionality reduction methods can also be used to understand the behavior of the input-output relationship. An example of this is the application of ASM to detect the subspace of the input variables corresponding to the maximum variance of the output (\textit{e.g.}, see one recent application in supersonic flame stabilization~\cite{wei2023exploiting}). If the function exhibits low-dimensional subspaces, projection onto these subspaces can reveal the intrinsic low-dimensional directions and the characteristics of the function (\textit{e.g.}, linearity, unimodality). In other words, the low-dimensional features are used to explain the behavior of the function. Another example, unsupervised learning techniques such as kernel PCA and linear or nonlinear embedding have been applied to reduce the dimensionality for faster optimization (\textit{e.g.}, see~\cite{gaudrie2020modeling} and~\cite{binois2022survey}).


In a sense, GSA also explains the characteristics of the problem since it paved the way for discovering the importance of input variables. Non-important variables can be neglected to reduce the problem's dimensionality, and the surrogate model can be reconstructed with the truncated inputs. If there are two significant variables, a scatter plot of these two variables can give a mental depiction of the problem's characteristics, in which the impact of variables appears as "noise" on this three-dimensional scatter plot. Methods such as PDP, ICE, and SHAP do not provide straightforward mechanisms to reduce dimensionality, unlike \textit{e.g.}, ASM. That is, although they provide information on input importance, they are not equipped with inherent mechanisms to reduce dimensionality. Instead, the derived GSA metrics can be used to post-process the problem by discarding insignificant input variables. For instance, SHAP feature importance has been used for such a purpose in the context of standard ML, where the reconstructed model yields higher accuracy~\cite{kumar2020dimensionality}. However, this comes with the risk of underestimating the true complexity of the problem.

\subsection{\textcolor{black}{Robust Design Under Uncertainty}}
In real-world engineering design, the presence of uncertainties is unavoidable, and it is crucial to consider and address them. \textcolor{black}{The role of simulations, often approximated by surrogate models to enable extensive sampling~\cite{sudret2017surrogate}, is important for assessing how different realizations of uncertain parameters propagate to system performance.} Uncertainty can be at least classified into types, namely, aleatoric uncertainty, which is due to the inherent randomness of a phenomenon, and epistemic uncertainty, which arises from a lack of knowledge or data~\cite{hullermeier2021aleatoric, der2009aleatory}. Aleatoric uncertainty is often described in the language of probability and necessitates the definition of the input probability density function. On the other hand, epistemic uncertainty is often characterized by non-probabilistic measures such as Dempster-Shafer evidence theory~\cite{dempster2008upper}. Irrespective of the categories, the primary goal is to describe the uncertainty in the output of interest stemming from uncertainties in the input.

The notion of uncertainty suggests the existence of multiple possible input values or certain but unknown values, prompting the need for experimentation with various values. For instance, representing the random input variables $\boldsymbol{\xi}=\{\xi_{1},\xi_{2},\ldots,\xi_{k}\}^{T}$ using a probability distribution $\rho(\boldsymbol{\xi})$ implies that the output can also be represented with a probability distribution $\rho(f(\boldsymbol{\xi}))$, in which quantities such as expected value and variance can be extracted. The most conventional method to perform such a task is \textcolor{black}{Monte Carlo} simulation, which involves generating numerous random samples of the input variables and computing the corresponding outputs. Clearly, this becomes impractical if one query is computationally expensive. A surrogate model $\hat{f}(\boldsymbol{\xi})$ can then be employed to act as the proxy of $f(\boldsymbol{\xi})$ to reduce the number of queries for accurate estimation of the output's uncertainty~\cite{sudret2017surrogate}. Besides, quantifying the individual impact of uncertain variables is important for variable screening. The field of GSA  aims to provide answers to this problem through methods such as variance-based decomposition~\cite{Iooss_2015}. 

Comprehending the influence of uncertainty on design can be enhanced by gaining a deeper understanding of the impact of $\boldsymbol{\xi}$ on $f(\boldsymbol{\xi})$. To that end, the model $\hat{f}(\boldsymbol{\xi})$ can be passed through an explainability module. Similar to design exploration, one can then enquire deeper into how specifically the random inputs affect the output of interest. As an illustration, the nonlinearity in the bending moment of a wind turbine influenced by random wind conditions can be identified through the application of SHAP~\cite{palar2023enhancing}. In this manner, it becomes feasible to pinpoint the factors causing the condition in relation to the random input variables that result in the maximum bending moment. A recent application involves the use of SHAP and PDP for investigating the stochastic re-entry trajectory of a capsule, revealing complex discontinuity~\cite{palar2024global}. Another study investigates SHAP for explaining the robust design of electric vehicle charging stations~\cite{silva2025carbon}. Further, combinations of XGBoost and SHAP have been utilised to investigate the impact of uncertain variables on aeroengine fan performance~\cite{cheng2025aerodynamic}. The ASM has also been applied to various designs under uncertainty and uncertainty analysis applications, in which the goal is to detect the presence of active subspaces and relate it to the input variables, \textit{e.g.}, for supersonic flame~\cite{wang2021active}, turbomachinery~\cite{song2024sensitivity}, and wind energy~\cite{panda2021multi}.

The potential of explainable ML in addressing design uncertainty has been recognized; however, there is still a limited amount of research dedicated to exploring this specific area. Utilizing explainable ML for epistemic uncertainty has the potential to enhance comprehension regarding strategies for acquiring additional knowledge and, consequently, further reducing epistemic uncertainty. One potential application of explainable ML is in tackling issues involving mixed uncertainty, where both aleatoric and epistemic uncertainties coexist. This is because analyzing mixed uncertainty can be inherently challenging due to the distinct nature of these two types of uncertainties~\cite{shah2015mixed}.

\subsection{\textcolor{black}{Enhancing Practitioner Comprehensibility}}
\label{sec:human_comprehensibility}

The most important goal of XAI for engineering with simulations is to provide understanding to users (both technical and non‑technical) to grasp how models arrive at their outcomes~\cite{arnejo2025automatic}. In contrast to the "pure-prediction" machine viewpoint,  the emphasis now shifts to the significance of human comprehensibility since the primary goal is understanding. Various explainability methods provide different levels of comprehensibility and can be achieved through three means, namely, post-hoc scenario generation, mathematical expression, or visuals. 

\bigbreak

Mathematical representations reveal the precise input–output relationships through explicit formulas or rule sets. Examples include power‑law or logarithmic transformations, generalized additive models, decision‑tree paths, polynomial regressions, sparse regressions, and symbolic expressions derived via genetic programming or gene expression programming~\cite{doumard2022comparative,ferreira2001gene}. These formats make it possible to quantify each variable’s influence via coefficients or rule predicates. Methods like Sparse Identification of Nonlinear Dynamics (SINDy) produce compact, data‑driven models of complex systems~\cite{brunton2016discovering}, though they require careful term pruning to maintain readability. 

Empirical models can be useful for further applications, such as preliminary design~\cite{forrester2008engineering}. Empirical models offer quick estimates and predictions during the initial design stages when the detailed analysis might be time-consuming, eventually enabling engineers to explore multiple design options efficiently. Such models are often transparent, providing a quick understanding of the relationship between the input variables and the output. The transparent nature of such empirical models also gives a sense of trustworthiness since one can see the model's expression.

Nonetheless, employing explainability methods for visual analysis can assist users in gaining a deeper understanding of the relationship, especially when interpreting it solely based on the expression proves to be challenging. \textcolor{black}{Visualization is often an essential component of explainability, as it aims to distill knowledge embedded in data and models through informative plots and diagrams~\cite{yin2024data}. In this regard, visualization not only aids interpretation but also serves as a practical tool for model refinement~\cite{choo2018visual}. Widely used explainability methods such as PDP/ICE, SHAP, and ALE provide systematic means for visualizing model behavior, which in turn facilitate knowledge discovery. In scientific and engineering contexts, this is particularly valuable, as it allows one to relate model findings back to physical principles, validate consistency with domain knowledge, and potentially generate new hypotheses about the underlying phenomena~\cite{roscher2020explainable}.} Visual explanations use feature‑attribution techniques to map inputs to outputs in a user‑friendly manner. ICE plots, PDP, and SHAP summary plots depict how each variable affects predictions, both locally and globally~\cite{knab2025lime,idrissi2025beyond}.  

\bigbreak

\textcolor{black}{Moving beyond passive explanations, explanations must act as interactive dialogue interfaces. This allows decision-makers to validate, refine, or contest outputs, thus closing a feedback loop that improves trustworthiness~\cite{ferreira2020people, amershi2019guidelines}. However, recent works highlight a persistent gap in stakeholder engagement after model deployment~\cite{arnejo2025communicating}. To address this and avoid bias amplification or over-reliance~\cite{hall2022still}, practitioners should maintain interactive dashboards that evolve with the model, embed real‑time demonstrations of emergent behaviors, and institute standardized evaluation protocols to collect ongoing feedback~\cite{cartwright2016communicating}.}




\section{Open Challenges and Perspectives}
\label{sec:challenges}

The surrogate–XAI workflow surveyed in the previous sections raises a set of recurring technical and epistemic challenges that we present hereinafter. These challenges are not isolated problems but cross-cutting tensions that influence method choice, validation protocols, and how results are communicated to stakeholders. \textcolor{black}{When explainability is combined with surrogate modeling, these challenges are further compounded by surrogate-specific factors such as approximation error.}

\textcolor{black}{The open challenges discussed in this section can be broadly organized into two complementary categories: (1) \emph{Methodological and technical challenges} and (2) \emph{human-centered challenges}. These perspectives highlight that progress in surrogate–XAI workflows demands both methodological advancement and a socio-technical framework that incorporates human understanding and responsible use. In what follows, we examine these topics in turn, connecting each to the methods and diagnostics reviewed earlier and highlighting open problems where further methodological work or empirical validation is required. Each subsection outlines the core problem, summarizes existing approaches, and indicates practical recommendations or research directions for integrating these concerns into a simulation-driven, co-design workflow.}



\subsection{Methodological and technical challenges}

\textcolor{black}{The first concern, \emph{methodological and technical challenges}, encompasses the reframing of explanations as a new design space, the integration of prior knowledge into surrogate models, the extension of explainability to dynamical systems, reliability-based modeling, and the treatment of mixed, correlated, and high-dimensional variables in the context of explainable surrogate model. This category also includes practical questions such as how to choose appropriate explainability methods and the needs for the uncertainty quantification associated with explanation metrics.}

\subsubsection{Explanations as a New Design Space}

Explainability has become central to high-stakes decision making: XAI closes the gap between model outputs and human understanding, providing auditability and justification for algorithmic decisions~\cite{barredo2019lies}. Framing explanations into new data or data features turns them into a new design space from which automatic model recommendations can be derived: explanations can be encoded, queried, and compared, enabling meta-models or recommender systems to suggest models, hyperparameters, or corrective actions that better satisfy human requirements~\cite{lipton2018mythos}.
Collectively, these works illustrate that explanations, when leveraged as structured data, can transform not only \emph{what} decisions are made, but also \emph{how} and \emph{why} they are made.   As local, post-hoc explainability methods transform each instance of a dataset into an \emph{explanation vector}, a feature-wise attribution of the model’s prediction (\textit{e.g.}, LIME, SHAP, CIU). Rather than treating these vectors solely as interpretive artifacts, we propose regarding them as a \emph{new design space} for downstream tasks. In this view, explanations become additional data features, capturing model behavior in a way that complements the original input representation and unlocks novel analytic and decision-making workflows. 

\bigbreak

First, this explanation space can be exploited for \textbf{model selection}. Traditional AutoML approaches optimize accuracy and resource usage; by incorporating \emph{explanation metrics}, such as stability, consistency, and sparsity, into the selection criterion, practitioners can choose models whose decision logic aligns with domain expectations and regulatory requirements. For instance, integrating explanation stability into an industrial AutoML pipeline improves user trust and system reliability~\cite{garouani2022towards1}. Also, AutoXAI frameworks can automatically recommend both predictive models and explanation methods tailored to stakeholder needs~\cite{cugny2022autoxai}, and models selected with explanation-aware criteria achieve competitive accuracy with significantly higher \textcolor{black}{explainability} in biomedical tasks~\cite{chkifa2025autism,wang2023explanations, Ben_Yahia_2025}.

Second, explanations enrich \textbf{feature selection}. Conventional approaches focus on predictive power (\textit{e.g.}, mutual information, recursive elimination) without regard for the interpretability of resulting models~\cite{brown2012conditional}. By contrast, it can be shown that selecting features whose attributions are both stable and sparsely distributed yields models that maintain accuracy while producing more coherent and actionable explanations~\cite{excoffier2022local,escriva2023make}. This “explanation-aware” feature pruning is especially valuable in high-stakes settings, such as healthcare, where domain experts demand not only performance but also clarity in how features drive decisions.

Third, clustering and selection of \textbf{representative instances} can be driven by explanation vectors. Every instance’s explanation encodes its unique decision rationale; grouping these vectors reveals prototypical decision patterns and outliers. Many coalitional strategies can be used for summarizing explanation clusters~\cite{ferrettini2022coalitional}, and these concepts have been applied to identify patient archetypes in clinical datasets, enabling targeted follow-up analyses and bias detection~\cite{escriva2023make}. In operational settings, this facilitates sampling of instances for model auditing, user review, or active learning.

Finally, explanation vectors support \textbf{exploratory data analysis}. By treating explanations as features, one can uncover latent structures not visible in the raw feature space. For example, clustering in explanation space has revealed non-linear interactions among biomarkers and enabled rule extraction for clinically meaningful subgroups, insights that standard unsupervised techniques on raw data failed to surface~\cite{escriva2023make,aligon2024vers}. This approach thus provides a bridge between black-box prediction and transparent data exploration.

Underpinning these applications is the recognition that explanations are not merely post-hoc agenda but constitute a rich \textcolor{black}{explanatory} representation. By embracing explanation space as a design substrate, we open new avenues for \textbf{model recommendation}, feature engineering, active sampling, and exploratory analysis, which is pivotal for building trustworthy, human-centered AI systems.

\subsubsection{Injecting Prior Information}
Injecting prior knowledge into the ML model can potentially improve the prediction quality~\cite{karniadakis2021physics}. Such prior knowledge can be in the form of physical constraints, empirical models, or governing equations. For instance, incorporating governing equations into neural networks has been demonstrated to improve predictions by aligning them more closely with the principles of physical laws~\cite{cuomo2022scientific, cai2021physics, karniadakis2021physics}. In the context of design exploration, constraints such as non-negativity or monotonicity, which are derived from prior knowledge, can be incorporated into the ML model. Several physical properties, such as stress, are positive quantities, and it might be worth incorporating such constraints when modeling stress. In GP, several approaches to inject prior information include constraining ML kernels until giving physics-based mean/trend function, and physical covariance~\cite{cross2024spectrum}. Other models have also been extended so as to incorporate prior knowledge (\textit{e.g.}, physics-informed neural network~\cite{raissi2019physics} and PCE~\cite{sharma2024physics}). Regardless, one must be careful in incorporating prior knowledge. One should then always perform procedures such as cross-validation to ensure potential gain in accuracy.

At this point, it is important to reiterate the goal of design exploration, which is to extract knowledge. If the knowledge is already known beforehand, would injecting prior knowledge into the model reveal more insight? The answer might depend on the case. Applying explainability techniques to ML models that are enriched by prior information can potentially still unravel new knowledge. In particular, the integration of knowledge into explainable ML has the potential to improve human understanding of the model and incorporate particular objectives or requirements into the learning process~\cite{beckh2021explainable}. The inclusion of prior knowledge can further reinforce or readjust the existing knowledge while also producing new knowledge. To illustrate, imagine a scenario where the function expression is precisely known but challenging for humans to grasp, meaning that we already have knowledge of the relationship itself. Explainable ML can assist in enhancing human understanding by offering visualizations that depict the relationships within the function.


\subsubsection{Explainability for Dynamical Systems}
Most explainability techniques are applied under the assumption that the data is static, that is, the response does not vary with time. This way, the application of explainability methods becomes relatively straightforward, as they aim to reveal the influence of input variables on a time-independent dataset. However, the use of explainability methods for time-series datasets in scientific and engineering contexts remains relatively uncommon~\cite{lee2015complexities, saluja2021towards}. \textcolor{black}{When surrogate models are used to approximate dynamical systems (\textit{e.g.}, see~\cite{zhao2023surrogate}), explainability faces additional challenges beyond those encountered in static settings. Explanations are obtained for an approximate model whose errors may accumulate over time, making the interpretation of long-term behavior particularly delicate.}

It is important to recognize that the term "time-series data" can refer to different scenarios. In time-series dataset, each observation is given by $x = \{x(t_1), x(t_2), \ldots, x(t_{n_t}) \} \in \mathbb{R}^{n_t \times d}$. The case with $d=1$ and $d>1$ refers to the univariate and multivariate time series data, respectively. In this regard, the response itself can be static (\textit{e.g.}, classification~\cite{theissler2022explainable}) or dynamic (\textit{e.g.}, forecasting~\cite{arsenault2024survey} or the time response of the QoI). The term time-series data may also refer to scenarios where the input variables remain static, while the output of interest is now a time response. Such cases are common in surrogate modeling of dynamical systems (\textit{e.g.},~\cite{zhao2023surrogate, bhattacharyya2022uncertainty, chakraborty2021role, schar2024emulating}), in which each $y^{(i)}$ becomes a function of time, representing a temporal signal associated with the $i$-th observation. A complex data structure arises when both the input (which might correspond to parameters and initial conditions) and the output are time-dependent, \textit{i.e.}, a time-series to time-series (or sequence-to-sequence) problem. The complexity further increases when both input and output exhibit spatio-temporal characteristics, involving variations across both space and time. A simple example of such a case is when the system is subjected to a time-varying boundary condition, and the corresponding output is a spatial distribution of the system’s response evolving over time, such as a transient temperature field or pressure distribution.

The main challenge in the context of dynamical systems is how to generate meaningful explanations for time-dependent responses. A straightforward approach is to summarize the time response (such as by taking the average) and treat this summary as the output. However, this effectively reduces the problem to a scalar-output setting. For example, the influence of input features on the Narendra-Li system has been investigated by integrating genetic programming with SHAP~\cite{calapristi2024interpretability}. In this approach, genetic programming was first proposed to identify the underlying dynamical equations, followed by the application of SHAP to provide global explainability of the model. Another possible approach is to use techniques like POD to represent the time response through a set of basis functions and their corresponding coefficients. An explainability model can then be constructed on these coefficients to gain insight into the system behavior (\textit{e.g.}, via GP~\cite{bhattacharyya2022uncertainty}). Other alternatives, such as autoencoders, can also be used, although this might complicate interpretations. It is important to note that the governing differential equations of the dynamical system may already be known; however, if they need to be identified from data, techniques like parameterised SINDy can be employed to infer them~\cite{brunton2016discovering, lemus2025multi}. Recall that the goal here is to investigate the impact of the parameters on the specific time response. 

\textcolor{black}{These considerations highlight that explainability for dynamical surrogate models is inherently conditional on temporal horizon, surrogate fidelity, and modeling assumptions, reinforcing the need to couple explainability with uncertainty assessment and validation. We believe that explainability for dynamical systems is an important research agenda since surrogate models are increasingly employed to approximate complex time-dependent simulations~\cite{schar2024emulating, scharmNARX2026}.
}

\subsubsection{Reliability-based Modeling}

Reliability is a central concept in simulation-driven decision support, even though it refers to different concepts depending on the context.
In engineering or critical system modeling, such as an ABM for epidemiology or a complex aerospace system, reliability refers to the probability that the system performs its intended function under specified conditions, accounting for uncertainties, rare events, and operational variability~\cite{raheja2012design}. 
In contrast, in a surrogate ML model, reliability refers to the model’s ability to provide consistent and accurate predictions, including previously unseen or slightly shifted inputs, often accompanied by quantified uncertainty (\textit{e.g.}, Bayesian models or ensembles). Both notions of reliability are essential: system-level reliability ensures that the underlying system behaves robustly, while surrogate-level reliability ensures that ML approximations can be trusted. The latter is particularly critical for decision-making and uncertainty propagation, guiding practitioners towards reliable and actionable insights~\cite{balasubramanian2014conformal}. \textcolor{black}{In this context, explainability plays a complementary role by clarifying the mechanisms through which surrogate predictions contribute to reliability assessments. Beyond numerical accuracy, reliable decision support requires understanding why a system is predicted to approach or violate a limit-state, making explainability a promising tool for diagnosing model behavior in safety-critical regions.}

We outline a three-level reliability analysis framework: (i) the \emph{underlying system model}, which captures the physics or agent interactions and defines the true reliability landscape; (ii) the \emph{surrogate ML model}, which emulates the system efficiently while providing predictive uncertainty; and (iii) the \emph{XAI technique}, which explains model outputs in a way that is actionable and faithful to both the system and surrogate behavior. 

At each level, reliability considerations inform model construction, validation, and interpretation: system-level reliability guides design and policy evaluation, surrogate-level reliability ensures trustworthy predictions under limited data or computational budgets, and XAI-level reliability guarantees that explanations are robust, understandable, and useful for stakeholders~\cite{balasubramanian2014conformal}.

In goal-oriented XAI, explanations are instruments aligned with the decision tasks they support (\textit{e.g.}, calibration, inspection planning, certification). In reliability contexts, objectives are to clarify \emph{why} a design lies near or inside the failure domain and \emph{how} minimal, actionable changes would reduce failure probability. Explanation primitives such as counterfactuals, contrastive rules, or more generally, global \textit{versus} local attributions should maximize actionability and fidelity in failure-relevant regions, with evaluation criteria including fidelity, sufficiency for the decision, robustness, and quantified uncertainty~\cite{slack2021reliable}.

Rare-event and low-probability regimes pose interpretability \textcolor{black}{and explainability} challenges: failure-critical regions are typically data-scarce and dominated by epistemic uncertainty. Explanation strategies must be uncertainty-aware: local attributions (\textit{e.g.}, Shapley-like factors) should include confidence bounds from surrogate ensembles or uncertainty-aware models. Counterfactuals are valuable for proposing small, interpretable modifications to inputs that move the system out of the failure region, but generating trustworthy counterfactuals in extreme-tail regimes requires constrained sampling and calibrated surrogates~\cite{maalej2025counterfactual}.

Current strategies include: (i) uncertainty-aware surrogates (GP, ensembles, quantile regressors) producing predictive distributions~\cite{espoeys2024overview}; (ii) multi-fidelity and transfer learning leveraging related conditions or simplified simulators~\cite{bocquet2025control}; (iii) decision-aware active-learning acquisition functions biased toward the limit-state~\cite{moran2025balancing}; and (iv) hybrid explanation pipelines combining global interpretable surrogates with precise local attributions (PDP/ICE, conditional Shapley)~\cite{mehdiyev2023interpretable}.

Reliability-Based Design and Optimization (RBDO) extends deterministic design by embedding probabilistic performance constraints, ensuring the probability of exceeding a limit-state remains below a threshold. Canonical RBDO couples an outer optimization loop with an inner reliability analysis~\cite{espoeys2024multi}. FORM/SORM approximate the limit-state at the most probable point in standard normal space; efficient for mild nonlinearity, they lose accuracy for highly curved surfaces or very low failure probabilities~\cite{bourinet2018reliability}. Variance-reduction techniques such as importance sampling and subset simulation are widely used for rare-event estimation~\cite{tabandeh2022review,dubourg2011reliability}, achieving orders-of-magnitude efficiency gains over naive \textcolor{black}{Monte Carlo}~\cite{morio2015estimation}.

Surrogates, particularly GP, provide predictive means and uncertainty estimates. Adaptive Kriging \textcolor{black}{Monte Carlo} Simulation iteratively refines a GP near the limit-state for efficient failure-probability estimation~\cite{echard2011ak,menz2020variance}. Surrogate-based RBDO extends to multi-physics and multi-fidelity frameworks, reducing costly simulations while maintaining accuracy~\cite{moustapha2018comparative,espoeys2024overview}. Additional methods include the cross-entropy method~\cite{el2021improvement}, quantile-based formulations~\cite{moustapha2016quantile}, and high-dimensional transfer learning for tail estimation~\cite{sabourin2021extreme}.

A typical workflow: define the performance/limit-state and objectives; characterize input uncertainties (aleatory \textit{versus} epistemic); propagate uncertainties through the surrogate model, and estimate reliability metrics via FORM/SORM, \textcolor{black}{Monte Carlo}, subset simulation, or surrogate methods; perform inverse analysis or sensitivity studies (\textit{e.g.}, Bayesian calibration, Sobol’ indices, Shapley factors) to refine inputs or models before iterating~\cite{idier2013bayesian,menz2020variance,gautier2025modelling}.

\color{black}
Despite these established workflows, several key challenges remain that require dedicated research. First, on the algorithmic front, establishing \textbf{joint convergence diagnostics} for both sampling and surrogate errors is critical~\cite{menz2020variance,owhadi2013optimal}. Practitioners must monitor these simultaneously to ensure that estimated failure probabilities are not artificially skewed by model bias or Monte Carlo variance. Closely related is the \textbf{rigorous propagation of epistemic uncertainty}~\cite{ha2017time,seoni2023application, nannapaneni2016reliability}. Bounding the parametric and model-form uncertainties embedded in the surrogate model remains mathematically demanding in rare-event regimes where training data is inherently scarce.

Second, computational scalability dictates a need for \textbf{scalable dimension reduction}~\cite{el2021improvement} and \textbf{multi-fidelity integration}~\cite{yi2021active}. As engineering systems grow in complexity, efficiently projecting high-dimensional parameter spaces into lower-dimensional manifolds, without distorting the topology of the limit-state boundary, is vital. Furthermore, sampling strategies must transition to \textbf{decision-aware acquisition functions}. Rather than aiming for global surrogate accuracy, active learning criteria must be strictly biased toward exploring failure boundaries to maximize sampling efficiency~\cite{moustapha2022active}.

Finally, the deployment of these models introduces significant human and operational hurdles. Generating \textbf{robust, uncertainty-aware counterfactuals}~\cite{maalej2025counterfactual} is essential so that explanations do not suggest physically impossible or highly uncertain design modifications to escape a failure domain. This complexity is compounded in \textbf{time-dependent reliability and life-cycle modeling}, where component degradation and stochastic environmental loads shift the limit-state over time. Most importantly, bridging the gap between advanced XAI metrics and \textbf{human-centered evaluation and regulatory compliance}~\cite{mehdiyev2023interpretable} remains the final barrier, as explanations in safety-critical engineering must be legally and operationally auditable, not just mathematically sound.

Addressing these challenges will enable RBDO workflows that are computationally efficient, transparent, and actionable, combining robust uncertainty quantification, decision-aware sampling, calibrated surrogates, and human-centered evaluation to deliver trustworthy rare-event forecasts for complex engineering systems.

\color{black}

\subsubsection{Mixed Variables}
Dealing with various types of variables in design exploration is also an issue to tackle. That is, the model should not be able to deal with just continuous variables but also integer and categorical variables. Typically, integer variables are easier to handle than categorical variables since there is at least a level of hierarchy for the former, while it is not clear for the latter. Neural networks are versatile methods capable of accommodating these variable types. Recent advancements in surrogate modeling include the development of a mixed-categorical GP~\cite{saves2023mixed, munoz2020global, moustapha2022multi}, which has been successfully applied in general approximation and optimization tasks, \textit{e.g.}, mixed variable aircraft design~\cite{yan2024explainable}. The primary challenge in kernel-based methods, such as GP for mixed variables, is how to define the distance for such variables. Often, variable encoding is needed for handling categorical variables to represent them in a numerical format. 

From the viewpoint of explainability, the methods should also be able to extract insights from mixed variable models in a way that humans can understand, though interpreting these models can be more challenging than working with continuous variables alone. Notably, explainability techniques developed in ML, like SHAP, are generally capable of handling mixed variables, including categorical ones (some examples in engineering applications include geology with soil type as a categorical variable~\cite{kannangara2022investigation}). With some treatments, PDP and ICE have also been extended and used for categorical variables~\cite{inglis2022visualizing}. However, some methods are not compatible with mixed variables, making it essential to recognize each method's limitations. For instance, GSA techniques using Sobol' indices are inappropriate for categorical variables, as variance decomposition is grounded in continuous variables. Similarly, the ASM is limited to continuous variables because it relies on computing the covariance matrix of the gradient, making it inapplicable to mixed variable cases. 

One particular challenge is when hierarchical variables are present. The challenge primarily lies in quantifying the impact of a certain variable that depends on the activation of another input variable, which is a hierarchy above the former. Such variables can be found in several engineering applications, including aircraft design~\cite{fouda2022automated}. To the best of our knowledge, no specific explainability method exists that can properly handle such problems. Therefore, the development of explainability methods capable of handling hierarchical variables is an important research direction. 

Explainability in the mixed-variables domain must also account for the semantic meaning of different variable types and their interactions~\cite{saves2023mixed}. For example, the influence of a categorical design choice cannot be interpreted in the same manner as that of a continuous parameter, and interactions between discrete and continuous variables may be obscured or exaggerated by surrogate approximation. This introduces an additional layer of complexity, where explanations reflect both the modeled relationships and the representational assumptions used to handle mixed variables. 

Explainability for mixed-variable surrogate models must be interpreted with particular care, as explanations are conditioned not only on data fidelity and approximation accuracy, but also on variable representation and encoding choices. \textcolor{black}{For example, using smooth surrogates such as naive GP to model integer variables may impose artificial continuity~\cite{garrido2020dealing}. In that case, some explainability methods, such as PDP and ICE may attribute gradual trends to what are, in reality, discrete transitions. Similarly, when categorical variables are represented using one-hot encoding but without the rounding and projection mechanisms, the naive surrogate model might treat each category as a separate binary variable, without explicitly accounting for the fact that only one category can be active at a time. As a result, naive explainability methods may assign separate contributions to each encoded variable, which can make it harder to interpret the original categorical feature~\cite{robani2025}.}


\subsubsection{Correlated Input Variables}
Correlated input variables continue to pose significant challenges in explainability, as many attribution and importance methods rely on independence assumptions that may not hold in practice~\cite{aas2021explaining}. While correlated inputs are generally not an issue in design optimization and exploration, they can be present in areas such as uncertainty quantification and reliability analysis~\cite{nohreliability2009, liurigorous2021}. 

In the context of explainability, correlated inputs can obscure the individual contributions of features to model predictions, making it difficult to accurately attribute the importance and impact of each individual input~\cite{hooker2021unrestricted}. The possibility of misleading interpretations increases when dealing with such a characteristic, and thus, it is important to understand the limitations of each explainability method. In a pure data-driven setting, it is highly important to check the dependency of inputs by simple means, such as correlation coefficients. The dependency can be characterized through various methods, such as Copulas~\cite{charpentier2007estimation}, which distinguish each variable's marginal distribution from its dependence structure. \textcolor{black}{These challenges become even more pronounced in surrogate modeling. Since surrogate models are approximate, any bias introduced by correlated inputs may propagate through both the surrogate construction and the subsequent explainability analysis.}

There are several advancements in this direction. In the context of GSA, Sobol' indices become challenging to interpret when there is statistical dependence between input variables. On the other hand, in contrast to Sobol' indices, Shapley effects are naturally able to handle correlated inputs since they rely on the conditional variances~\cite{iooss2019shapley, demange2023shapley, song2016shapley}. Both PDP and ICE rely on marginal distribution and ignore any dependency between inputs, making them unsuitable for correlated inputs~\cite{apley2020visualizing}. Despite its wide use, SHAP is also grounded on the assumption that the inputs are independently correlated. Some methods address this issue by implementing specific treatments; for example, ALE uses the conditional distribution to compute differences in predictions rather than averages~\cite{apley2020visualizing}, which aids in both visualizing and determining the importance of inputs. \textcolor{black}{
Shapley values have been extended to the dependent cases and are called “target Shapley effects”, allowing for \textcolor{black}{the computation of} sensitivity measures under dependent inputs~\cite{idrissi2021developments,demange2023shapley}. Furthermore, the KernelSHAP algorithm has been modified by avoiding the independence assumption through the estimation of the conditional distribution~\cite{aas2021explaining}. Moreover, an extension of ASM to handle dependent inputs has been proposed, where the formulation is based on a dependent gradient rather than the standard gradient~\cite{lamboni2026active}.} 


\subsubsection{High-dimensional Inputs and Outputs}
Explainability methods for high-dimensional input variables remain challenging. One particular challenge is the high computational cost of explaining itself. For example, KernelSHAP is computationally prohibitive for high-dimensional problems, which is why the development of analytical methods to perform such a task can be greatly beneficial. Regardless, KernelSHAP itself is an approximation method; otherwise, it is impractical to use exact methods for computing Shapley values for high-dimensional inputs. Indeed, there exist fast methods for methods such as neural networks (deepSHAP)~\cite{lundberg2017unified}, PCE~\cite{palar2023enhancing}, and random forest~\cite{lundberg2018consistent}. However, for other surrogate models, such an analytical form might not be available. This leads to the issue of scalability in explainability.

The next issue is the interpretation of high-dimensional inputs, although arguably this is not a bigger issue than the first issue mentioned in the paragraph before. One simple method is that the GSA and explainability can be performed in two stages. Initially, the GSA identifies the most crucial variables within a dataset. Subsequently, the explainability process is applied exclusively to this selected set of variables. This approach enables users to concentrate their analysis on these essential variables, rather than exhaustively examining every variable. Insignificant variables, on the other hand, may not require further explanation, as they have minimal impact on the desired output. 

It is also important to distinguish high-dimensional inputs and high-dimensional outputs (\textit{i.e.}, if the surrogates are constructed by considering high-dimensional outputs). One simple method to perform explainability is to treat each output individually. However, for correlated high-dimensional outputs, point-wise explainability analysis may be inadequate when dealing with physical field variables (\textit{e.g.}, stress field or velocity field), as interpreting individual points disregards the spatial correlations. Examples of such surrogate modeling cases include flow-field predictions in fluid mechanics~\cite{li2023fast, duru2022deep}, but these applications did not discuss explainability. One viable approach to address this challenge involves first applying dimensionality reduction techniques (\textit{e.g.}, principal component analysis) to the output space. This enables subsequent explainability analysis to be performed on the reduced-dimensional representation, significantly simplifying the interpretation process. Indeed, recent advances in surrogate modeling have demonstrated that such issues can be effectively managed through appropriate dimensionality reduction strategies, such as singular value decomposition and diffusion mapping~\cite{guo2023investigation}.

\textcolor{black}{In the context of multi-objective design, the application of explanation ML becomes more challenging as the number of objectives increases. While the explainable ML framework for two objectives~\cite{palar2024multi} can be used by analyzing two objectives one at a time, the development of an efficient framework for multi-objective in a high-dimensional objective space has the potential to streamline the analysis process. To the best of our knowledge, this has not yet been thoroughly explored. Nevertheless, it is important to consider that simplifying the complexity of the objective by reducing the number of objectives is sometimes deemed advantageous. Having a high number of objective functions is not always necessary and should be avoided if the optimization problem can be well described with only a few objectives~\cite{purshouseconflict2003}}

One interesting research direction is to combine explainability with dimensionality reduction. That is, the explainability methods can also provide the means for dimensionality reduction. Although not in the context of dimensionality reduction, information from SHAP has been used to refine the bounds of an optimization problem for a more targeted optimization~\cite{weeratunge2025interpretable}. We then expect that information derived from the explainability method can also be leveraged for dimensionality reduction.

\textcolor{black}{High-dimensional input and output spaces pose a fundamental challenge for explainable analysis in surrogate modeling, \textcolor{black}{but the nature of data limitations differs from typical ML settings. In typical  surrogate modeling applications, data are generated from numerical simulators and each sample may be computationally expensive, leading to a constrained evaluation budget~\cite{saves2024high}. On the other hand, some general ML applications are constrained by scarce or difficult-to-acquire data (\textit{e.g.}, sensor measurements, such as fluid-flow measurements~\cite{bao2024integrating}), for which additional data cannot be readily generated. As a result, high-dimensionality in surrogate modeling is primarily associated with limited sampling density under a fixed or limited computational budget~\cite{hou2022dimensionality} (the well known curse-of-dimensionality as first coined by Richard Bellman in the context of dynamic programming~\cite{keogh2017curse}) rather than intrinsic data scarcity. In the context of explainability, this leads to sparse coverage in high-dimensional spaces and can lead to unstable or biased explanations~\cite{holzinger2018machine}.} For this reason, it is important to quantify the uncertainty associated with explainability measures, as discussed in the next section. Notably, the need to assess such uncertainty applies not only to high-dimensional settings but also to low-dimensional cases.}

\subsubsection{Uncertainty Estimation of Explainability Metrics}
\textcolor{black}{Explanations may appear deceptively precise, potentially leading scientists or engineers to overtrust specific input attributions or model rationales that are, in reality, might  be sensitive to the underlying data, approximation error, or computational approximations. Estimating uncertainty in explainability metrics is then important to ensure robust model interpretations~\cite{alvarezmelisrobustness2018, deuschelrole2024}.}

 It is important first to understand that uncertainty in the explainability metrics might come from at least two different sources. The first stems from limited data, as previously discussed. The second originates from the approximation methods used to compute the metrics themselves, for instance, the finite number of samples in a \textcolor{black}{Monte Carlo} simulation~\cite{slack2021reliable}. While both sources are relevant, we argue that the former is more critical, as the only way to reduce this uncertainty is by acquiring more data, an option that is not always feasible. Nevertheless, the latter should not be overlooked and must also be considered in the analysis. \textcolor{black}{Furthermore, the data used to construct the surrogate may already incorporate uncertainty due to numerical discretization, modeling assumptions, or measurement noise~\cite{kennedybayesian2001}, which further propagates into the explainability results. From a verification, validation, and uncertainty quantification perspective, this highlights that explainability metrics should not be treated as deterministic quantities, but rather as uncertain estimates conditioned on data fidelity, sampling design, and surrogate accuracy. Quantifying and communicating this uncertainty is therefore important for responsible interpretation of explainability results and for avoiding overconfident conclusions.}

Within a wider context, the uncertainty estimate of ML is an emerging important research~\cite{abdar2021review}. Particularly in scientific ML, which deals with differential equations containing unknown parameters, estimation of uncertainty becomes a paramount task~\cite{psaros2023uncertainty}. Some methods, like GP and Bayesian neural networks~\cite{martin2024}, inherently incorporate uncertainty structures because of their probabilistic definitions. For deterministic models, a simple method is to apply resampling methods to estimate the associated prediction uncertainties. For example, bootstrapping has been used to estimate uncertainties due to the limited sample size~\cite{marelli2018active, modak2024enhanced, regis2022bootstrap}. In the context of explainability, it is crucial to indicate the reliability of the model's output~\cite{poyiadzi2021understanding} so bootstrapping can be leveraged in the context of local explanations~\cite{gosiewska2019ibreakdown}. 

An attempt has been made to quantify the uncertainty in the PDP extracted from a probabilistic surrogate model~\cite{moosbauer2021explaining}. In this sense, it is possible to provide uncertainty of a PDP to, for example, GP and Bayesian neural networks. Arguably, one of the current issues is how to provide a reliable uncertainty estimate for a deterministic model in a cheap manner. The conformal prediction is one important technique currently investigated, and its use for providing uncertainty estimates has been demonstrated in the context of explainability~\cite{yapicioglu2024conformasight, alkhatib2024estimating, ibrahim2024transparent}.

We argue that providing uncertainties on explainability metrics is an important research direction and should be actively pursued. As explainable ML methods become more integrated into scientific and engineering workflows, understanding not just what a model explains but also how confidently it does so becomes essential.

\subsubsection{Choosing Explainability Methods}
A specific challenge in selecting among various explainability methods lies in the need for a thorough understanding of the factors that these methods elucidate. Failure to comprehend what these methods reveal may lead to potential misunderstandings. Possible sources of pitfalls might include the use of unsuitable ML models, those that stem from the limitation of explainability methods, and wrong applications of the explainability methods.
To steer clear of potential pitfalls, users of explainable ML methods should ensure a thorough understanding of the techniques employed. Further, the lack of quantitative evaluation metrics to compare various explainability methods remains a challenge~\cite{verma2021pitfalls}, although some quantitative metrics have been proposed. It is worth noting that, if applied improperly, the use of explainability methods might lead to several pitfalls, including an unsuitable ML model, the limitation of the explainability method itself, and the wrong application of the explainability method~\cite{molnar2020general}. Explainability methods might also disagree in practice~\cite{krishna2022disagreement}, which is why it is important to understand the quantities calculated by each explainability method.

Users should initially state their primary objective in employing explainability methods. Is it specifically for GSA? Or is it to enhance the understanding of the input variables' impact? Perhaps it is to analyze potential trade-offs between objectives? By having a clear goal in mind, we can tailor our approach to the specific requirements of the task and maximize the effectiveness of the chosen explainability methods. One should also exercise restraint in the application of explainability methods~\cite{rudin2019stop}. If the problems are relatively simple and can be adequately approximated using simple methods such as linear regression, there may be no necessity to employ sophisticated explainability methods.

It is worth noting that there is no single "silver bullet" explainability method that can uncover all knowledge in a single shot (\textit{i.e.}, NFL in explainability~\cite{holzinger2018machine}). Rather, all methods have their advantages and disadvantages and can reveal knowledge from different viewpoints. For instance, ASM is useful for detecting the presence of single or multiple optima by projecting the input samples into the reduced space, while it is not trivial to do this task using methods such as SHAP and PDP. On the other hand, SHAP and PDP are more capable of illuminating the individual impact of input variables in a more decipherable way than ASM. When users are armed with sufficient knowledge of the advantages and potential pitfalls of various explainability methods, they can apply several of these methods simultaneously and derive maximum benefits by gaining insights from various perspectives.

\subsection{Human-centered challenges}
\textcolor{black}{The second category involves \emph{human-centered challenges}, which address how explanations are interpreted, communicated, and critically evaluated in real-world decision making context. This category encompasses the relationship between explainability, fairness, and privacy; the development of automatic explanation systems; immersive and stakeholder-oriented communication strategies; and broader criticisms concerning the conceptual foundations and limitations of XAI.}

\subsubsection{Relationships with Fairness and Privacy}
\label{sec:fairness_privacy}

Interpretability and \textcolor{black}{explainability} in ML do not stand alone: it interacts with fairness and privacy in ways that, even if generally opposed, can produce both trade-offs and opportunities~\cite{ferry2023sok}. 
Simplifying a model to improve human comprehensibility can reduce predictive accuracy for underrepresented subgroups and reveal information on the training data, while fairness interventions that force better fits on rare cases may increase the influence of individual training examples and thus raise membership or reconstruction-attack risks. 
Similarly, post-hoc explanations (for example, feature importances, counterfactuals, or influential examples) can unintentionally disclose sensitive inputs or reveal rare data traces~\cite{shilova2025fairness}.
These tensions have been observed both theoretically and empirically, and a careful characterization is essential for informed design choices. Important open questions include precisely accounting for the (direct or indirect) privacy cost introduced by fairness mechanisms and rigorously quantifying the privacy risks posed by different explanation formats~\cite{saifullah2024privacy}. At the same time, synergies are identified in~\cite{ferry2023sok}: explainable ML can ease fairness audits and debugging, certain fairness constraints may act as regularizers that improve model simplicity, and, in some designs, privacy-preserving mechanisms can be adapted to support forms of individual fairness.

Concrete examples illustrate these trade-offs and reconciliations. Work that analyzes or modifies the distribution of noise injected for privacy has been used both to improve and to certify statistical fairness~\cite{ignatiev2020towards}; pruning and sparsity-promoting techniques have been employed to learn more interpretable models under fairness constraints; and explanation methods have been leveraged to detect privacy leakages. Extensions of differential privacy, such as metric differential privacy, offer promising tools for enforcing individual fairness by generalizing the notion of neighbouring databases to metric spaces~\cite{lafargue2025exposing,nguyen2025privacy}.

Missing data mechanisms, for example, privacy in the medical domain, can further complicate fairness and interpretability: the choice of imputation or deletion may disproportionately affect rare component behaviors or hierarchical levels, leading to algorithmic bias and interpretability distortions in problems such as in system architecture design~\cite{saves2025hierarchical,josse2024consistency}.

From a game-theoretic standpoint, XAI and fairness can also be framed as cooperative or competitive games: individual features or design variables form coalitions in attribution methods (\textit{e.g.}, Shapley methods) or fairness interventions, and their interactions may lead to privacy risks or shifts in fairness outcomes based on the chosen value-functions~\cite{idrissi2025beyond}. This approach of treating XAI as a fair allocation problem can lead to many GSA metrics, such as the proportional marginal effect, that are well-adapted for feature selection in an engineering setup~\cite{herin2024proportional}.

For engineering and high-stakes systems, we therefore recommend that practitioners: (i) assess fairness and privacy alongside \textcolor{black}{explainability} at each hierarchical level (system/subsystem/component), (ii) empirically evaluate explanation quality and subgroup/hierarchical‐level effects and privacy risk, and (iii) involve interdisciplinary review (technical, legal, domain) when selecting trade-offs. While differentially-private explanation methods exist, they impose explicit trade-offs between explanation fidelity, privacy guarantees, and model utility, and their effects in hierarchical engineering workflows r.main an active research frontier~\cite{le2024algorithmic}.

\subsubsection{Automatic Explanations}
Despite the potential of knowledge extraction from explainability methods, sometimes it is still difficult to decipher for untrained users, especially when the explainability methods used are non-intuitive. Further, there is a possibility that one misses important information even when it is already extracted by the explainability method. In this regard, we believe that this role can be filled by an AI-assistant who helps users in explaining the results from explainability. 

One potential direction is to leverage trustworthy Large Language Models (LLMs) that can provide text-based explanations of or at least automate the data analysis process (as demonstrated in several papers,  \textit{e.g.}, ~\cite{jansen2025leveraging}) from the results of explainable models. To that end, a LLM can be incorporated, trained on explanatory texts written by expert users, to aid inexperienced or even moderate users in extracting important information from the explainability method. Also, we think that such a model can also give meaningful suggestions to the user on how to perform explainability analysis on the data and model. In this sense, our aim is to alleviate the complexity of explanations provided by various methods for users who may come from diverse backgrounds and have different levels of expertise. This will, without doubt, accelerate the adoption of explainability techniques in data-driven engineering.

\color{black}

\subsubsection{Immersive Communication to Stakeholders}

While standard 2D XAI visualizations provide rigorous mathematical attribution, they often remain opaque to non-experts. Effectively informing high-stakes decisions requires translating statistical metrics into intuitive, multimodal experiences~\cite{do2024exploring}. 
First, transitioning to natural language via LLMs allows systems to frame complex behaviors as causally linked narratives rather than abstract graphs~\cite{arnejo2025automatic}. When combined with surrogate-based argumentation, this enables interactive "debates" where emulators transparently justify predictions using specific physical or behavioral rules~\cite{albini2020deep}.
Second, \emph{interactive explainability} via dashboards and digital twins allows stakeholders to query models in real-time~\cite{predhumeau2025sustainable}. Users can inject counterfactuals (such as adjusting a policy parameter) to immediately observe dynamic shifts in both the prediction and its underlying explanation network, fostering a deeper understanding of system tipping points.
Finally, for spatially complex domains like fluid dynamics or urban modeling, 2D plots are fundamentally insufficient. Integrating XAI with Virtual and Augmented Reality (VR/AR) allows metrics to be projected directly onto 3D geometries. For instance, mapping SHAP values as heatmaps onto a VR Computer Aided Design (CAD) model enables instant visual comprehension of the geometric drivers behind structural failures~\cite{maathuis2025integrating}. Developing real-time attribution algorithms to support this immersive explainability remains a highly promising research frontier.

\color{black}

\subsubsection{Criticism of Explainability}
While explainability in ML holds great promise, it is essential to critically examine its limitations to ensure informed and effective application in real-world scenarios. These criticisms range from the criticism of the explainability framework itself to that directed toward specific methods. This is quite a broad topic by itself, hence we only discuss essential parts to understand the main points of criticism that are relevant within the context of \textcolor{black}{this review}. A good summary of this issue is discussed in Herrera~\cite{herrera2024attentiveness}.

As Freisleben and König~\cite{freiesleben2023dear} argue, much of current XAI research lacks solid conceptual and methodological foundations. The development of new techniques has often prioritised presentation over purpose. They suggest that research should first clarify the intended purpose of explainability before moving toward benchmarking. Accordingly, appropriate evaluation metrics need to be established to enable fair comparisons between different explanation methods. Indeed, research on developing evaluation metrics for explainability is progressing (\textit{e.g.}, measuring them in terms of simplicity, completeness, soundness, and stability of explanations~\cite{coroama2022evaluation}. This is grounded on an important issue that different explanations are neither standardised nor assessed in a systematic manner~\cite{gilpin2018explaining}. One important thing is that, should we want to develop such evaluation metrics, they should be objective. For example, Rosenfeld presents four objective measures~\cite{rosenfeld2021better}, namely\,:
\begin{itemize}
    \item D: Measures the disagreement between the explanations from a model and the best transparent model
    \item R: The number of rules used in the explanation (\textit{i.e.}, simplicity)
    \item F: The number of input variables (\textit{i.e.}, features) used for the explanation.
    \item S: Stability, that is, robustness to small perturbations.
\end{itemize}
 Meaningful comparisons between explainability methods remain a challenge. In this context, we argue that the most pragmatic perspective is to recognise that each method uncovers knowledge from a different viewpoint (in the context of knowledge discovery). For instance, consider two GSA methods: Sobol' indices and DGSM. While Sobol' indices rely on variance decomposition, DGSM evaluates sensitivity through the average of local derivatives. Determining which method is superior is challenging (and is it meaningful?) since each captures global sensitivity from a distinct perspective.

Another important point that should not be overlooked is that explainability systems lack a self-regulative feature~\cite{peters2023explainable}, unlike human reasoning, which triggers self-regulation, making human decision-making more transparent and trustworthy. In the context of knowledge discovery, we argue that AI lacks this dynamic process of reflection and alignment between reasons and internal mechanisms. Human reason-giving often leads to the refinement of beliefs, the correction of inconsistencies, and even the generation of new hypotheses. \textcolor{black}{In this regard, advancing the field of ML, especially explainable and interpretable ML, fundamentally requires fostering effective human-ML interaction~\cite{dengintegrating2020}}  Therefore, while current state-of-the-art explainable ML methods can support the scientific discovery process, the human element remains crucial for enabling the dynamic, self-regulatory reasoning that such discovery often requires~\cite{mosqueirareyhuman2023, tocchettirole2022}. 

It is also worth reflecting on the question: "Do I truly need complex explainability methods?" Echoing the principle of Occam's Razor, it is also important to always encourage simpler explanation outputs in the context of knowledge discovery~\cite{domingos1999role}. When a straightforward, interpretable model performs comparably to a more complex black-box model, the simpler option is often the better choice. However, when accurate approximations are difficult to obtain from simple models, using post-hoc explanation is a plausible way to move forward.



\section{Conclusions}
\label{sec:conclusion}

\textcolor{black}{
As complex simulations become increasingly central to engineering design, scientific exploration, and policy analysis, the tension between computational efficiency and model transparency has grown acute. For years, the standard solution has been to deploy surrogate models to reduce computational costs, often at the expense of exacerbating black-box opacity. Meanwhile, XAI has matured into a powerful discipline, yet it has largely evolved separately, struggling to adapt to the rigorous constraints of physical and rule-based simulators.   This survey has demonstrated that these two paradigms, \textit{i.e.}, surrogate modeling and XAI, must no longer remain disjoint. By systematically integrating them, we can mutually improve both: surrogates gain the missing diagnostic transparency required for stakeholder trust, while XAI methods benefit from the computational efficiency and controlled environments that surrogates provide. Ultimately, leveraging the synergy between global and local tools through Visual XAI is a critical enabler for robust, actionable decision-making. }

\textcolor{black}{Moving forward, the research community must tackle the pressing challenges identified in this review. Standard XAI metrics must be adapted to handle highly correlated input structures, mixed-variable spaces, and the temporal complexities of dynamical systems, to name a few. Furthermore, the robustness and uncertainty of the explainability metrics themselves must be rigorously quantified. Most importantly, achieving trustworthy simulation-driven design requires a paradigm shift. Explainability can no longer be treated as a post-hoc reporting tool or an optional add-on. Instead, it must become a core, embedded element of the modeling workflow, integrated from the initial design of experiments through surrogate training, validation, and final decision support.}

\appendix
\section{Glossary of key terms and concepts}
\label{Glossary}

The integration of explainability methods into surrogate modeling for engineering and science introduces terminology from both ML and engineering sciences. Given the interdisciplinary nature of this domain, the absence of a standardized lexicon can pose challenges for effective communication and knowledge transfer among researchers and practitioners. To address this, we include in this appendix a selection of key terms and definitions used throughout the paper. These definitions, adapted in part from the IAPP AI Governance Glossary\,\cite{iappai_glossary}, are intended to promote conceptual clarity and support a shared understanding of the foundational concepts relevant to explainable surrogate modeling and knowledge discovery in engineering contexts.

\begin{itemize}

    \item \textcolor{black}{\textbf{Active Learning / Adaptive Design of Experiments (DoE).}
An iterative sampling strategy where an algorithm actively queries the most informative data points, which are then evaluated next using the underlying model or experiment (\textit{e.g.}, Bayesian optimization). This term is highly interdisciplinary. The engineering community refers to it as Adaptive DoE: optimizing the parameter space to minimize expensive simulation runs. The ML community refers to it as Active Learning (strategic data acquisition for minimizing the time and cost of data collection).}
    
   \item  \textcolor{black}{\textbf{Agent-Based Model (ABM).} A bottom-up computational simulation paradigm that models the simultaneous operations and interactions of multiple autonomous agents to assess their effects on the system as a whole. It is characterized by emergent behavior, stochasticity, and decentralized decision-making.}

    \item \textcolor{black}{\textbf{Black-Box Model.} A highly complex predictive model or simulator (such as a deep neural network or large ensemble) whose internal mechanics, mathematical structures, and decision-making processes are opaque and cannot be straightforwardly understood by a human observer without auxiliary post-hoc tools.}
    
    \item \textcolor{black}{\textbf{Counterfactual Explanation.} A contrastive explanation method that identifies the minimal necessary changes to an input variable required to alter the model's predicted outcome to a desired target state. It essentially answers the "what-if" or "how to change" questions, making it highly actionable for engineering design.}

    \item \textbf{Decomposability.} The ability of the model to express its output as a sum (or composition) of contributions from individual input variables or their interactions. This property enables clear interpretation of how each input affects the output, making the model more transparent and analyzable.

    \item \textbf{Explainability.} The ability to describe or provide sufficient information about how a ML system generates a specific output or arrives at a decision in a specific context to a predetermined addressee. XAI is important in maintaining transparency and trust in ML.

    \item \textbf{Fairness.} An attribute of a ML system that prioritizes relatively equal treatment of individuals or groups in its decisions and actions in a consistent, accurate manner. Every model must identify the appropriate standard of fairness that best applies, but most often it means the ML system's decisions should not adversely impact, whether directly or disparately, sensitive attributes like race, gender, or religion.
    
    \item \textcolor{black}{\textbf{Feature Attribution.} A specific class of explainability methods (such as SHAP or LIME) that distributes the model's prediction score among the input variables, quantifying the exact directional impact and magnitude that each feature contributed to the final output.}
    
    \item \textcolor{black}{\textbf{Fidelity.} A highly overloaded, context-dependent term. In \textcolor{black}{equation-based} simulations, it refers to the accuracy with which a computational solver represents the true physical phenomena, as in high-fidelity (precise) \textit{vs}. low-fidelity (fast) physics. In the context of XAI, \textit{explanation fidelity} refers to how accurately a local interpretable explainer (like LIME) mimics the underlying black-box model's actual decision boundary.}

   \item  \textcolor{black}{\textbf{Global Explainability.} Techniques that capture the overarching, system-wide behavior of a model across the entire input space. These methods (such as GSA or global surrogate models) identify which input features are the most dominant drivers of the model’s predictions overall.}

    \item \textbf{Interpretability.} The ability to explain or present a model's reasoning in human-understandable terms. Unlike explainability, which provides an explanation after a decision is made, interpretability emphasizes designing models that inherently facilitate understanding through their structure, features, or algorithms. Interpretable models are domain-specific and require significant domain expertise to develop.
    
    \item \textcolor{black}{\textbf{Local Explainability.} Techniques that diagnose the model's behavior at the level of a single, specific prediction or instance. These methods detail exactly why a particular input vector resulted in a specific output, isolating the local feature interactions that drove that exact decision.}

    \item  \textcolor{black}{\textbf{Rashomon Effect (Model Multiplicity).} A statistical phenomenon where multiple, fundamentally distinct predictive models can achieve the exact same level of accuracy on a given dataset. This creates profound ambiguity in XAI, because different, yet equally accurate surrogate models will yield entirely conflicting feature attributions and physical explanations for the same system.}
    
    \item \textbf{Reliability.}      In the context of a surrogate ML model, reliability refers to the model’s ability to generalize. A reliable ML system delivers both consistent and accurate predictions or calibrated uncertainty estimates, even on new or slightly shifted inputs. This notion emphasizes the trustworthiness of the model behavior and not physical robustness.  
    
    In contrast, in engineering design or ABM contexts, reliability denotes the probability that a system or component will perform its intended function under specified conditions and for a defined period of time. It is fundamentally about operational resilience, failure rates, maintainability, certification, and endurance in the face of varying, uncontrolled environments.  

    \item \textbf{Robustness.} An attribute of a ML system that ensures a resilient system that maintains its functionality and performs accurately in a variety of environments and circumstances, even when faced with changed inputs or adversarial attacks.

    \item \textbf{Safety.} The development of ML systems that are designed to minimize potential harm, including physical harm, to individuals, society, property, and the environment.
    
  \item   \textcolor{black}{\textbf{Sensitivity Analysis.} The study of how uncertainty in the output of a mathematical model can be allocated to different sources of uncertainty in its inputs. While often used interchangeably with "feature importance" in ML, traditional engineering sensitivity analysis explicitly maps variance propagation  across the physical design space, rather than just assigning predictive weights.}

   \item  \textcolor{black}{\textbf{Surrogate Model (or Emulator).} A computationally efficient ML approximation of a complex, resource-intensive high-fidelity simulator. It is trained on a limited set of input-output data points to enable rapid design space exploration, though it inherently introduces approximation errors.}

    \item \textbf{Transparency.} The extent to which information regarding a ML system is made available to stakeholders, including disclosing whether ML is used and explaining how the model works. It implies openness, comprehensibility, and accountability in the way ML algorithms function and make decisions.

    \item \textbf{Trustworthy.} In most cases, used interchangeably with the terms responsible ML and ethical ML, which all refer to principle-based ML governance and development, including the principles of security, safety, transparency, explainability, accountability, privacy, nondiscrimination/non-bias, among others.

   \item  \textcolor{black}{\textbf{Uncertainty (Aleatoric \textit{vs.} Epistemic).} A bifurcated concept frequently conflated in black-box modeling. \textit{Aleatoric} uncertainty refers to the inherent, irreducible stochastic variability in a physical system or agent-based environment. \textit{Epistemic} uncertainty refers to reducible ignorance or lack of knowledge about the system, which can be mitigated by acquiring more simulation data or refining the surrogate architecture.}

   \item  \textcolor{black}{\textbf{Uncertainty Quantification.} The mathematical science of characterizing and estimating the uncertainties associated with computational model predictions.} 

    \item \textcolor{black}{ \textbf{Verification and Validation (V\&V).} Distinct but complementary evaluation procedures often erroneously used as synonyms. \textit{Verification} asks "are we building the model right?" (ensuring the computational/code implementation strictly matches the mathematical description). \textit{Validation} asks "are we building the right model?" (ensuring the surrogate or simulator actually aligns with the real-world physical or social ground truth).}

\end{itemize}

\subsection*{Author contributions}

Conceptualization: [P.P., P.S., N.V., M.G.];
Methodology: [P.P., P.S., M.G., N.V., B.G.]
Formal analysis and investigation: [P.P., P.S., N.V., M.G., K.S., J.A.];
Writing - original draft preparation: [P.P., P.S., M.G., J.A., K.S., N.V.]; 
Writing - review and editing: [P.P., P.S., B.G., M. R., M.G., N. V.]; 
Funding acquisition: [P.P., N.V.]; 
Resources: [M.G., J.A., J.M., N.V.];
Validation: [M.G., K.S., J.A., N.V.]
Applications: [P.P., P.S., M.R., J.M., N. V., B.G.]; 

\subsection*{Data availability}
As this is a review and position paper, there are no datasets associated with this study. All results are derived from previously published work, which is cited accordingly.

\subsection*{Funding}
This work was supported by Institut Teknologi Bandung through the Riset Kolaborasi Internasional 2024 scheme. This research was also funded by the Indonesian Endowment Fund for Education (LPDP) on behalf of the Indonesian Ministry of Higher Education, Science and Technology and managed under the EQUITY Program (Contract No. 8128/IT1.B07.1/TA.00/2025).  This work is part of the activities of ONERA - ISAE - ENAC joint research group. The research presented in this review paper has been performed in the framework of the MIMICO research project funded by the Agence Nationale de la Recherche (ANR) n$^o$ ANR-24-CE23-0380. 

\section*{Declarations}
\subsection*{Conflict of interest}
The authors have no Conflict of interest to declare.


\begingroup
\tiny
\bibliography{sn-bibliography}
\endgroup

\end{document}